\documentclass[opendfm]{xlance/xlance}

\usepackage{tikz}
\usepackage[edges]{forest}
\usetikzlibrary{mindmap, trees, shadows, arrows.meta}
\useforestlibrary{edges}

\usepackage{booktabs}
\usepackage{multirow}
\usepackage{dcolumn}
\usepackage{longtable}
\usepackage{threeparttable}
\usepackage{threeparttablex}
\usepackage{array}
\usepackage{float}
\usepackage{tcolorbox}
\usepackage{fvextra}
\usepackage{tabularx}
\usepackage{ragged2e}
\usepackage{fontawesome5}
\usepackage{makecell}
\usepackage[table]{xcolor}
\tcbuselibrary{breakable}

\definecolor{xcgreen}{HTML}{2E7D32}
\definecolor{xcorange}{HTML}{F0A202}
\definecolor{xcred}{HTML}{B00020}
\definecolor{xcblue}{HTML}{EAF2FF}

\newcommand{\Yes}{\textcolor{xcgreen}{\faIcon{check-circle}}}
\newcommand{\Part}{\textcolor{xcorange}{\faIcon{adjust}}}
\newcommand{\No}{\textcolor{xcred}{\faIcon{times-circle}}}

\definecolor{takeawayBg}{RGB}{250,252,255}
\definecolor{takeawayFrame}{RGB}{220,230,246}
\definecolor{takeawayAccent}{RGB}{76,119,190}
\definecolor{takeawayTitle}{RGB}{34,78,145}
\definecolor{takeawayTitleBg}{RGB}{235,243,255}

\newtcolorbox{takeaway}[1][]{
    enhanced,
    colback=takeawayBg,
    colframe=takeawayFrame,
    colbacktitle=takeawayTitleBg,
    coltitle=takeawayTitle,
    fonttitle=\bfseries,
    title={Key Takeaway},
    rounded corners,
    arc=3pt,
    boxrule=0.4pt,
    borderline west={2pt}{0pt}{takeawayAccent},
    left=9pt, right=9pt, top=6pt, bottom=6pt,
    #1
}

\usepackage[most]{tcolorbox}
\usepackage{xcolor}

\definecolor{instructionBg}{RGB}{248,252,251}
\definecolor{instructionFrame}{RGB}{218,235,231}
\definecolor{instructionAccent}{RGB}{42,139,130}
\definecolor{instructionTitle}{RGB}{29,112,106}
\definecolor{instructionTitleBg}{RGB}{235,248,246}

\newtcolorbox{instruction}[2][]{
    enhanced,
    breakable,
    colback=instructionBg,
    colframe=instructionFrame,
    colbacktitle=instructionTitleBg,
    coltitle=instructionTitle,
    fonttitle=\bfseries,
    title={#2},
    rounded corners,
    arc=3pt,
    boxrule=0.4pt,
    titlerule=0pt,
    borderline west={2pt}{0pt}{instructionAccent},
    left=9pt,
    right=9pt,
    top=6pt,
    bottom=6pt,
    toptitle=4pt,
    bottomtitle=4pt,
    before skip=8pt,
    after skip=8pt,
    #1
}

\newcolumntype{Y}{>{\RaggedRight\arraybackslash}X}
\renewcommand{\tabularxcolumn}[1]{m{#1}}

\newcommand{\reporttitle}{Making AI Scientists Auditable from Evidence to Claim}
\title{%
    \reporttitle
}

\small{
\author[1]{\equalcontribution{Zijian Wang}}
\author[1,3]{\equalcontribution{Hanqi Li}}
\author[1]{Ziyue Yang}
\author[1]{Zijian Hu}
\author[1]{Shenghan Zuo}
\author[1]{Yunzhe Zhang}
\author[1]{Da Ma}
\author[1]{Danyu Luo}
\author[1,2]{Chenrun Wang}
\author[1]{Jing Peng}
\author[1]{Tiancheng Huang}
\author[1]{Sijia Guo}
\author[1]{Huayang Wang}
\author[1]{Zichen Zhu}
\author[1]{Senyu Han}
\author[1]{Yilu Cao}
\author[3]{Bo Chen}
\author[3]{Xin Chen}
\author[1,3]{Kai Yu}
\author[1,2,3]{\correspondence{chenlusz@sjtu.edu.cn}{Lu Chen}}
}

\small{
\affiliation[1]{X-LANCE Lab, School of Computer Science, Shanghai Jiao Tong University, Shanghai, China\newline}
\affiliation[2]{Shanghai Innovation Institution, Shanghai, China\newline}
\affiliation[3]{Suzhou Laboratory, Suzhou, China}
}
\newcommand{\xcientist}{\textsc{Xcientist}}
\usepackage{needspace}
\usepackage{pdflscape}
\newenvironment{HAclaimcase}[2]{%
  \begin{instruction}[title after break={#1 (continued)}]{#1}\label{#2}\small\raggedright
}{\end{instruction}}
\abstract{

AI research systems can report improved results without establishing that the tested implementation matches the method described in their conclusions. We introduce \xcientist{}, a research harness that connects literature review, component-based idea generation, experiment validation and reporting. An evidence graph retains source passages, methods, baselines and evaluation conditions; component records track design revisions; and validation contracts specify the implementation, comparisons and artifacts required at each stage. We define claim drift as an unresolved mismatch between an accepted claim and its attributed sources, code or experiments. A retrospective audit across agent memory, traffic forecasting and physics-informed learning found audit claim drift rates of 3.6--16.7\% for \xcientist{}, lower than those of each of three comparison systems in all nine task-by-relation comparisons. This measure includes both confirmed mismatches and unverifiable claims; most of the difference arose from fewer unverifiable records. The case studies also document performance gains, unsuccessful revisions and limits on component attribution. This technical report describes the system, its implementation and the records needed to check how a proposed idea became a tested method and a final claim.
}

\metadatainfo{
    \small\makebox[\linewidth][c]{
        \href{https://github.com/OpenDFM/Xcientist}{\github~\textbf{Code}} \quad
        \href{https://kotohanon.github.io/Xcientist/}{\web~\textbf{Project Page}}
    }
}

\begin{document}
\publishdate{\today}
\maketitle
\fancyhead[L]{\small\nouppercase{\reporttitle}}

\tableofcontents

\section{Introduction}
\label{section:intro}

\begin{figure}[!h]
    \centering
    \includegraphics[width=0.8\linewidth,height=\textheight,keepaspectratio]{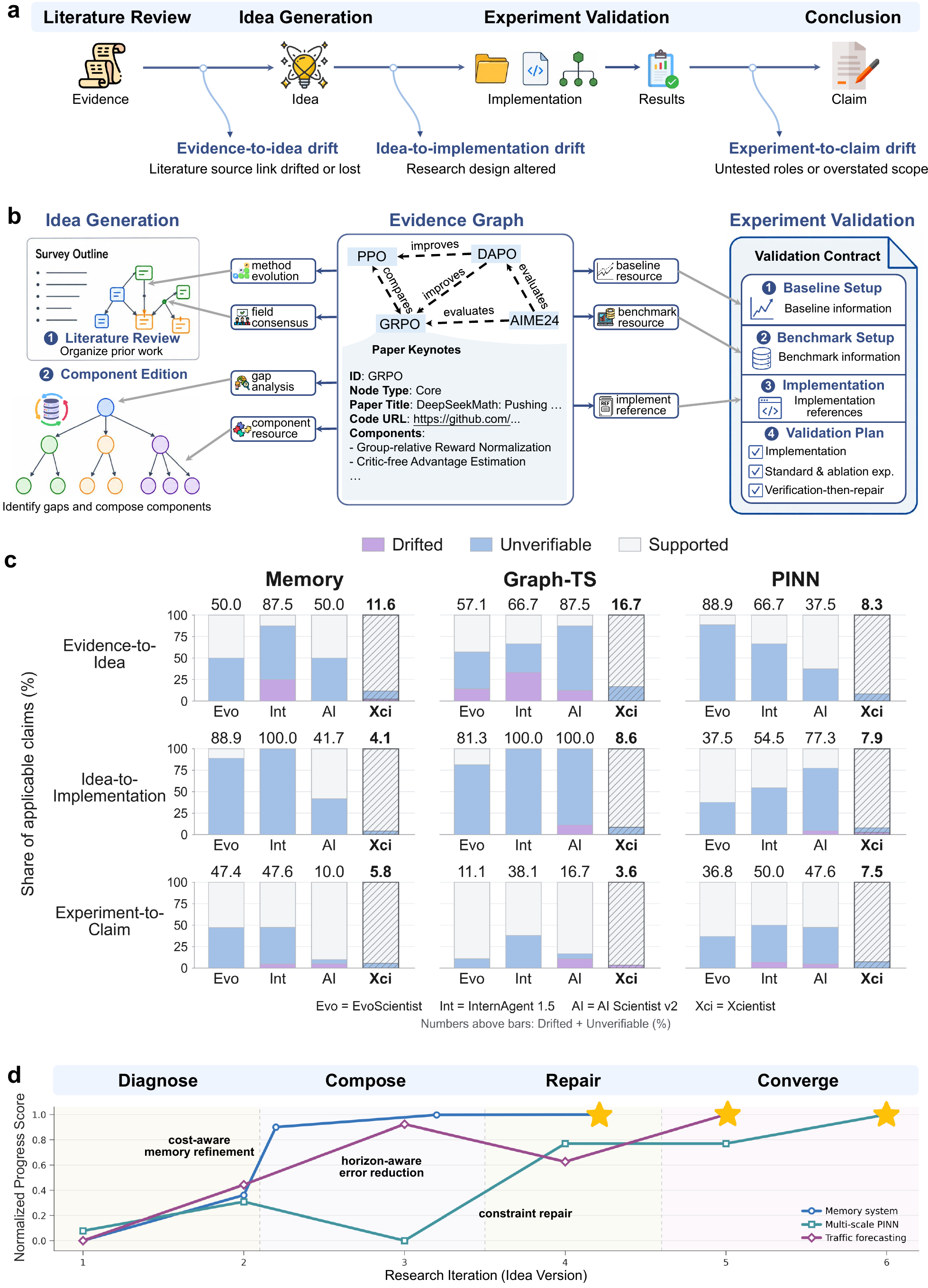}
    \caption{\textbf{From evidence to claim in \xcientist{}.}
\textbf{a}, Three support relations used in the audit. \textbf{b}, The evidence graph, component-based ideation and validation contracts. \textbf{c}, Supported, drifted and unverifiable records across three tasks; labels report audit CDR, \((D+U)/N\). \textbf{d}, Normalized progress across design versions.}
    \label{fig:teaser}
\end{figure}

Scientific results depend on the correspondence between a stated method, the experiment that tested it and the conclusion drawn from that experiment. Automated research makes this correspondence harder to follow when agents revise a proposal, replace code or reinterpret a result across successive stages. A proposed addition to an existing model, for example, may become a replacement during implementation. Its score then supports a different comparison from the one originally planned.

Recent systems combine hypothesis generation, coding, experiment execution and manuscript drafting~\cite{lu2026towards,yamada2025aiscientistv2,lyu2026evoscientist}. They produce many of the artifacts needed to inspect a study, but those artifacts must remain connected as the study changes. Readers need to recover the source behind a design rationale, the code and configuration behind a reported result, and the comparisons supporting a component-level claim. Calls for automated falsification and constraints on autonomous scientists address related concerns~\cite{liu2025falsify,zhu2025safescientist}. We call an unresolved mismatch between an accepted claim and its attributed evidence \textbf{claim drift}. We distinguish evidence-to-idea, idea-to-implementation and experiment-to-claim relations. Missing records leave a claim unverifiable; they do not by themselves establish a mismatch. A documented design revision is also not drift when the method description and conclusion are updated consistently.

\xcientist{} maintains these relations through a research harness (Fig.~\ref{fig:teaser}). Its evidence graph records source passages, method components, limitations, baselines, datasets and experimental conditions. Literature review and idea generation use these records to frame a problem and propose component changes. Experiment validation returns to the same records when selecting starting code, controls and evaluation protocols. This extends literature-grounded idea generation~\cite{baek2025researchagent,liu2025literaturedata} with an explicit connection to implementation and testing.

Each selected idea is translated into validation contracts specifying inputs, permitted operations, deliverables and acceptance checks. Workers implement and execute the plan; separate validators inspect the outputs before a stage advances. Failed checks, repairs and incomplete comparisons remain in the run record. These contracts apply formal agent-governance principles~\cite{ye2025agentcontracts} to the practical question of what evidence is required to support a research claim.

We evaluate retained records in three settings: training-free memory for LLM agents, graph-structured traffic forecasting and multiscale physics-informed neural networks. The evaluation combines claim-level audits with the design histories and task results. Across the nine task-by-relation comparisons, \xcientist{} had lower audit CDR than EvoScientist, InternAgent-1.5 and AI-Scientist-v2. The case studies explain the scope of this result: records were more often sufficient to verify claims, while component attribution and benchmark generalization still required separate checks. The following sections describe the architecture, results and methods; the appendices retain implementation details, worked examples and prompt contracts.

\section{\textsc{Xcientist} Architecture Overview}
\label{sec:architecture_overview}

\begin{figure}[!h]
    \centering
    \includegraphics[width=\linewidth]{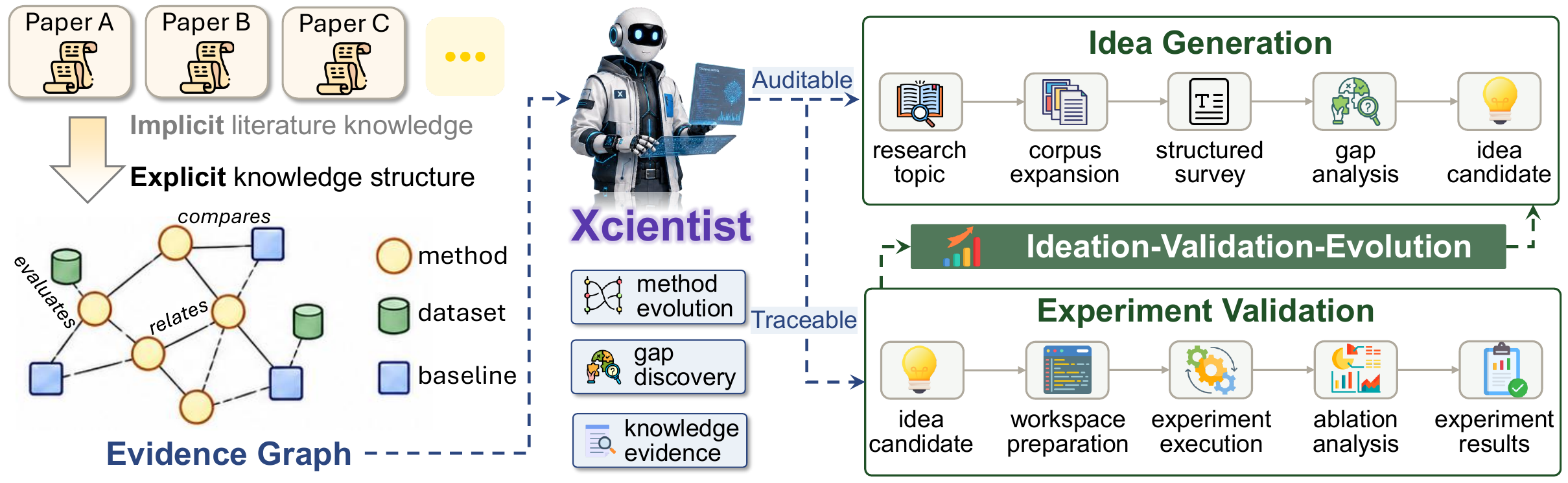}
    \caption{\textbf{\xcientist{} architecture.} The evidence graph supplies source records to component-based idea generation and contract-based experiment validation. The harness retains design changes, executed comparisons and repair decisions.}
    \label{fig:architecture_overview}
\end{figure}

\xcientist{} comprises an evidence graph, a research harness and a user interface. Figure~\ref{fig:architecture_overview} shows how the first two connect literature evidence to implementation and validation. The interface exposes run state, artifacts and decisions for inspection.

The \textbf{evidence graph} stores source passages alongside descriptions of methods, components, limitations, baselines, datasets and evaluation conditions. Both idea generation and experiment planning query these records. A passage used to motivate a design change can therefore be revisited when deciding how to test it.

The \textbf{research harness} coordinates literature review, idea generation, experiment validation and report writing. Ideas have identifiable components, each with a role, source attribution, risks and a proposed test. Search and fusion edit these components, and implementation maps the selected component list to code. Validation contracts define the baseline, controlled changes and required outputs. A component-disable switch makes a comparison possible; an executed, matched comparison is still required to support a contribution claim. Weak or incomplete evidence can prompt a repair or restrict the conclusion to the complete system.

The \textbf{system user interface} groups runs under research topics and exposes review, ideation and experiment lanes. The backend stores artifacts, messages, trace events and approval decisions. Users can inspect failed stages and previous revisions as well as final outputs. Section~\ref{sec:methods} describes the implementation of each module.

Table~\ref{tab:ai_scientist_comparison} summarizes the relevant capabilities of representative research systems. These are design-level comparisons; the claim-level evaluation in Section~\ref{sec:results} assesses the retained records from the three study tasks.

\begin{table}[t]
\centering
\footnotesize
\setlength{\tabcolsep}{1.5pt}
\renewcommand{\arraystretch}{1.18}
\caption{\textbf{Process-accountability capabilities of representative AI scientist systems.} The table records explicit design support, not measured claim reliability.}
\label{tab:ai_scientist_comparison}
\begin{tabular}{@{}lcccccc@{}}
\toprule
\textbf{System}
& \makecell{\faIcon{project-diagram}\\\textbf{Structured}\\\textbf{Knowledge}}
& \makecell{\faIcon{code-branch}\\\textbf{Structured}\\\textbf{Idea Evolution}}
& \makecell{\faIcon{clipboard-check}\\\textbf{Contracted}\\\textbf{Validation}}
& \makecell{\faIcon{tools}\\\textbf{Evidence-driven}\\\textbf{Repair}}
& \makecell{\faIcon{puzzle-piece}\\\textbf{Component}\\\textbf{Attribution}}
& \makecell{\faIcon{search}\\\textbf{Claim-boundary}\\\textbf{Audit}} \\
\midrule
AI-Scientist-v2~\cite{yamada2025aiscientistv2}           & \No   & \No   & \Part & \No   & \No   & \No   \\
DeepScientist-v1.5~\cite{weng2025deepscientist}        & \No   & \No   & \Yes  & \Part & \Part & \No   \\
AI-Researcher~\cite{tang2026ai}             & \Part & \No   & \No   & \No   & \No   & \No   \\
AiScientist~\cite{chen2026toward}        & \No   & \No   & \Yes  & \Part & \No   & \No   \\
InternAgent-1.5~\cite{feng2026internagent}     & \Part & \No   & \Yes  & \Part & \Part & \No   \\
EvoScientist~\cite{lyu2026evoscientist}      & \No   & \Yes  & \Yes  & \Part & \Part & \No   \\
ARIS~\cite{yang2026aris}        & \No   & \No   & \Yes  & \Part & \Part & \Part \\
AlphaEvolve~\cite{novikov2025alphaevolve}       & \No   & \Part & \Part & \Part & \No  & \No   \\
\midrule
\rowcolor{xcblue}
\textbf{\textsc{Xcientist} (Ours)}
                           & \Yes
                           & \Yes
                           & \Yes
                           & \Yes
                           & \Yes
                           & \Yes \\
\bottomrule
\end{tabular}

\vspace{2pt}
\footnotesize{
\Yes{} denotes an explicit system-level mechanism;
\Part{} denotes limited or adjacent support;
\No{} denotes that the capability is not central to the system design.
}
\end{table}

\section{Results}
\label{sec:results}

We examine whether the records support the reported claims and how experimental evidence changed each design. The claim audit compares \xcientist{}, EvoScientist, InternAgent-1.5 and AI-Scientist-v2; the task evaluations compare the methods produced during each design trajectory with their task-specific baselines. Major versions denote mechanism changes, and minor versions denote local refinements, including changes in the memory performance--cost trade-off.

\subsection{Motivating example: Claim drift in automated research}
\label{sec:claim_drift}
\label{sec:results_claim_drift}

\begin{takeaway}
Claim drift is a confirmed mismatch between a claim and its support. Audit CDR counts both drifted and unverifiable claims, so its interpretation requires the separate drifted and unverifiable counts. A design revision with consistently updated claims is not itself drift.
\end{takeaway}

We first examined three AI-Scientist-v2 runs~\cite{yamada2025aiscientistv2} as diagnostic examples (Table~\ref{tab:ed_diagnostic_observations}). Their retained records included a proposed plug-in implemented as a replacement model, component roles without direct tests, and design changes without preserved source links. These cases required two judgements: whether a support relation was broken and whether the available record was sufficient to decide. We therefore separated confirmed mismatches from missing evidence before comparing research systems.

We define \textbf{claim drift} as an unresolved mismatch between a scientific claim and its attributed evidential or experimental basis during research. A claim drifts when it remains presented as supported despite inconsistent source attribution, design identity or experimental support. A new hypothesis, a negative result or an explicitly documented revision is not itself drift.

Formally, let \(\tau=(s_0,\ldots,s_T)\) be a research trajectory and \(\mathcal C_t^{\mathrm{acc}}\) the claims accepted as supported at stage \(t\). For a claim \(c\), \(\mathcal B_t(c)\) denotes its attributed source, design, execution and experimental records. The set \(\mathcal R_t(c)\) contains the relations required to support its current wording. A relation \(r(c,\mathcal B_t(c))\) equals one when the required correspondence holds and zero otherwise. Then
\begin{equation}
\operatorname{Drift}_t(c)
\iff c\in\mathcal C_t^{\mathrm{acc}}
\;\land\;
\exists r\in\mathcal R_t(c):\ r(c,\mathcal B_t(c))=0.
\label{eq:claim_drift_definition}
\end{equation}
The three relation types are \textbf{evidence-to-idea}, \textbf{idea-to-implementation} and \textbf{experiment-to-claim} (Table~\ref{tab:ai_scientist_claim_drift}). Relations are assessed against the corresponding design and claim versions, so a revision that consistently updates the attribution and validation requirements resolves the mismatch.

\begin{table}[!htbp]
\centering
\small
\renewcommand{\arraystretch}{1.15}
\caption{\textbf{Three support relations used to identify claim drift.}
A mismatch concerns a claim still presented as supported. Missing information without an established mismatch is recorded separately as an auditability gap.}
\label{tab:ai_scientist_claim_drift}
\begin{tabularx}{\linewidth}{@{}p{0.23\linewidth}YY@{}}
\toprule
\textbf{Type} & \textbf{Required correspondence} & \textbf{Example of a mismatch} \\
\midrule
evidence-to-idea & The attributed source, its findings and their conditions agree with the original evidence; new inference is distinguished from prior findings. & Evidence reported under a restricted condition is presented as established support for a broader design rationale. \\
\midrule
idea-to-implementation & The design to which a result is attributed is the design actually implemented and executed. & A replacement model is tested, but its result is attributed to an addition to the original model. \\
\midrule
experiment-to-claim & The claim's attribution, scope and certainty are supported by the completed experiments. & A whole-system gain is assigned to an unisolated component, or a conclusion conflicts with limiting results. \\
\bottomrule
\end{tabularx}
\end{table}

\Needspace{5\baselineskip}
An audit can confirm that a required relation holds, confirm that it fails, or lack the evidence needed to decide. We therefore distinguish \emph{supported}, \emph{drifted} and \emph{unverifiable} claims. A confirmed mismatch takes precedence over a missing relation. For system \(s\), task \(q\) and relation type \(k\), the audit claim drift rate (CDR) for a non-empty claim set is
\begin{equation}
\mathrm{CDR}_{\mathrm{audit}}^{s,q,k}
=\frac{N_{\mathrm{Drifted}}^{s,q,k}+N_{\mathrm{Unverifiable}}^{s,q,k}}
       {N_{\mathrm{Claims}}^{s,q,k}}.
\label{eq:cdr_audit}
\end{equation}
The denominator contains distinct accepted core claims requiring that relation. Under a fixed audit budget, the unverifiable fraction, \(N_{\mathrm{Unverifiable}}/N_{\mathrm{Claims}}\), measures how often the available evidence leaves a claim unresolved; a lower fraction indicates greater auditability. CDR records the combined fraction with a confirmed mismatch or unresolved support. We report the drifted and unverifiable counts separately (Table~\ref{tab:cdr_counts}). Hereafter, CDR refers to this audit measure.

\begin{table}[!h]
\centering
\small
\setlength{\tabcolsep}{4pt}
\renewcommand{\arraystretch}{1.18}
\caption{\textbf{Exact counts underlying the claim-level audit.} Each cell gives CDR (\%) followed by \((D,U,N)\), where \(D\) is the number of confirmed drifted claims, \(U\) the number of unverifiable claims and \(N\) the number of applicable core claims. Rates are computed as \((D+U)/N\) and rounded to one decimal. Categories can overlap; counts are not independent repeated research runs.}
\label{tab:cdr_counts}
\begin{tabularx}{\linewidth}{@{}l l *{3}{>{\centering\arraybackslash}X}@{}}
\toprule
\textbf{Task} & \textbf{System} & \makecell{\textbf{evidence-to-idea}} & \makecell{\textbf{idea-to-}\\\textbf{implementation}} & \makecell{\textbf{experiment-to-}\\\textbf{claim}} \\
\midrule
Memory & EvoScientist & \makecell{50.0\%\\(0,3,6)} & \makecell{88.9\%\\(0,8,9)} & \makecell{47.4\%\\(0,9,19)} \\
 & InternAgent-1.5 & \makecell{87.5\%\\(2,5,8)} & \makecell{100.0\%\\(0,16,16)} & \makecell{47.6\%\\(1,9,21)} \\
 & AI-Scientist-v2 & \makecell{50.0\%\\(0,2,4)} & \makecell{41.7\%\\(0,5,12)} & \makecell{10.0\%\\(1,1,20)} \\
\rowcolor{xcblue}
 & \textbf{\xcientist{}} & \makecell{11.6\%\\(1,4,43)} & \makecell{4.1\%\\(0,2,49)} & \makecell{5.8\%\\(0,4,69)} \\
\midrule
Graph-TS & EvoScientist & \makecell{57.1\%\\(1,3,7)} & \makecell{81.3\%\\(0,13,16)} & \makecell{11.1\%\\(0,2,18)} \\
 & InternAgent-1.5 & \makecell{66.7\%\\(2,2,6)} & \makecell{100.0\%\\(0,11,11)} & \makecell{38.1\%\\(0,8,21)} \\
 & AI-Scientist-v2 & \makecell{87.5\%\\(1,6,8)} & \makecell{100.0\%\\(1,8,9)} & \makecell{16.7\%\\(2,1,18)} \\
\rowcolor{xcblue}
 & \textbf{\xcientist{}} & \makecell{16.7\%\\(0,6,36)} & \makecell{8.6\%\\(0,3,35)} & \makecell{3.6\%\\(1,0,28)} \\
\midrule
PINN & EvoScientist & \makecell{88.9\%\\(0,8,9)} & \makecell{37.5\%\\(0,3,8)} & \makecell{36.8\%\\(0,7,19)} \\
 & InternAgent-1.5 & \makecell{66.7\%\\(0,2,3)} & \makecell{54.5\%\\(0,6,11)} & \makecell{50.0\%\\(1,6,14)} \\
 & AI-Scientist-v2 & \makecell{37.5\%\\(0,3,8)} & \makecell{77.3\%\\(1,16,22)} & \makecell{47.6\%\\(1,9,21)} \\
\rowcolor{xcblue}
 & \textbf{\xcientist{}} & \makecell{8.3\%\\(0,2,24)} & \makecell{7.9\%\\(1,2,38)} & \makecell{7.5\%\\(0,3,40)} \\
\bottomrule
\end{tabularx}
\end{table}

We evaluated retained records from \xcientist{}, EvoScientist~\cite{lyu2026evoscientist}, InternAgent-1.5~\cite{feng2026internagent} and AI-Scientist-v2 across three research tasks in two stages. The fixed-budget audit compared auditability using a common relation taxonomy applied to each system's own accepted claims. Expanded review then reassessed the initially unverifiable records to examine the reliability of their claim--evidence links (Section~\ref{sec:results_audit_expansion}). For each task, we pair the fixed-budget audit with the retained design trajectory, tracing how source evidence, implementation checks and experimental results informed revisions and the final claims.

\subsection{Task 1: Training-free memory system for LLM agents}
\begin{takeaway}
The final memory system increased overall F1 from 0.306 to 0.391 and reduced average token length by 64.2\%. Its audit CDR was 4.1\% for implementation identity and 5.8\% for experimental conclusions. Incomplete matched ablations limit attribution of the system-level gain to individual components.
\end{takeaway}

\label{sec:results_memory}

\begin{figure}[!h]
\centering
\includegraphics[width=\linewidth]{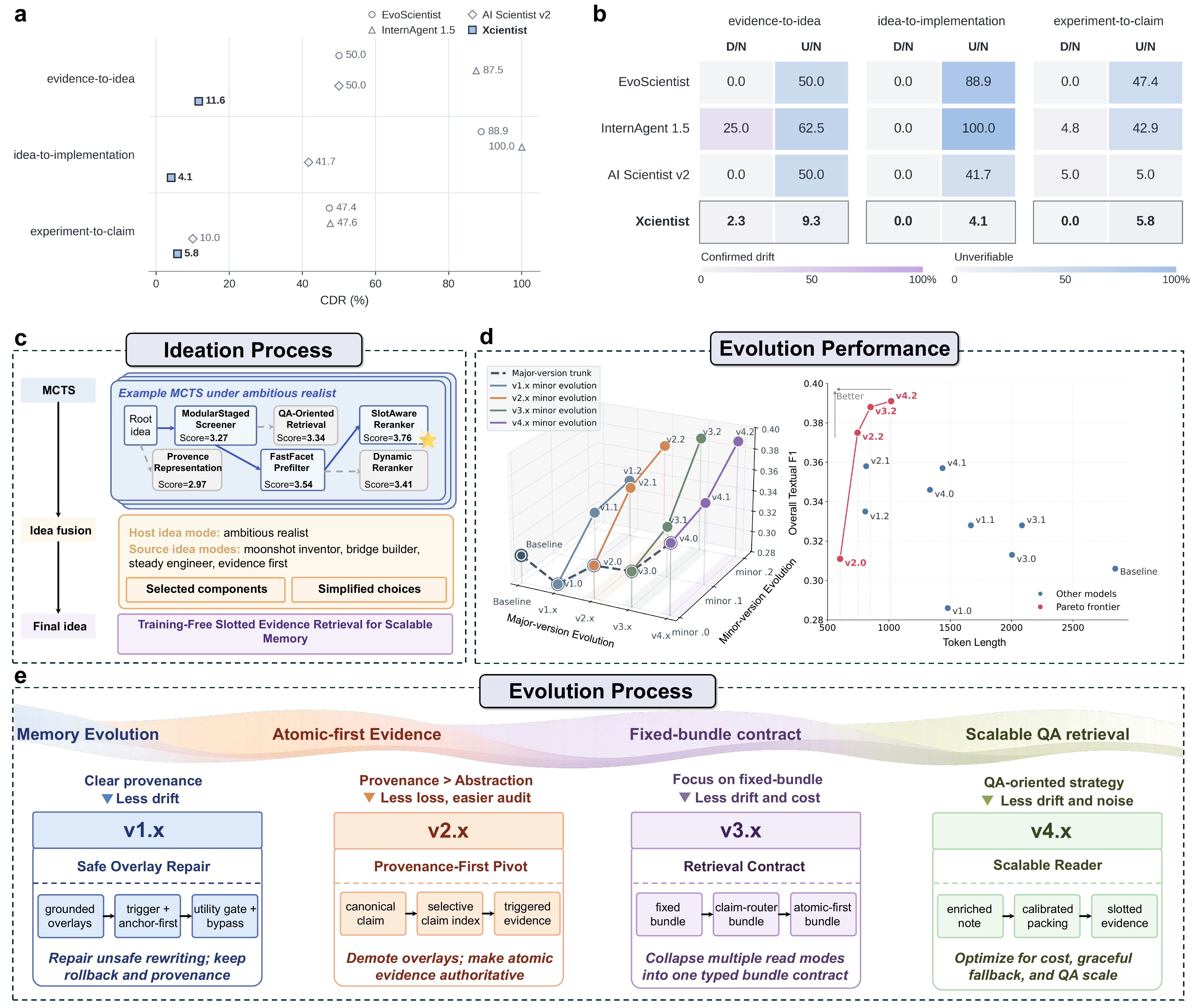}
\caption{\textbf{Training-free memory: audit and design history.} \textbf{a--b}, Audit CDR and its drifted/unverifiable split. \textbf{c}, MCTS and candidate fusion. \textbf{d}, F1 and token-length trade-offs. \textbf{e}, Four design families. Exact audit counts and version results appear in Tables~\ref{tab:cdr_counts} and \ref{tab:ed_detailed_case_results}a.}
\label{fig:memory_case}
\end{figure}

Under the same audit budget, \xcientist{}'s memory claims were easier to verify across revisions to storage and retrieval. CDRs were 11.6\% for evidence-to-idea, 4.1\% for idea-to-implementation and 5.8\% for experiment-to-claim. The corresponding ranges across EvoScientist, InternAgent-1.5 and AI-Scientist-v2 were 50.0--87.5\%, 41.7--100.0\% and 10.0--47.6\% (Fig.~\ref{fig:memory_case}a).

The audit linked 47 of 49 \xcientist{} implementation claims to execution records confirming that the code run matched the described method; two remained unverifiable. Among 43 claims about prior work, one did not match the cited evidence and four were unverifiable. Of 69 claims about experimental findings, four were unverifiable and none had a confirmed mismatch (Fig.~\ref{fig:memory_case}b; Table~\ref{tab:cdr_counts}).

Early versions, based on A-Mem\footnote{https://github.com/agiresearch/A-mem}~\cite{xu2026mem}, allowed stored text to be rewritten and restored. Later designs kept short notes linked to the original passages and changed how notes were selected for each question. The final design reserved space for relevant evidence, its source, timing and conflicting information (Fig.~\ref{fig:memory_case}c,e). The selection procedure could therefore change while the stored source notes remained available (Table~\ref{tab:experiment_contract_to_code}).

On the same LoCoMo subset~\cite{maharana2024evaluating}, the first version used fewer tokens but reduced answer quality. In the final selected version, v4.2, textual F1, which measures agreement with reference answers, rose from 0.306 for A-Mem to 0.391. Average token length per query fell from 2844.1 to 1017.2, a 64.2\% reduction (Fig.~\ref{fig:memory_case}d; Table~\ref{tab:ed_detailed_case_results}a). The complete system thus answered questions better with fewer tokens. Tests that removed individual components were unavailable for some components, limiting conclusions about which changes produced the gain (Table~\ref{tab:ed_claim_boundaries}).

\subsection{Task 2: Graph-structured spatio-temporal forecasting}
\begin{takeaway}
The retained records supported 32 of 35 implementation claims. Ablations provided little evidence for the hard projection and supported its replacement by innovation coverage. The final model reduced average MAE from 1.588 to 1.556 on PEMS-BAY; the conclusion remains specific to the evaluated protocol.
\end{takeaway}

\label{sec:results_graph}

\begin{figure}[!h]
\centering
\includegraphics[width=\linewidth]{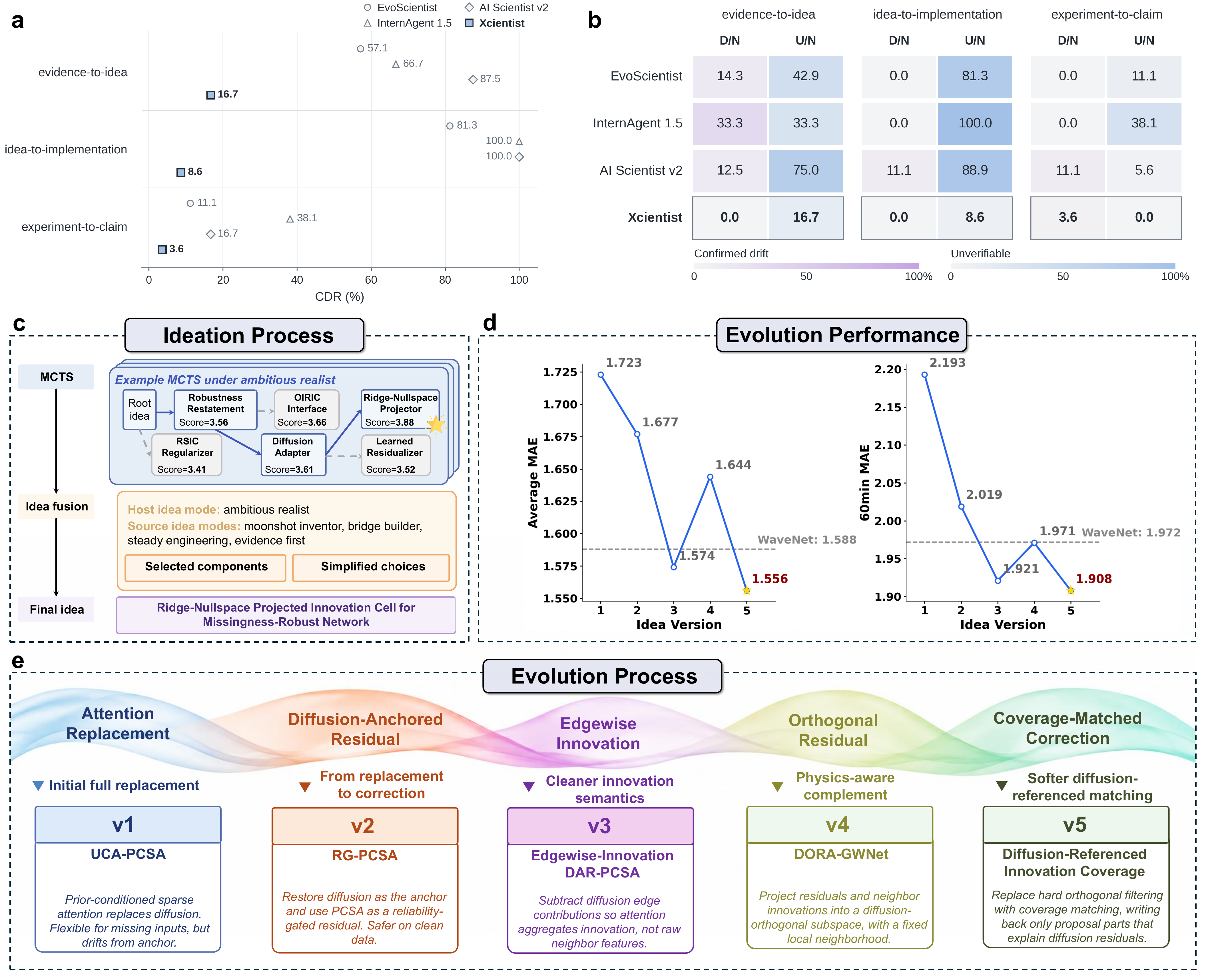}
\caption{\textbf{Graph forecasting: audit and design history.} \textbf{a--b}, Audit CDR and its drifted/unverifiable split. \textbf{c}, Search leading to the projection proposal. \textbf{d}, Average and 60-min MAE; dashed lines denote Graph WaveNet. \textbf{e}, Revisions ending in innovation coverage. Counts and version results appear in Tables~\ref{tab:cdr_counts} and \ref{tab:ed_detailed_case_results}b.}
\label{fig:graph_case}
\end{figure}

The fixed-budget audit verified a larger fraction of \xcientist{}'s implementation claims in traffic forecasting. Execution records supported 32 of 35 applicable claims, with three remaining unverifiable. Idea-to-implementation CDR was 8.6\%, compared with 81.3--100.0\% for the comparison systems (Fig.~\ref{fig:graph_case}a,b).

An early version shows why checking the executed method mattered. The task was to improve Graph WaveNet~\cite{wu2019graph} traffic forecasts on PEMS-BAY~\cite{li2017diffusion} by adding corrections to its predictions. The first candidate instead replaced Graph WaveNet and performed worse. Later versions restored it and changed only the added correction (Fig.~\ref{fig:graph_case}e).

The auditability advantage extended to literature attribution and experimental conclusions. Evidence-to-idea CDR was 16.7\%, compared with 57.1--87.5\% across the comparison systems; six of 36 \xcientist{} claims remained unverifiable and none had a confirmed mismatch. For experimental conclusions, the audit resolved all 28 \xcientist{} claims: 27 were supported and one conflicted with the recorded evidence. Experiment-to-claim CDR was 3.6\%, compared with 11.1--38.1\% for the comparison systems (Table~\ref{tab:cdr_counts}).

Separate component tests guided further revisions. Removing a filter intended to exclude information already captured by Graph WaveNet, termed hard orthogonal projection, changed clean-data mean absolute error (MAE) by only 0.0019. Under this protocol, the test gave little evidence of a benefit from the filter. Its replacement favoured corrections that added information the original predictor had missed, using proposal-space innovation coverage (Fig.~\ref{fig:graph_case}c,e). Removing coverage increased clean-data MAE by 0.0042 and worsened robustness in the block-40\% missing-input test (Table~\ref{tab:ed_claim_boundaries}). The final model reduced average MAE from 1.588 to 1.556 and 60-min MAE from 1.972 to 1.908 (Fig.~\ref{fig:graph_case}d; Table~\ref{tab:ed_detailed_case_results}b).

\subsection{Task 3: Multi-scale physics-informed neural networks}
\begin{takeaway}
The final PINN obtained the lowest reported mean errors on one-dimensional multiscale heat and PINNACLE heat, while external baselines remained better on two-dimensional heat. The experiment-to-claim audit supported 37 of 40 records and left three unverifiable, giving a CDR of 7.5\%.
\end{takeaway}

\label{sec:results_pinn}

\begin{figure}[!h]
\centering
\includegraphics[width=\linewidth]{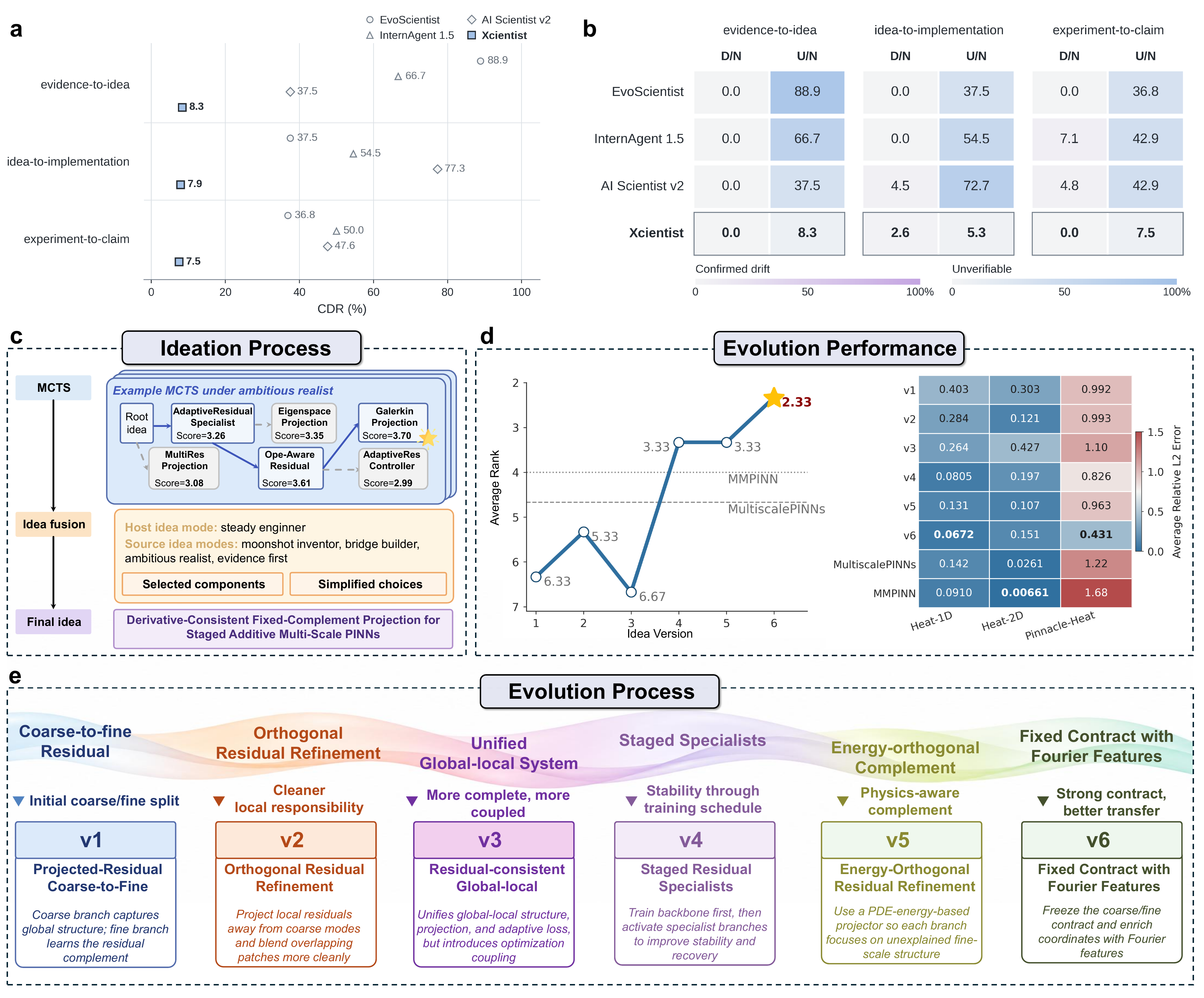}
\caption{\textbf{Multiscale PINNs: audit and design history.} \textbf{a--b}, Audit CDR and its drifted/unverifiable split. \textbf{c}, Search and fusion of the coarse--fine design. \textbf{d}, Average rank and benchmark errors. \textbf{e}, Six design versions. Counts and detailed results appear in Tables~\ref{tab:cdr_counts} and \ref{tab:ed_detailed_case_results}c.}
\label{fig:pinn_case}
\end{figure}

Claims from \xcientist{}'s physics-informed neural network (PINN) study were also more auditable under the fixed budget. The experiment-to-claim audit supported 37 of 40 claims, leaving three unverifiable and none with a confirmed mismatch. CDR was 7.5\%, compared with 36.8--50.0\% for the comparison systems (Fig.~\ref{fig:pinn_case}a,b). The retained conclusions distinguished lower mean errors on two heat-equation benchmarks from higher errors on the third.

For literature attribution, evidence-to-idea CDR was 8.3\% for \xcientist{}, compared with 37.5--88.9\% for the comparison systems; two of its 24 claims remained unverifiable. The audit also checked whether the described method matched the code actually run. Idea-to-implementation CDR was 7.9\%, compared with 37.5--77.3\%. Of 38 \xcientist{} claims, one did not match the implementation records and two remained unverifiable (Table~\ref{tab:cdr_counts}).

The models combined a broad solution to the heat equation with finer corrections. The run records described difficulty in training both parts together. Later versions learned the broad solution first and then trained the corrections. The final design kept the broad solution fixed and added Fourier features to represent variation over shorter scales (Fig.~\ref{fig:pinn_case}c,e). Across one-dimensional multiscale heat, two-dimensional multiscale heat and PINNACLE heat~\cite{hao2024pinnacle}, average rank improved from 6.33 for the first version to 2.33 for the sixth.

The individual results showed where the gains held. The final model had the lowest reported mean relative \(L_2\) errors on one-dimensional heat (0.067) and PINNACLE heat (0.431); lower error means a closer match to the reference solution. On two-dimensional heat, its error of 0.151 exceeded 0.007 for MMPINN and 0.026 for Multiscale PINNs~\cite{wang2024practical,wang2021eigenvector}, and also exceeded 0.107 for an earlier generated version (Fig.~\ref{fig:pinn_case}d; Table~\ref{tab:ed_detailed_case_results}c). These results supported an advantage on the first two benchmarks, while showing that the final model remained weaker on the third (Table~\ref{tab:ed_claim_boundaries}).

\subsection{Expanded audit and human review}

\label{sec:results_audit_expansion}

\begin{figure}[!h]
\centering
\includegraphics[width=\linewidth]{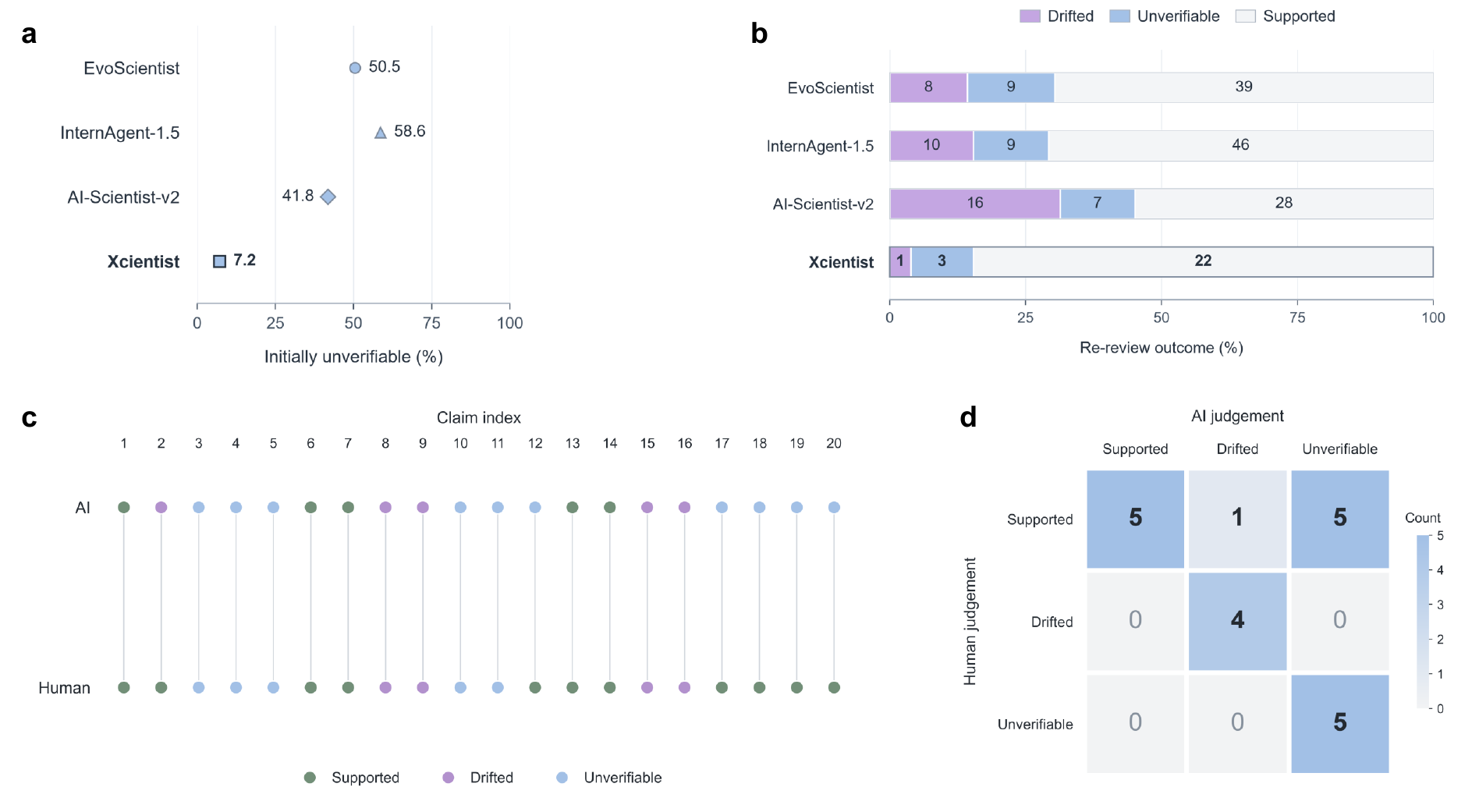}
\caption{\textbf{Expanded audit and human review.} \textbf{a}, Initially unverifiable fractions. \textbf{b}, Reassessment of 198 unresolved records; segment labels are counts. \textbf{c--d}, Paired labels and assessment matrix for 20 selected records. Human review retained 14 labels and could retrieve evidence beyond the automated bundle.}
\label{fig:claim-agreement}
\end{figure}

Across the three tasks and relation types, the fixed-budget audit left 7.2\% of \xcientist{} claim--relation records unverifiable, compared with 41.8--58.6\% for the comparison systems (Fig.~\ref{fig:claim-agreement}a). To assess claim reliability beyond the fixed budget, we used \texttt{gpt-6-astra} with web access to expand evidence retrieval and reassess all 198 initially unverifiable records.

Expanded review classified one of 26 initially unverifiable \xcientist{} records (3.8\%) as drifted, supported 22 and left three unresolved (Fig.~\ref{fig:claim-agreement}b). Across the comparison systems, 34 of 172 records (19.8\%) were classified as drifted, 113 became supported and 25 remained unverifiable. The fraction reclassified as drifted ranged from 14.3\% to 31.4\% across these systems. \xcientist{}'s greater auditability under the fixed budget was thus accompanied by a lower proportion of drift among the records reassessed during expanded review.

Human review of 20 selected records retained 14 automated labels (70.0\%; Fig.~\ref{fig:claim-agreement}c,d). Five unverifiable records became supported after execution manifests or completed-run metrics were recovered. One drifted record became supported after review clarified the scope of its literature attribution, and five records remained unverifiable. Reviewers could retrieve evidence beyond the automated bundle, so these changes reflected both evidence recovery and interpretation.

All 20 selected records are documented in Appendix~\ref{sec:supp-claim-cases}, with the claim, evidence and automated and human decisions. The human labels were not used to recalculate the fixed-budget CDRs. The audit compares selected records under different evidence access, not independent research runs or assessor agreement under identical information.

\begin{table}[!h]
\centering\small
\caption{\textbf{Expanded-review outcomes for initially unverifiable records.} The final three columns partition the initially unverifiable subset. Totals pool claim--relation records descriptively; a claim can occur in more than one relation category.}
\label{tab:expanded_audit_counts}
\begin{tabularx}{\linewidth}{@{}Yrrrrr@{}}
\toprule
System & \makecell{Initial\\records} & \makecell{Initially\\unverifiable} & Drifted & Unverifiable & Supported\\
\midrule
EvoScientist & 111 & 56 & 8 & 9 & 39\\
InternAgent-1.5 & 111 & 65 & 10 & 9 & 46\\
AI-Scientist-v2 & 122 & 51 & 16 & 7 & 28\\
\rowcolor{xcblue}\xcientist{} & 362 & 26 & 1 & 3 & 22\\
\bottomrule
\end{tabularx}
\end{table}

\section{Discussion}
\label{sec:discussion}

Across three computational studies, \xcientist{} had lower audit CDR than each comparison system in all nine task-by-relation comparisons, with rates ranging from 3.6\% to 16.7\%. Most of this difference arose from fewer unverifiable claims. The main observed advantage was therefore that the retained records more often allowed readers to recover and check the support for a claim. Confirmed-drift rates varied across systems and settings. These findings place the emphasis on maintaining access to the evidence needed to assess a changing study, while leaving the correctness of each claim subject to examination.

Recent systems have extended the automation of idea generation, implementation and experimentation~\cite{yamada2025aiscientistv2,lyu2026evoscientist,feng2026internagent}. As a proposal changes during these activities, its rationale and evaluation must remain connected to the relevant prior evidence. The shared evidence graph in \xcientist{} provides this connection by supplying source passages, methodological limitations, baselines and experimental conditions to both idea generation and validation. A source used to motivate a component can thus be revisited when deciding how to implement and test it. Retaining the original passage and its conditions also allows an auditor to distinguish a finding established by prior work from an inference introduced by the research agent. This distinction matters when a plausible design rationale extends beyond what the cited study actually demonstrated.

The memory study illustrates how component-based idea generation makes revisions easier to interpret. Its design evolved from reversible rewriting towards immutable source notes and question-dependent evidence selection. Representing storage, enrichment and retrieval as separate components allowed the selection procedure to change while the underlying source notes remained available. More generally, each component carries its intended role, source attribution and proposed test through search, fusion and implementation, so an edit can be examined against the assumption it changes. The resulting memory system improved answer quality and token efficiency, but incomplete component tests limited attribution of those gains. The component records made this gap explicit even where the complete system performed better.

Validation contracts connect these proposed component roles to executable comparisons. They specify the implementation, baseline, controlled component changes and outputs required for a claim, while separate validators inspect the artifacts produced by workers. Their role becomes particularly clear when experimental evidence challenges a selected idea. In forecasting, the proposed hard orthogonal projection had little observed benefit under the tested protocol, prompting a revision towards proposal-space innovation coverage while retaining the diffusion backbone. The contract and component mapping provided a way to test the revised operation without losing track of the intended correction path. In the PINN study, staged coarse--fine learning and the final fixed-complement design with Fourier features yielded lower mean errors on two benchmarks, while remaining weaker than external baselines on the two-dimensional multiscale task. These results required benchmark-specific conclusions despite an improved average rank. Validation therefore affected both the design pursued and the scope of the claim that could be retained.

The present evaluation does not isolate the contribution of the evidence graph, component representation or validation contracts to the observed CDR differences. The retrospective audit compared each system's own accepted claims, which differed in content and applicability, and depended on the records available for inspection. The findings characterize these records across three computational settings; they do not estimate claim-drift incidence across independent research runs. The workflow also remains dependent on retrieval coverage, the adequacy of the selected comparisons and validator judgement. Establishing how much each module contributes would require controlled module comparisons across repeated runs, together with comparable access to the evidence used in each audit.

Transparent reporting requires records that identify the tests, component revisions and evidence used to delimit a conclusion~\cite{vandis2023priorities,hanson2023garbage,kapoor2024reforms}. In \xcientist{}, the evidence graph, component histories and validation records serve these decisions and remain available for subsequent review. They allow a reader to follow a proposed idea through its revisions, identify the implementation behind a result and determine which completed comparisons support the final claim.
\section{Methods}
\label{sec:methods}

This section summarizes the implementation and specifies the audit and task protocols. Appendix~\ref{app:prompt_contracts} reproduces the prompt contracts.
\begin{figure}[!htbp]
    \centering
    \includegraphics[width=\linewidth]{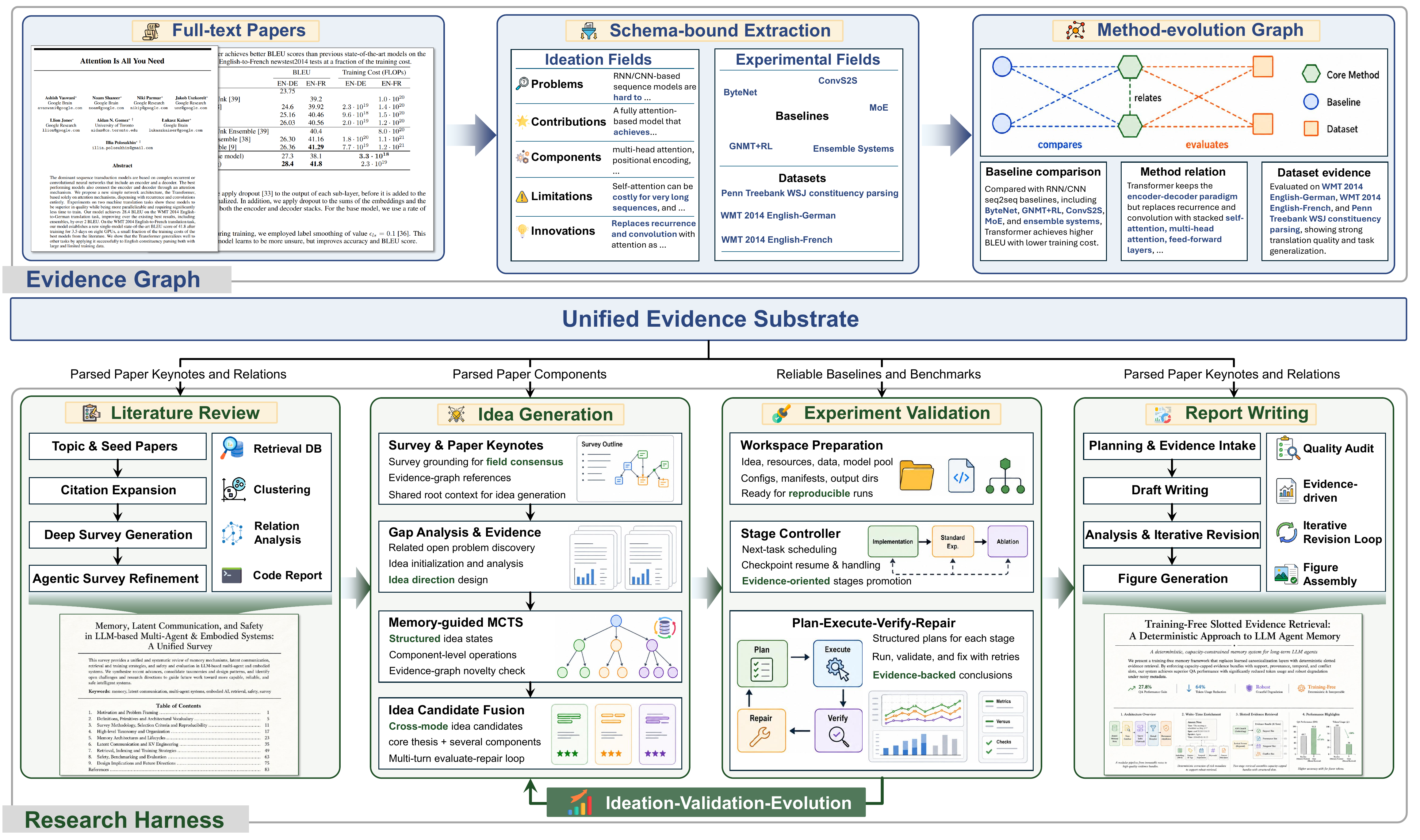}
    \caption{\textbf{Framework design of \xcientist{}.} The evidence graph supplies source records to idea generation and experiment validation. The harness maps selected components to code, checks declared comparisons and records repairs and unresolved requirements.}
    \label{fig:methods}
\end{figure}

\subsection{Evidence Graph} \label{app:evidence_graph}
The evidence graph records method-level information needed for idea generation and experiment planning: components, baselines, datasets, evaluation conditions and results. Citation links and abstract retrieval provide useful context but often omit these details.

\begin{figure}[htbp]
    \centering
    \includegraphics[width=\linewidth]{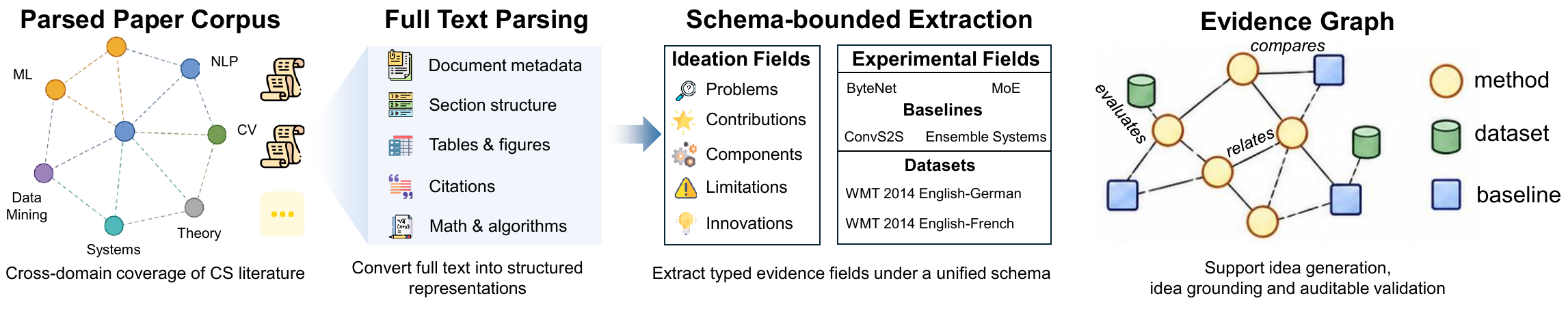}
    \caption{Construction of the heterogeneous method-evolution evidence graph. (1) Full-text Parsing, (2) Schema-bound Extraction showing the two-block schema, (3) Graph Construction with node types and edge types labeled.}
    \label{fig:graph_construction}
\end{figure}

Each paper is decomposed into core method entities linked to shared baseline and dataset nodes. Core entities retain components, limitations and future directions. Construction proceeds through full-text parsing, schema-bound extraction and graph assembly with entity resolution (Figure~\ref{fig:graph_construction}).

\subsubsection{Corpus Construction}

Our corpus is drawn from computer science papers indexed through Semantic
Scholar~\cite{kinney2023semantic}. Papers span multiple CS subfields;
the largest domain groups are cs.CL, cs.CV, cs.LG, and cs.AI, with a smaller
tail from neighboring areas. The full graph is constructed from nearly 50K source
papers after filtering.

\subsubsection{Full-text Structured Parsing}

A central design decision is to parse full paper text rather than abstracts
alone. Abstract-level content omits critical signals: baseline comparison
details, component-level descriptions, experimental conditions, and failure
modes are rarely recoverable from abstracts but are essential for method-level
evidence extraction. We parse each paper using
MinerU~\cite{wang2024mineru}, which merges the rendered Markdown text with
structured content lists (such as tabular data) into a unified representation.
This ensures that empirical evidence from tables is preserved and directly
available to the extraction stage.

Paper-level metadata include year, venue, title, domain label, a structured summary of the background, method and results, and a code-repository URL when available. These records are shared by all core entities extracted from the paper.

\subsubsection{Schema-bound Evidence Extraction}

From each parsed paper, we apply schema-bound extraction with constrained
structured outputs to obtain a set of typed evidence fields. The schema is
organized into two blocks. The first block, \texttt{ideation\_\allowbreak resource},
captures the scientific content most relevant to downstream ideation and
analysis: identified problems, core contributions, components, innovations,
limitations, and future work directions. The second block,
\texttt{graph\_\allowbreak data}, captures the experimental comparison structure:
baseline methods and benchmark datasets mentioned in the context of
evaluation.

\begin{table}[htbp]
    \centering
    \small
    \caption{Structured evidence schema used for per-paper extraction. Textual evidence retains the four fields \texttt{(keywords, summary, insight, quote)} for source checking.}
    \label{tab:schema_overview}
    \begin{tabularx}{\linewidth}{@{}>{\raggedright\arraybackslash}p{0.19\linewidth}>{\raggedright\arraybackslash}p{0.23\linewidth}YY@{}}
        \toprule
        \textbf{Block} & \textbf{Field} & \textbf{Description} & \textbf{Base Attribute Format} \\
        \midrule
        \texttt{ideation\_\allowbreak resource} 
        & \texttt{problems} & Identified research gaps \& motivations & \texttt{(keywords, summary, insight, quote)} \\
        & \texttt{core\_\allowbreak contributions} & Proposed methods and core insights & \texttt{(name, type, keywords, ...)} \\
        & \texttt{components} & Reusable technical building blocks & \texttt{(name, keywords, summary, ...)} \\
        & \texttt{innovations} & Novel approaches and improvements & \texttt{(keywords, summary, insight, quote)} \\
        & \texttt{limitations} & Stated weaknesses and failure modes & \texttt{(keywords, summary, insight, quote)} \\
        & \texttt{future\_\allowbreak work} & Suggested directions and next steps & \texttt{(keywords, summary, insight, quote)} \\
        & \texttt{core\_\allowbreak relations} & Methodological edges between cores & \texttt{(source, target, relation, ...)} \\
        \midrule
        \texttt{graph\_\allowbreak data} 
        & \texttt{baselines} & Compared baseline methods & \texttt{(name, metrics, keywords, ...)} \\
        & \texttt{datasets} & Benchmarks and evaluation datasets & \texttt{(name, metrics, keywords, ...)} \\
        \bottomrule
    \end{tabularx}
\end{table}

The fields accept natural-language descriptions, including relation descriptions, so the schema can represent methods, systems, empirical studies and theory papers. Each record is as detailed as the source evidence permits.

Each field is represented as a four-tuple \texttt{(keywords, summary, insight,
quote)}, where \texttt{keywords} supports structured retrieval,
\texttt{summary} provides a concise description, \texttt{insight} encodes an
interpretive observation about the field's significance, and \texttt{quote}
anchors the record to a verbatim passage from the source paper, enabling
downstream evidence tracing and consistency verification.

Extraction is performed in three passes. The first pass extracts all
\texttt{ideation\_\allowbreak resource} fields from the parsed full text. The second pass
extracts \texttt{graph\_\allowbreak data} fields---specifically baseline and dataset
mentions---in a dedicated call; separating this pass improves performance, as
identifying baseline comparisons requires focused attention to experimental
context that competes with the broader ideation fields when handled jointly.
The third pass grounds the extracted baseline and dataset entities against the
paper's actual reference list retrieved from Semantic Scholar, correcting
surface-form mismatches and attaching canonical metadata---including Paper ID,
canonical title, publication venue, and citation information---to each resolved
entity. This metadata serves both as a reliability signal for downstream
filtering and as the basis for cross-paper entity merging in the graph
construction stage.

\subsubsection{Graph Construction with Entity Resolution}

The extracted evidence records are assembled into a heterogeneous graph
$\mathcal{G} = (\mathcal{V}, \mathcal{E})$ with three node types,
\begin{equation}
    \mathcal{V} = \mathcal{V}_{\mathrm{core}} \cup
    \mathcal{V}_{\mathrm{base}} \cup \mathcal{V}_{\mathrm{data}},
\end{equation}
where $\mathcal{V}_{\mathrm{core}}$ contains core method entities,
$\mathcal{V}_{\mathrm{base}}$ contains baseline method entities, and
$\mathcal{V}_{\mathrm{data}}$ contains dataset and benchmark entities. A single
paper may decompose into multiple core nodes, each corresponding to a distinct
contribution; paper-level metadata (title, domain, venue, publication year) is
shared across all core nodes originating from the same paper.

The edge set is typed as
\begin{equation}
    \mathcal{E} = \mathcal{E}_{\mathrm{core}} \cup
    \mathcal{E}_{\mathrm{cmp}} \cup \mathcal{E}_{\mathrm{eval}},
\end{equation}
where $\mathcal{E}_{\mathrm{core}}$ denotes methodological relations between
core entities, $\mathcal{E}_{\mathrm{cmp}}$ denotes baseline comparison edges,
and $\mathcal{E}_{\mathrm{eval}}$ denotes evaluation edges linking core entities
to datasets. All edge relation fields are expressed as natural language,
consistent with the open-field design of the node schema. Comparison and
evaluation edges retain local experimental evidence as edge attributes,
including metrics, comparison context, summary, and supporting quotes.

Baseline and dataset nodes are maintained as globally unique entities across
the full corpus. When the same method or dataset appears across multiple papers,
the corresponding nodes are merged rather than duplicated, and all associated
comparison and evaluation edges are consolidated onto the single canonical node.
Merging is performed using the Paper IDs obtained during the third extraction
pass: two entity mentions are unified if and only if they resolve to the same
Semantic Scholar Paper ID, regardless of how they were named in the source
papers. This ensures that all experimental comparisons involving the same method 
are queryable through a single node.

After construction and merging, the graph contains approximately 72K core nodes,
250K baseline nodes, and 63K dataset nodes, for a total of 385K nodes and 1.15M
typed edges.

\begin{table}[htbp]
    \centering
    \caption{Summary statistics of the heterogeneous method-evolution evidence graph. Node counts reflect globally deduplicated entities after entity resolution.}
    \label{tab:graph_stats}
    \begin{tabular}{@{}lr@{}}
        \toprule
        \textbf{Metric} & \textbf{Count} \\
        \midrule
        Source papers & $\sim$50,000 \\
        Core method nodes ($\mathcal{V}_{\mathrm{core}}$) & 72,000 \\ 
        Baseline nodes ($\mathcal{V}_{\mathrm{base}}$) & 250,000 \\
        Dataset nodes ($\mathcal{V}_{\mathrm{data}}$) & 63,000 \\
        \midrule
        \textbf{Total nodes} ($\mathcal{V}$) & \textbf{$\sim$385,000} \\
        \textbf{Total typed edges} ($\mathcal{E}$) & \textbf{$\sim$1,150,000} \\
        \bottomrule
    \end{tabular}
\end{table}

\subsubsection{Integration with Downstream Modules}

The graph supplies shared evidence to literature review, idea generation and experiment validation.

For \textbf{research synthesis}, methodological edges trace relations between methods and experimental findings. Innovation, limitation and future-work fields suggest candidate questions; component records identify reusable or editable operations. Source passages allow downstream analyses and proposals to be checked against the cited work.

For \textbf{experiment validation}, baseline comparisons and dataset links identify starting implementations, metrics and evaluation conditions. These records guide controlled comparisons and component tests. The retained passages also preserve the conditions under which earlier findings were established, helping distinguish a supported result from a proposed extension.

\subsection{Literature Review}
\subsubsection{Overview}

\textsc{Xcientist} uses DeepSurvey~\cite{yang2026deepsurvey} for literature review. The module retrieves evidence from the evidence graph, analyzes the paper set and drafts a survey whose source records can be reused by idea generation and reporting.

Analysis precedes drafting: keynotes, relation graphs and comparison tables are prepared before the survey text. These outputs retain source-paper and passage links for subsequent checks.

\subsubsection{System Architecture}

\begin{figure}[t]
  \centering
  \includegraphics[width=0.8\linewidth]{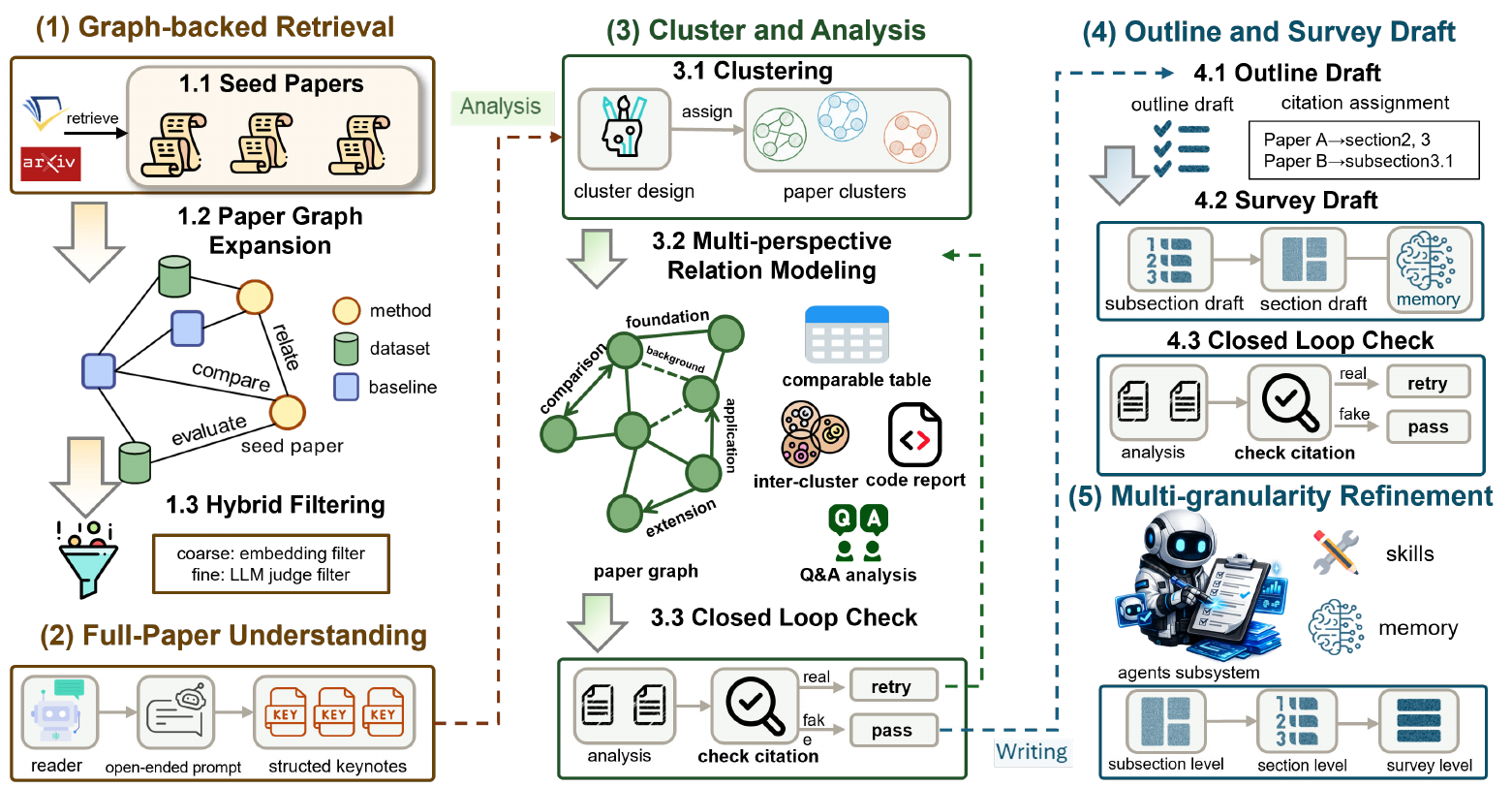}
  \caption{\textbf{Overview of DeepSurvey}. 
  Stage 1 performs graph-backed retrieval with hybrid filtering to collect evidence papers. Stage 2 extracts structured keynotes from full papers. Stage 3 clusters papers and conducts multi-perspective relation modeling (comparable tables, evidence graph, inter-cluster Q\&A) and code repository analysis. Stage 4 drafts an outline with citation assignment, then generates the survey subsections in parallel. Stage 5 applies multi-granularity agentic refinement (section, subsection, survey levels) to ensure reliability and citation correctness.}
  \label{fig:DeepSurvey}
\end{figure}

DeepSurvey follows a staged pipeline organized into five principal stages, as Figure~\ref{fig:DeepSurvey} shows. The first three stages build the analysis substrate; the final two stages generate and refine the survey document.

\subsubsection{Graph-backed Literature Retrieval}

Given a research topic, DeepSurvey retrieves seed papers via the Evidence Graph mentioned in Section~\ref{app:evidence_graph} with Semantic Scholar API as fallback and expands along citation and reference edges in both directions to a bounded depth. This graph-based expansion captures inter-paper dependencies that flat semantic search misses, producing a structurally connected evidence collection.

A title/abstract similarity filter removes weak candidates at low cost. An LLM then checks relevance against the topic and seed context. The retained set supplies the full-text analysis stages.

\subsubsection{Full-text Keynotes}

For each retained paper, DeepSurvey generates a structured keynote from the full text rather than from the abstract alone. This design choice is motivated by the observation that key technological details---implementation mechanisms, experimental configurations, applicability conditions, and failure modes---are often absent from or heavily condensed in abstracts, yet are essential for meaningful cross-paper comparison and mechanism-level reasoning.

Keynotes adopt an open-ended schema that adapts to different contribution types (method, system, or theory papers) without imposing rigid categorical templates. Each keynote records contributions, methods, experiments, assumptions, and limitations in a domain-adaptive format. The resulting keynotes preserve the contribution-specific details used in later comparisons.

\subsubsection{Clustering and Multi-perspective Relation Analysis}

After keynote generation, DeepSurvey organizes papers into thematic clusters. A clustering agent designs cluster themes from the complete keynote collection, and a partitioning agent assigns papers accordingly. Papers spanning multiple topics may be assigned to multiple clusters, reflecting the cross-topic nature of complex contributions.

Within each cluster, three complementary analytical perspectives are applied:

\begin{itemize}[nosep]
    \item \textbf{Relation graphs} model typed citation relationships (e.g., foundation, extension, substitution, complementarity) with abstracted descriptions, capturing technical lineage and method evolution trajectories.
    \item \textbf{Comparison tables} align papers along key technical dimensions---method objectives, core mechanisms, experimental setups, and limitations---enabling systematic horizontal comparison within a research theme.
    \item \textbf{Guided Q\&A synthesis} integrates cluster evidence to answer high-value research questions, producing compact analytical results that support cross-paper pattern extraction and gap identification.
\end{itemize}

A code-analysis agentic subsystem extends the textual analysis to source-code repositories, retrieving implementation-level details and exposing technological dimensions overlooked by paper prose alone. Inter-cluster analysis is also performed to capture cross-cluster differences and broader research landscape patterns. All analytical outputs require explicit source attribution, with checks for missing or unsupported source attribution.

\subsubsection{Outline-driven Drafting and Citation Enforcement}

DeepSurvey generates a hierarchical outline from the clustering results and iteratively refines it against retrieved paper keynotes. Papers are then explicitly assigned to appropriate sections based on their keynotes and cluster analyses. This citation-anchoring process creates a localized evidence constraint: each writing unit depends only on its assigned paper subset, reducing hallucination risk by limiting the context available to the generation model.

Writing proceeds bottom-up: subsection drafts receive the outline, relevant cluster analyses, and assigned paper keynotes; section drafts aggregate subsection content with overview-level analysis. After each paragraph, a citation verification mechanism checks consistency and authenticity against the assigned paper set. Detected errors trigger retry generation with error memory, enforcing strict alignment between generated text and the evidence collection.

\subsubsection{Multi-granularity Agentic Refinement}

After initial draft generation, DeepSurvey applies iterative refinement at three granularities: section-level (organization and logic), subsection-level (local analysis, readability, and citation-claim validation), and survey-level (cross-section coherence). A centralized planning agent inspects the draft, outline, revision history, and relevant evidence, then coordinates specialized roles---keynote reader, reviewer, and reviser---through an explicit action plan. Global memory preserves reviewer feedback, score changes, and revision states across rounds, ensuring cross-round consistency. A simplified skill-loop fallback is available for weaker backbone models.

\subsection{Idea Generation}
Idea generation searches editable component descriptions under several Idea Taste Modes. Each mode starts from the same topic, retrieved evidence and prepared root idea, but changes the preferences used to propose and evaluate edits. Modes include exploratory transfer, conservative engineering, ambitious but implementable design and evidence-first reasoning.

\begin{table}[h]
\centering
\caption{Stages of the \textsc{Xcientist} ideation pipeline and their roles.}
\label{tab:stage_of_ideation_pipeline}
\begin{tabular}{@{}p{0.18\linewidth} p{0.36\linewidth} p{0.36\linewidth}@{}}
\toprule
\textbf{Stage} & \textbf{Definition} & \textbf{Role in Ideation} \\
\midrule
Knowledge Acquisition & Retrieve survey-grounded and graph-backed references related to the topic. & Build the evidence base for ideation and reduce problem drift. \\
Advanced Analysis & Extract mechanisms, limitations, open problems, and promising directions from the retrieved context. & Convert raw literature context into structured design signals. \\
Re-analysis \& Replanning & Optionally revise the focus, mature idea, or retrieval direction when prior context or partial ideas already exist. & Keep the search aligned with the most relevant problem framing. \\
Idea Generation & Run structured idea search to produce candidate ideas and finalize a high-quality proposal. & Turn grounded analysis into concrete, researchable ideas. \\
\bottomrule
\end{tabular}
\end{table}

\subsubsection{Workflow Stages}
The workflow begins with evidence acquisition and analysis, optionally revises an existing idea or retrieval plan, and then searches and combines candidate designs (Table~\ref{tab:stage_of_ideation_pipeline}).

\subsubsection{Idea Taste Modes}
Idea Taste Modes change the relative emphasis on novelty, feasibility, mechanism grounding and experimental verifiability while keeping the research problem fixed.

\begin{figure}[h] 
    \centering
    \includegraphics[width=\linewidth]{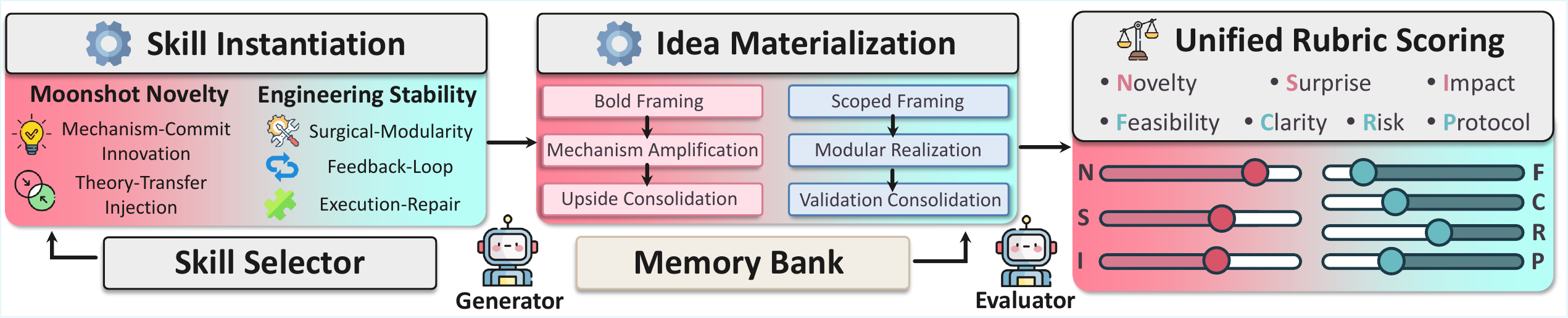}
    \caption{Idea Taste Modes as structured generation and evaluation priors over the same grounded problem.}
    \label{fig:idea_taste_mode}
\end{figure}

Figure~\ref{fig:idea_taste_mode} summarizes these preferences. Some modes favor high-risk proposals or cross-domain transfer; others favor modular implementation or direct ablation. Search returns candidates developed under each preference for later comparison and fusion.

\subsubsection{Memory-Guided MCTS}
We model idea generation as a Monte Carlo Tree Search (MCTS) procedure over structured idea states. Starting from a root idea, \textsc{Xcientist} expands child candidates, evaluates them through a referee-style simulation step, and backpropagates the resulting reward along the search path to guide future selection. The Select and Backpropagate phases follow standard MCTS, so the key system-specific design lies in the Expand and Simulate operations.

\paragraph{Expand.} 
Unlike prior work that treats ideation as largely free-form generation \cite{lyu2026evoscientist,feng2026internagent}, \textsc{Xcientist} uses a structured action space during expansion. It treats each idea as a set of editable components and applies component-level operators such as adding, removing, replacing, or rewiring modules, as well as introducing validation protocols. These operations are composed into a higher-level edit skill $S$, allowing \textsc{Xcientist} to perform targeted architectural revisions while keeping the search space interpretable and experimentally grounded.

To make expansion adaptive to past search experience, we further introduce a vector-memory mechanism. When \textsc{Xcientist} receives feedback during simulation, it stores that experience and retrieves relevant signals in later expansions. The retrieved memory provides local edit patterns, failure corrections, or validation hints observed in other search trajectories, thereby guiding the generation of new child ideas:
\begin{equation}
    I^i_l = \mathrm{Expand}(S^i_l, I^j_{l-1}, M_v \mid \mathrm{taste}),
\end{equation}
where $I^i_{l}$ denotes the $i$-th idea at layer $l$, $S^i_{l}$ denotes the instantiated edit skill for that idea, and $M_{v}$ denotes the retrieved vector memory. In effect, \textsc{Xcientist} implements an implicit form of cross-over: instead of explicitly merging two parent ideas, it reuses and recombines useful local patterns from other branches. Cross-domain transfer is handled within this expansion stage by retrieving mechanisms from other fields only when they address a diagnosed weakness in the current idea and can be adapted back into the target domain.

\paragraph{Simulate.} 
For each candidate idea, \textsc{Xcientist} builds an evaluation prompt from the topic, fixed root-domain constraints, the root-idea anchor, the candidate method description, and retrospective symbolic-memory hints. The evaluator then produces a multi-dimensional judgment over novelty, surprise, impact, feasibility, clarity, alignment, complexity, risk, and protocol quality. These signals are aggregated into a composite score used for MCTS backpropagation and search control. We use symbolic memory here as a grounded hint because it is derived from component-wise ablation results obtained during experimental validation, which allows \textsc{Xcientist} to reason about the functional role each component played in prior ideas:
\begin{equation}
    \{r^i_l, f^i_l\} = \mathrm{Simulate}(I_l^i, I_0, \mathcal{T}, M_s \mid \mathrm{taste}),
\end{equation}
where $r^i_l$ and $f^i_l$ denote the reward and verbal feedback for the $i$-th idea at layer $l$, $\mathcal{T}$ denotes the topic, and $M_s$ denotes the retrieved symbolic memory. Novelty is assessed in a reference-grounded and mechanism-centered way: rather than relying on surface-level difference, the evaluator compares the candidate against nearby prior work in the evidence graph and assesses the proposed component change.

\subsubsection{Idea Fusion}
Fusion selects a host candidate that defines the central research question and evaluates components from other candidates against that question. It retains components that address a recorded weakness or improve the validation plan and records why other components were rejected.

Formally, let $\mathcal{M}$ denote the set of idea taste modes, and let $\hat{I}_m$ be the best candidate returned by mode $m$:
\begin{equation}
    \hat{I}_m = \arg\max_{I \in \mathcal{S}_m} \mathrm{Score}(I \mid m),
\end{equation}
where $\mathcal{S}_m$ is the search tree explored under mode $m$. Because all $\hat{I}_m$ are generated from the same prepared root context, they are diverse in search preference but comparable in problem framing. \textsc{Xcientist} therefore performs fusion over $\{\hat{I}_m\}_{m \in \mathcal{M}}$ while conditioning on the shared topic, root-domain constraints, the current mature idea, the optional refinement scope, and the structured analysis summary.

\paragraph{Fusion draft.}
Fusion is formulated as a constrained composition problem rather than a symmetric merge. \textsc{Xcientist} first selects one host idea whose central mechanism defines the fused thesis, and then evaluates each external component according to whether it can support that thesis as a support module, protocol, or guardrail. The fused result is therefore required to expose exactly one dominant core mechanism; additional retained components are allowed only when they sharpen the causal story, strengthen validation, or resolve a concrete weakness without introducing a second competing novelty. It also preserves scope: when a refinement scope is provided, candidates are not allowed to appear stronger merely because they move the contribution to a different subsystem.

The output of the first fusion pass is an initial fused draft together with explicit synthesis metadata:
\begin{equation}
    I_{\mathrm{fuse}}^{(0)}, \Omega = \mathrm{Fuse}\big(\{\hat{I}_m\}_{m \in \mathcal{M}}, \mathcal{C}_{\mathrm{shared}}\big),
\end{equation}
where $I_{\mathrm{fuse}}^{(0)}$ is the initial fused idea and $\Omega$ records the host mode, selected components, rejected components, conflict-resolution decisions, the fused core thesis, and a minimal validation plan. Retaining this metadata makes the synthesis stage inspectable by exposing why specific components were preserved, simplified, or discarded.

\paragraph{Post-fusion repair.}
A first-pass fusion may still contain local redundancy or unresolved incompatibilities, so \textsc{Xcientist} does not accept the fused draft directly. Instead, it sends the fused candidate back to the evaluator under a fixed referee mode and runs a local repair loop:
\begin{equation}
    I_{\mathrm{fuse}}^{(t+1)} = \mathrm{Repair}\big(I_{\mathrm{fuse}}^{(t)}, f_{\mathrm{ref}}^{(t)}, \{\hat{I}_m\}_{m \in \mathcal{M}}\big),
\end{equation}
where $f_{\mathrm{ref}}^{(t)}$ denotes the evaluator feedback for the current fused draft. A repaired draft is accepted only if it improves the composite evaluation score over the previous fused idea; otherwise, the old version is kept. The final output retains the selected component list, evaluation feedback and accepted repair history.

\subsection{Experiment Validation}
The experiment module translates a structured proposal into runnable code, standard comparisons and component tests. Every stage records a contract, execution artifacts and a validator decision. A textual progress summary alone cannot satisfy a completion condition.

The implementation separates planning, execution and verification. Workers perform a stage's operations, while validators judge the resulting artifacts against its contract. Each experiment runs in a self-contained project with versioned paths, permissions and configuration.

A master scheduler checks workspace state at the start of each iteration. In priority order, it checks code self-containment, implementation validation, standard experiment outputs and component-test coverage. A missing prerequisite routes the run back to the relevant phase. Full convergence requires validator pass states, no unresolved blockers and the required component-to-evidence mapping (Table~\ref{tab:experiment_phases}).

\begin{table}[h]
\centering
\caption{Principal phases of the experiment validation pipeline and their validation roles.}
\label{tab:experiment_phases}
\begin{tabular}{@{}p{0.18\linewidth} p{0.38\linewidth} p{0.34\linewidth}@{}}
\toprule
\textbf{Phase} & \textbf{Objective} & \textbf{Validator Mandate} \\
\midrule
Prepare 
& Acquire repositories, configure the environment, stage datasets and models, and synthesize handoff documents. 
& Each of the five sequential stages must receive a stage-specific PASS before synthesis. \\

Code 
& Implement the full idea in \texttt{project/} as self-contained runnable code with ablation hooks. 
& Every plan step, including the mandatory \texttt{final\_\allowbreak integration\_\allowbreak smoke}, must pass contract validation. \\

Standard Science 
& Run baseline and full-method experiments on prepared targets to collect primary metrics. 
& All experiment steps must produce real execution artifacts under \texttt{standard\_\allowbreak results\_\allowbreak dir}. \\

Ablation Science 
& Systematically disable each canonical component and measure its marginal contribution. 
& The number of ablation steps must exactly match the number of canonical components, and each must receive PASS. \\
\bottomrule
\end{tabular}
\end{table}

\subsubsection{Preparation: Building the Execution Surface}
The preparation phase establishes the material foundation upon which all later phases depend. It proceeds through five sequential stages---repository acquisition, environment construction, dataset staging, model staging, and synthesis---each governed by an explicit contract that specifies goals, input paths, permitted write roots, required outputs, and a done condition. A step executor mediates a repair loop between a worker, which performs concrete operations such as file editing and command execution, and a validator, which checks whether the step output satisfies the contract. Only when the validator returns pass does the executor advance to the next stage; otherwise, the feedback is routed back to the worker for repair until either the contract is satisfied or a repair round limit is reached.

Preparation ends with two handoff files. \texttt{prepare\_\allowbreak target\_\allowbreak inventory.json} records verified repository paths, dataset files, model identifiers, environment variables and benchmark entry points. \texttt{prepare\_\allowbreak idea.md} records the research objective, canonical component list, implementation guidance and resource provenance. Subsequent stages use these files as their declared execution and design specification.

\begin{figure}
    \centering
    \includegraphics[width=\linewidth]{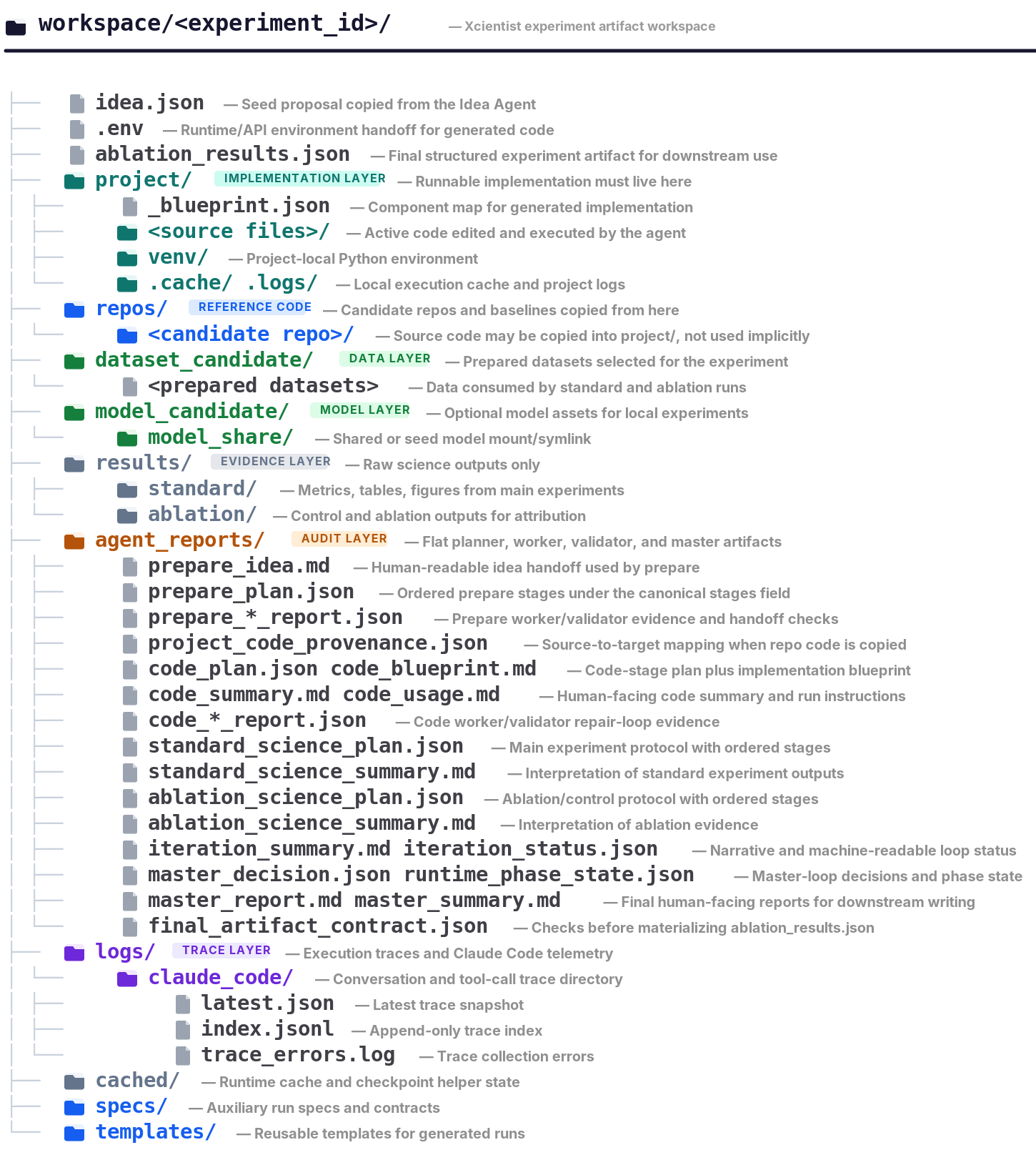}
    \caption{The artifact tree of Xcientist in experiment validation stage.}
    \label{fig:experiment_artifact_tree}
\end{figure}

\subsubsection{Master Scheduling and Convergence Judgment}
The master scheduler implements an outer loop driven by workspace state. At the beginning of each iteration, it evaluates four gates in strict priority order: self-containment of the project code, completion of the implementation phase, completion of standard experiments, and completion of ablation science. Self-containment is evaluated first through a static scanner that inspects the \texttt{project/} directory for imports or path references reaching outside the project root, including \texttt{sys.path} injection, editable installs, and local path imports. If any violation is found, the scheduler immediately demands code repair regardless of other validator statuses, because unresolved local dependencies prevent reconstruction of the declared execution context.

The scheduler writes its decision, phase status, blocking issues, and evidence file paths into structured runtime artifacts. These files serve both as an audit trail and as resumption checkpoints; if the process is interrupted, the scheduler can reload the last recorded state and continue from the appropriate phase. Convergence is defined strictly: every required phase must carry a validator-backed pass status, no unresolved blocking issues may remain, and all canonical components must have corresponding ablation evidence. A phase marked partial or blocked is never accepted as sufficient, regardless of textual justification.

\subsubsection{Code Enablement and Ablation Exposure}
The implementation phase translates the research proposal encoded in the handoff documents into self contained runnable code under \texttt{project/}. The planning layer produces an ordered sequence of implementation steps, each specifying goals, input paths, permitted write roots, and a verification command. The plan must conclude with a mandatory integration smoke test that runs the full experiment command on the real prepared dataset and model, confirming that the entire implementation chain is materially executable.

The code must expose a disabling or replacement condition for each canonical component while preserving the remaining implementation. Each condition documents what changes. Runnable code and entry points reside under \texttt{project/}; selectively copied external code is recorded in a source-to-destination provenance manifest. These controls make component tests possible, but attribution still depends on executing appropriate comparisons.

\subsubsection{Standard and Ablation Science}
The science phase splits into two validation lanes that share the same worker-validator infrastructure but pursue distinct epistemic goals. Standard science assesses whether the full method outperforms the baseline condition. It constructs an ordered sequence of experiment steps that run the baseline using standard components and the full method using all idea components, both operating on the prepared dataset and producing comparable metrics. Each step advances only after the validator confirms that real execution artifacts exist under \texttt{standard\_\allowbreak results\_\allowbreak dir}.

Ablation science measures the marginal contribution of each canonical component by systematically disabling one component at a time while holding the rest constant. The ablation plan produces a step list whose length exactly equals the number of components in the research proposal, and each step corresponds to one component in the same order. For each step, the experiment variant with that component disabled is executed, and the validator evaluates whether the collected metrics, confidence estimates, and method context descriptions satisfy the phase contract. All raw outputs are written to \texttt{ablation\_\allowbreak results\_\allowbreak dir}. The required mapping makes omissions detectable. Missing or invalid component tests remain unresolved requirements and prevent a fully validated status.

\subsubsection{Iteration Integration and Final Evidence}
After each master iteration completes, an iteration reporter summarizes the current experiment status by reading the validator reports, raw result windows, and decision history. It produces two artifacts: a human-readable summary and a machine-readable status file that follows a strict schema recording completion status, readiness flags, and blocking issues for each phase. The scheduler reads this file at the start of the next iteration to ground its routing decision in the most recent validated state rather than in stale assumptions.

When the scheduler finally selects converged, the pipeline synthesizes a canonical ablation results file containing per-component results, metric values, confidence levels, and method context analyses. This file summarizes the completed component comparisons and their validation status. Should the process reach the maximum iteration limit without achieving convergence, the scheduler returns the best available progress and explicitly records that execution stopped due to the iteration limit, preventing incomplete evidence from being presented as a validated conclusion.

\subsection{Report Writing}
The report-writing module converts source code, experimental outputs, iteration records and cited literature into a technical account of the study. Its checks connect implementation details and numerical claims to the artifacts from which they were drawn.

The module follows a write--audit--repair loop. Named functions, parameter values and file paths are checked against the project; scientific claims are checked against results and source papers. Table~\ref{tab:report_writing_stages} lists the stages and their outputs.

\begin{table}[h]
\centering
\caption{Principal stages of the report writing pipeline and their communicative roles.}
\label{tab:report_writing_stages}
\begin{tabular}{@{}p{0.22\linewidth} p{0.34\linewidth} p{0.34\linewidth}@{}}
\toprule
\textbf{Stage} & \textbf{Objective} & \textbf{Output Mandate} \\
\midrule
Project Exploration & Ingest the experiment workspace, identify key files, search for relevant papers, and produce a structured outline. & Must produce \texttt{blog\_\allowbreak idea.md} with verified project overview, architecture, and candidate papers for citation. \\
Article Composition & Write the full blog article section by section, grounded in source code and downloaded papers. & Must produce \texttt{blog\_\allowbreak article.md} with faithful code excerpts, proper citations, and image placeholders. \\
Quality Analysis & Audit the article against source fidelity, research integrity, and engineering depth criteria. & Must produce \texttt{blog\_\allowbreak analysis.md} with a 0-100 score and a prioritized issue list. \\
Iterative Refinement & Repair verified issues until the quality score exceeds a threshold or the loop limit is reached. & Must update \texttt{blog\_\allowbreak article.md} and graph method files without altering correct content. \\
Image Generation & Generate figures from graph method files and replace placeholders with embedded images. & Must produce \texttt{figure\{N\}.png} for every \texttt{graph\{N\}.md} and update the article in place. \\
\bottomrule
\end{tabular}
\end{table}

\subsubsection{Project Exploration and Outline Synthesis}
The module reads project documentation and core source files, then queries the evidence graph for related papers. Candidate papers are retrieved through Semantic Scholar and inspected before inclusion in the citation pool.

The exploration stage writes \texttt{blog\_\allowbreak idea.md}, containing the project overview, architecture, technology stack, outline and candidate citation table. The table records each paper's relevance and intended location. Composition is restricted to these papers, and each retained reference must be cited in the article.

\subsubsection{Structured Composition}
For each outline section, the writer reads the relevant code fragments and assigned papers. The writing configuration favors connected prose, limits pseudocode to fifteen lines and excludes routine imports, logging and error-handling boilerplate.

Citation is performed during composition using superscript markers such as \texttt{<sup>[1]</sup>}. Every paper listed in the References section must be cited in the article body; conversely, no paper may be listed that is not actually referenced. This bidirectional constraint prevents both dangling references and missing attributions. The module also inserts \texttt{<graphN>} placeholders at positions where visual explanation would strengthen the narrative, and writes a corresponding \texttt{graph\{N\}.md} method file that describes the desired image in terms of visual concept, main elements, and key message.

\subsubsection{Quality Analysis and Source Fidelity}
The audit scores Content Quality, Engineering Depth, Source Fidelity, Research Integrity, E-E-A-T Signals and AI Citation Readiness out of 100 points in total. Source Fidelity has the largest weight. The auditor searches the workspace to check named functions, parameter defaults and paths.

Research Integrity is verified by checking that every cited paper exists as a PDF in the blog workspace and that the claims attributed to it can be confirmed from the PDF content. A penalty of twenty points is applied if papers are missing or citations cannot be verified. Image quality is evaluated separately on description clarity, position, content match, and uniqueness, then combined with the article score through a weighted formula. The final output is \texttt{blog\_\allowbreak analysis.md}, which contains the score, a rating, a detailed breakdown, and a prioritized list of critical, high, medium, and low priority issues.

\subsubsection{Iterative Refinement and Convergence}
Refinement enters a repair loop that continues until the quality score exceeds 90 or a maximum of three iterations is reached. In each iteration, the module reads \texttt{blog\_\allowbreak article.md} and \texttt{blog\_\allowbreak analysis.md}, verifies every flagged issue against the source code before fixing it, and updates the article and graph method files accordingly. Critical issues such as nonexistent functions or parameter mismatches are always fixed; medium and low priority issues are addressed only when the improvement is unambiguous. After refinement, the audit is rerun to produce a new score, and the loop repeats.

The loop retains verified content and repairs confirmed issues. The score is a workflow criterion, not a guarantee that every claim is correct. Code-fence delimiters, indentation and code-block completeness are preserved during revision.

\subsubsection{Image Generation and Integration}
After the article converges, the pipeline enters the image generation stage. For every \texttt{graph\{N\}.md} method file, the system invokes an image generation provider to produce a corresponding \texttt{figure\{N\}.png}. The method file serves as a structured prompt that describes the visual concept, main elements, and key message, ensuring that generated images are aligned with the article content rather than decorative. After generation, the \texttt{<graphN>} placeholders in \texttt{blog\_\allowbreak article.md} are replaced with standard Markdown image tags, producing a self contained article with embedded figures.

\subsubsection{Checkpointing and Resumability}
After each stage, \texttt{workflow\_\allowbreak status.json} records completion. An interrupted workflow resumes from the last completed stage.

\subsubsection{Integration with the Research Harness}
The final output is an article accompanied by an audit report and figures. Its inputs remain available under the experiment workspace, including \texttt{project/}, result files and iteration reports. Unresolved evidence gaps should remain visible in the report; passing a writing-quality check does not replace experiment validation.

\subsection{System Interface}
The frontend and backend coordinate multi-stage runs and retain their artifacts and event histories. Users can launch a workflow, inspect partial results and failures, and return to an earlier run. This supporting infrastructure implements the user interface described in the main text.

\begin{table}[h]
\centering
\caption{High-level roles of the frontend--backend stack in \xcientist.}
\label{tab:frontend_backend_roles}
\begin{tabular}{@{}p{0.21\linewidth} p{0.24\linewidth} p{0.45\linewidth}@{}}
\toprule
\textbf{Layer}              & \textbf{Primary Role}        & \textbf{Main Responsibilities} \\
\midrule
Frontend\newline interaction layer  & User-facing control surface  & Topic creation, run launch, run inspection, artifact viewing, trace browsing, message inspection and approval interaction. \\
Backend orchestration layer & Execution coordination layer & Run lifecycle management, stage dispatch, event emission, artifact registration, runtime materialization and failure propagation. \\
Deployment/\allowbreak runtime layer    & Operating environment        & Persistent storage, service hosting, per-run workspace isolation, environment binding and containerized stage execution. \\
\bottomrule
\end{tabular}
\end{table}

\subsubsection{System-Level Design Objectives}

The stack maintains stable interfaces for launching and inspecting runs, maps a topic request to execution stages, and persists runtime events and artifacts. The current deployment targets a single host for internal use and controlled end-to-end evaluation.

\subsubsection{Frontend Interaction Abstraction}

The frontend groups work by research topic. Its main operations are topic creation, run launch, state monitoring, message inspection and artifact viewing.

Three persistent lanes, \texttt{review}, \texttt{ideation} and \texttt{experiment}, summarize workflow status. The backend maps these interface labels to the internal execution stages.

Each topic retains its run history, stage states, event streams and artifacts. Trace views, message drawers and approval interactions let users inspect the context of individual decisions and failures.

\subsubsection{Snapshot--Stream Runtime State Model}

To synchronize the interface with active execution, \xcientist{} uses a snapshot--stream runtime state model. When a page is opened, the frontend requests a snapshot containing the materialized topic state, recent events, current agent states and known artifacts. This provides an authoritative starting view without requiring the client to reconstruct history from incremental messages.

After initialization, WebSocket events update lane status, subtasks and artifact registrations. Reconnecting clients request a fresh snapshot before resuming the stream; backend persistence retains the durable run history.

\subsubsection{Backend Orchestration and Interface Preservation}

The backend preserves the REST and WebSocket interfaces while mapping user-visible lanes to internal stages. It creates run-local contexts, dispatches work and emits completion or failure events.

The backend translates user requests into run objects and stage plans, and normalizes stage outputs into artifacts and events. It also applies module-availability and approval rules and persists state changes.

\subsubsection{Per-run Runtime Materialization}

Each run receives a separate directory structure and generated configuration files. Stage outputs are assigned to run-specific locations to reduce interference between concurrent or successive studies.

Configuration views bind paths, output roots and workspace references to the active run. This gives survey, ideation and experiment modules a consistent namespace despite differences in their internal layouts. Preserving this execution context helps reconstruction but does not by itself guarantee scientific reproducibility.

Partial survey outputs, candidate ideas and tool logs remain in the run directory after an interruption and can be used for diagnosis or resumption.

\subsubsection{Artifacts, Events and Observability}

Artifacts store content such as surveys, ideas, logs and experiment results. Events record state transitions, progress, warnings and failures. Both are registered with the run and exposed through the frontend.

Events locate an interruption or revision in time; artifacts preserve the associated evidence. Users can revisit these records and compare successive runs under the same topic.

\subsubsection{Deployment Topology and Operational Scope}

The current deployment uses a relational database for persistent state, a backend for orchestration, a separately served frontend and, where needed, containerized stage execution. It assumes a single host.

\subsubsection{Current Constraints}

Lane identifiers and event contracts are stable interface assumptions. Artifact persistence uses the local filesystem, and orchestration does not provide cross-host scheduling or artifact management. These choices limit scaling and isolation but keep the current deployment simple.

The interface supports inspection and runtime organization. Scientific validity still depends on the research methods, completed comparisons and interpretation of the resulting evidence.

\label{sec:methods_cdr}

\paragraph{Scope and evidence preparation.}
The CDR analysis compared the retained records of \xcientist{}, EvoScientist, InternAgent-1.5 and AI-Scientist-v2 within each research task. This comparison of research systems is distinct from evaluating the task-specific methods they produced. The offline audit organized reports, literature records, design revisions, code, execution information and experimental outputs into source-linked evidence bundles. Baseline adapters used the available research-trajectory and literature-evidence records, inheriting earlier material only through explicit continuation links. The \xcientist{} records also linked selected idea documents to experimental workspaces. Text was divided into retrievable blocks while retaining source locations and coverage gaps. Missing, unread or truncated material was not treated as proof that an experiment had not occurred.

\paragraph{Claim extraction and applicability.}
A language model extracted distinct accepted core claims from the designated research endpoints, retaining the conditions, comparison, attribution and original wording. Exact and semantic deduplication combined repeated statements of the same conclusion. Hypotheses, plans, withdrawn claims, isolated table entries and execution metadata were not counted as separate conclusions. Selected idea documents could additionally supply accepted literature-based design rationales for the evidence-to-idea audit; expected performance was not treated as an experimental finding. Each claim entered a relation-type denominator only when its wording required that relation. Claims and evidence were interpreted in their corresponding design and revision context.

\paragraph{Evidence review.}
BM25 retrieval and explicit file references supplied literature, implementation, execution and result evidence for each claim. The language-model judge specified the necessary support relations, their status and supporting source quotations, with targeted retrieval when more evidence was needed. Programmatic checks verified evidence identifiers and quotations, checked that support did not consist solely of a report repeating its own assertion, and required execution as well as implementation evidence for design-identity judgments. Numerically bound claims were checked by recalculating the relevant quantities from the retrieved values. A numeric match did not by itself establish source attribution, component contribution or generalization. Semantic judgments and any subsequent adjudication remained part of the audit record.

\paragraph{Classification and reporting.}
A claim was supported for a relation type when all necessary relations were verified, drifted when at least one necessary relation was confirmed to fail, and otherwise unverifiable when missing evidence prevented resolution. Confirmed drift took precedence over unverifiability within the same type. Evaluation errors and unresolved review states were kept separate from unverifiable claims; the latter denotes missing scientific support records, not a failed model call. The reported CDRs were recomputed from the resolved \(D/U/N\) counts in Table~\ref{tab:cdr_counts}. We also display \(D/N\) and \(U/N\) separately. A claim may require more than one relation type, so category counts are not added to infer a unique total. The primary figures report each task and relation separately. The expanded-review summary pools claim--relation records only descriptively; neither analysis treats records within a run as independent repetitions. Table~\ref{tab:cdr_counts} provides the counts needed to recompute the fixed-budget rates.

\paragraph{Expanded and human review.}
Expanded review used \texttt{gpt-6-astra} with web access to reassess all 198 initially unverifiable claim--relation records. Records still unresolved after retrieval remained in the denominator. Human review covered 20 selected records: five evidence-to-idea, seven idea-to-implementation and eight experiment-to-claim. Reviewers could recover evidence beyond the automated bundle. The paired labels therefore reflect both evidence recovery and interpretation; they were not used to recalculate the aggregate fixed-budget CDRs. Appendix~\ref{sec:supp-claim-cases} records all 20 decisions. These retrospective audits do not isolate the causal contribution of individual harness modules or estimate variation across independent research runs.
\subsection{Case-study protocols}

\label{sec:methods_cases}

We evaluated the research harness in three settings chosen to vary the amount and type of prior constraint. Each run began from an existing method, retrieved task-specific evidence and evaluated successive component-based designs against that starting point. The training-free memory task provides a relatively open design space. Graph-structured traffic forecasting adds a fixed backbone, spatial structure and task-specific robustness tests. Multiscale physics-informed neural networks add governing equations and coarse--fine constraints that restrict both admissible designs and evaluation. Detailed version-level results are reported in  Table~\ref{tab:ed_detailed_case_results}.

\subsubsection{Training-free memory for LLM agents}

The memory case began from an A-Mem implementation~\cite{xu2026mem} and evaluated versions on the same subset of LoCoMo~\cite{maharana2024evaluating}. Evidence retrieval focused on provenance preservation, long-context failure modes, retrieval reliability and prompt cost. The search space was restricted to training-free memory operations and retained the language model as an external generator. The selected idea comprised immutable atomic-note memory, deterministic write-time enrichment, sparse handle support, slotted evidence retrieval and a deduplication guardrail.

The baseline, complete candidate and available component-disabled conditions used a shared execution path. Embeddings were generated with \texttt{all-MiniLM-L6-v2}, and answers were generated with \texttt{gpt-4o-mini}. Evaluation used category-level textual F1 for single-hop, multi-hop, temporal, open-domain and adversarial questions, together with overall textual F1 and average token length per query. Baseline and complete-candidate comparisons used matched data, embedding and answer-generation settings.

\subsubsection{Graph-structured traffic forecasting}

The traffic-forecasting case began from Graph WaveNet~\cite{wu2019graph} and the PEMS-BAY dataset~\cite{li2017diffusion}. Evidence retrieval prioritized diffusion-based backbones, spatial attention designs, missing-data robustness and horizon-specific evaluation. The intended design space retained the diffusion backbone as the main predictor and restricted changes to residual branches and their interfaces. The initial replacement candidate departed from that boundary; subsequent revisions restored the backbone. Its retained result is reported as part of the design history, not as evidence for a backbone-preserving extension.

The standard comparison evaluated Graph WaveNet and successive candidates on clean PEMS-BAY forecasting. Performance was measured by mean absolute error averaged across horizons and at 15-, 30- and 60-min horizons. Robustness was evaluated under encoder-side block missingness, including the block-40\% condition. Component tests removed or replaced the proposed projection or coverage operation while holding the backbone and remaining residual path fixed. These tests were used to determine whether the corresponding component changed clean forecasting error or robustness degradation under the declared protocol.

\subsubsection{Multiscale physics-informed neural networks}

The physics-constrained case addressed multiscale heat equations with physics-informed neural networks. Evidence retrieval focused on spectral bias, optimization coupling, coarse--fine interference and prior multiscale PINN designs. Candidate ideas were required to remain trainable under partial-differential-equation, boundary-condition and initial-condition losses. The search space was further restricted so that the coarse branch represented the global solution and residual branches supplied fine-scale corrections without overriding the coarse prediction.

Successive candidates were evaluated on \texttt{heat1d\_multiscale}, \texttt{heat2d\_multiscale} and \texttt{pinnacle\_heat}~\cite{hao2024pinnacle}. External comparisons included MMPINN~\cite{wang2024practical} and MultiscalePINNs~\cite{wang2021eigenvector}. The primary outcome was mean relative \(L_2\) error with the reported variation for each task.

Average rank across the three settings summarized version-level progress without replacing the per-task results. Validation also compared progressive and simultaneous residual activation and checked whether residual branches preserved the designated role of the coarse backbone. Conclusions were assessed separately for each benchmark so that improvement in one- or two-dimensional settings was not generalized across tasks without supporting results.

\clearpage
\bibliographystyle{xlance/IEEEtran2}
\bibliography{xlance/venues,reference}
\newpage
\appendix

\section{Literature Review Case Study}
The AutoSurvey example follows DeepSurvey from retrieval through analysis, drafting and refinement. Intermediate keynotes, clusters and comparison records supply the evidence used in the generated survey.

\subsection{Graph-backed Literature Retrieval}

Given the topic \emph{AutoSurvey}, DeepSurvey retrieves seed papers via the Semantic Scholar API and then expands along edges of Evidence Graph and citation graph to a bounded depth. This graph-based expansion is the key difference from flat semantic search: it discovers papers that are structurally connected to the seed set but may not share high surface-level similarity with the query. Table~\ref{tab:case_retrieval} illustrates how graph expansion enriches the evidence corpus beyond what seed-level retrieval alone would capture.

\begin{table}[h]
\centering
\caption{
\textbf{Graph expansion  case from seed papers for the topic \emph{AutoSurvey}.}
For each seed paper, citation/reference traversal discovers structurally connected papers that flat semantic retrieval would miss.
}
\label{tab:case_retrieval}
\begin{tabular}{@{}p{0.15\linewidth} p{0.20\linewidth} p{0.32\linewidth} p{0.22\linewidth}@{}}
\toprule
\textbf{Seed paper} & \textbf{Graph-expanded papers} & \textbf{Relationship type} & \textbf{Missed by flat retrieval?} \\
\midrule
Auto\-Survey~\cite{wang2024autosurvey}
& IterSurvey~\cite{zhang2025deep}, SurveyG~\cite{nguye2025surveyg}
& Papers that cite AutoSurvey and propose iterative extensions.
& Yes---low query similarity to ``AutoSurvey'' topic. \\

Auto\-Survey~\cite{wang2024autosurvey}
& Survey\-Gen~\cite{chen2025surveygen}, SurGE~\cite{su2025surge}
& Evaluation papers that reference AutoSurvey as a baseline or refer to its evaluation metric.
& Yes---low query similarity to ``AutoSurvey'' topic. \\

Auto\-Survey~\cite{wang2024autosurvey}
& SciSage~\cite{Shi2025SciSageAM}
& Papers that cite AutoSurvey and propose agentic method to solve ``AutoSurvey'' tasks.
& Yes---low query similarity to ``AutoSurvey'' topic. \\
\bottomrule
\end{tabular}
\end{table}

After graph expansion, a two-stage hybrid filter prunes the corpus: a coarse semantic filter retains candidates with sufficient title/abstract similarity to the seed set, and a fine-grained LLM re-ranker removes papers that are topically irrelevant despite structural connectivity. For the \emph{AutoSurvey} topic, this process produces a final evidence corpus of approximately 100 papers, substantially broader than the 8 seed papers, yet tightly focused by the filtering stages.

\subsection{Full-text Keynote Extraction}

For each retained paper, DeepSurvey generates a structured keynote from the full text. Table~\ref{tab:case_keynote} shows a condensed excerpt (condensed from 1,015 to 105 words) from the keynote for the AutoSurvey paper, illustrating the granularity of information captured.

\begin{table}[h]
\centering
\caption{
\textbf{Excerpt from the keynote for AutoSurvey.}
Full-text reading captures methodological mechanisms, experimental configurations, and limitations that are absent from the abstract.
}
\label{tab:case_keynote}
\begin{tabular}{@{}p{0.18\linewidth} p{0.72\linewidth}@{}}
\toprule
\textbf{Field} & \textbf{Content} \\
\midrule
Core contribution
& A four-phase pipeline (retrieval, outline, parallel drafting, integration) for automated survey generation using LLMs with RAG-based citation grounding. \\
Methodology
& Embedding-based retrieval over 530K arXiv papers; parallel subsection drafting by specialized LLMs; integration with coherence refinement; Multi-LLM-as-Judge evaluation. \\
Key results
& Near-human citation quality (recall 82.25\%, precision 77.41\% at 64K tokens); scores highly (e.g., at 16k tokens: coverage 4.66, structure
4.33, relevance 4.86, average 4.60) compared to human (4.66, 4.38, 5.00, 4.66) and naive RAG(4.46, 3.66, 4.73, 4.23). \\
Limitations
& Overgeneralization accounts for 51\% of citation errors; system focus only on CS topics; reliability of the automatic evaluation. \\
\bottomrule
\end{tabular}
\end{table}

The full-text keynote identifies overgeneralization as the main reported citation-error category, providing a specific limitation for later analysis.

\subsection{Clustering and Multi-perspective Relation Analysis}

DeepSurvey organizes the retrieved papers into thematic clusters. Table~\ref{tab:case_cluster} shows part of selected clusters for the \emph{AutoSurvey} topic, with their primary research questions.

\begin{table}[h]
\centering
\caption{
\textbf{Selected clusters for the topic \emph{AutoSurvey}.}
Papers are grouped by thematic questions, enabling local comparison and layered analysis.
}
\label{tab:case_cluster}
\begin{tabularx}{\linewidth}{@{}c c Y@{}}
\toprule
\textbf{ID} & \textbf{\# Papers} & \textbf{Primary research question} \\
\midrule
1 & 25 & How do explicit hierarchical structures (knowledge trees, citation graphs) compare with iterative outline refinement in capturing research evolution? \\
3 & 33 & What is the fundamental trade-off between structural organization and content depth in automated survey generation? \\
7 & 17 & How do iterative and recurrent planning mechanisms differ in memory handling, state updates, and error accumulation? \\
12 & 29 & How do evaluation philosophies differ across SurveyLens~\cite{guo2026surveylens}, SurGE~\cite{su2025surge}, SurveyBench~\cite{sun2025surveybench}, and DeepSurvey-Bench~\cite{zhang2026deepsurvey} in defining survey quality? \\
\bottomrule
\end{tabularx}
\end{table}

\begin{figure}[t]
  \centering
  \begin{subfigure}[t]{0.48\linewidth}
    \centering
    \includegraphics[width=\linewidth]{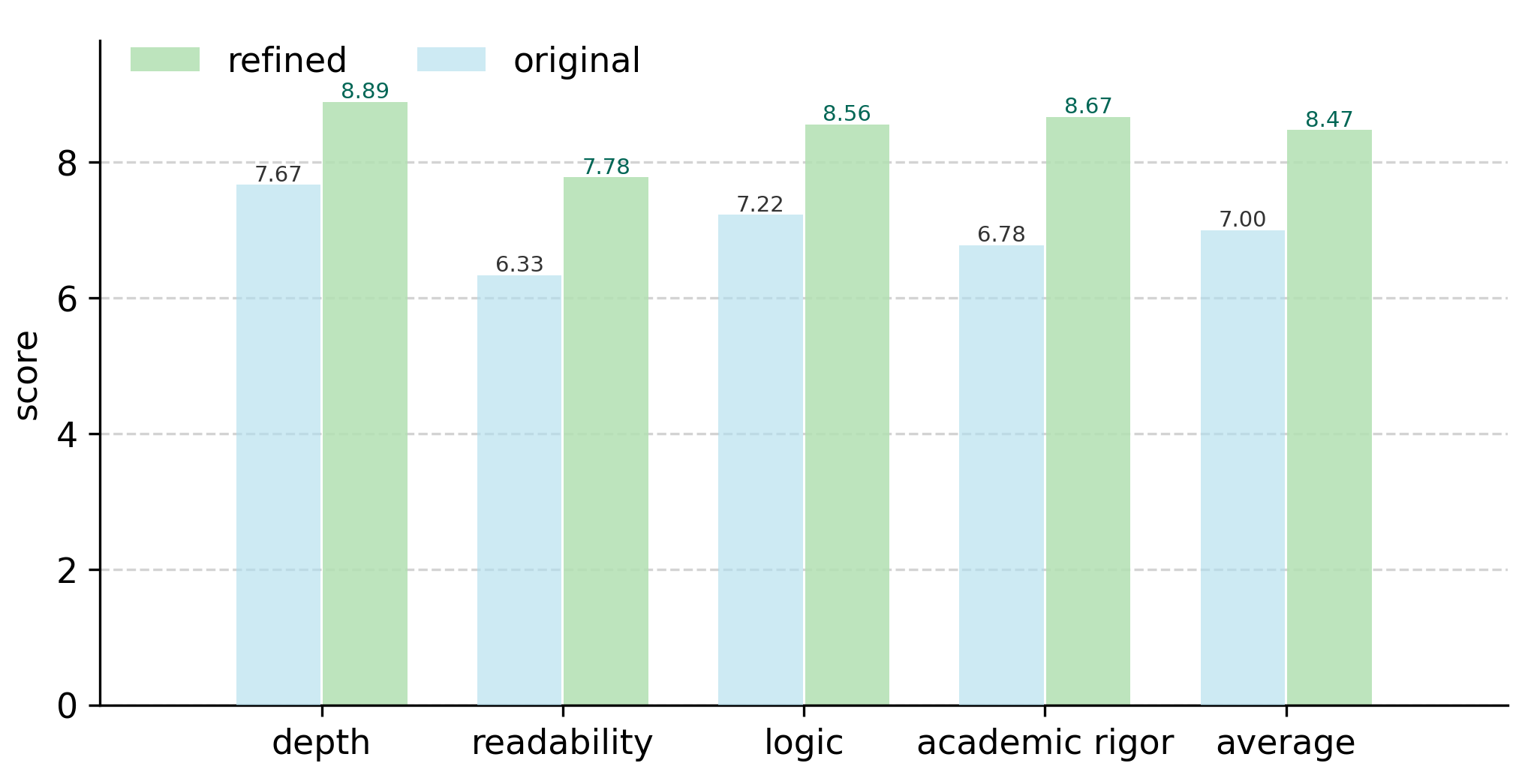}
    \caption{\textbf{Subsection refinement score comparison before and after.}
    Multi-granularity agentic refinement improves content quality across all evaluated subsections, with the largest gains in analytical depth and citation-claim alignment.}
    \label{fig:case_refinement}
  \end{subfigure}
  \hfill
  \begin{subfigure}[t]{0.48\linewidth}
    \centering
    \includegraphics[width=\linewidth]{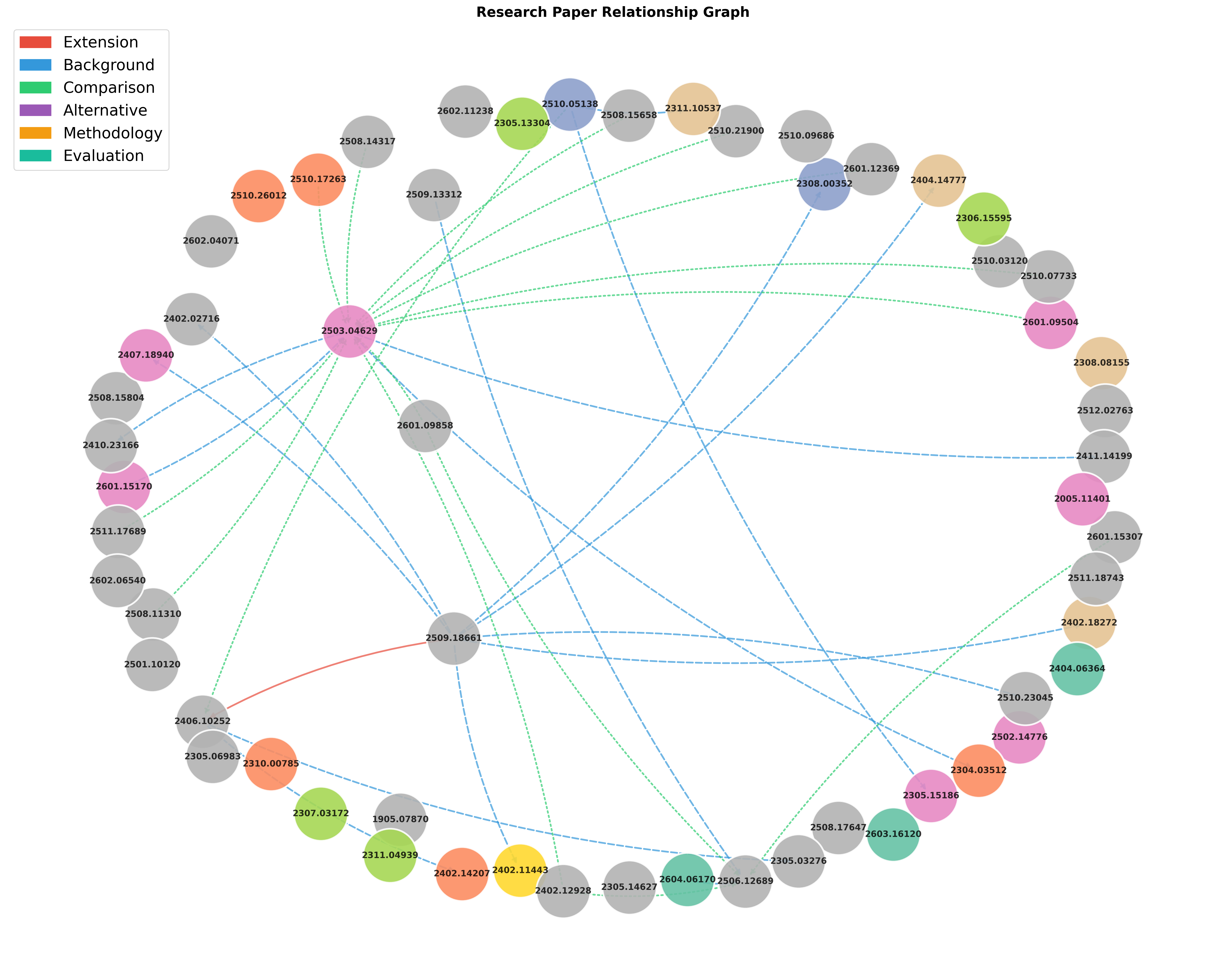}
    \caption{\textbf{Relation graph for the topic \emph{AutoSurvey}.}
    Typed citation edges (foundation, extension, substitution) capture technical lineage and method evolution trajectories, providing the Idea Agent with explicit methodological dependencies.}
    \label{fig:case_relation_graph}
  \end{subfigure}
  \caption{Comparison of subsection refinement and relation graph for \emph{AutoSurvey}.}
  \label{fig:case_two_subfigures}
\end{figure}

Within each cluster, three complementary analytical perspectives are applied. The \textbf{relation graph} (Figure~\ref{fig:case_relation_graph}) models typed citation relationships (e.g., foundation, extension, substitution), capturing technical lineage. For instance, it reveals that AutoSurvey's single-pass pipeline is extended by Agentic AutoSurvey through multi-agent orchestration, while RecurrentGPT~\cite{Zhou2023RecurrentGPTIG} proposes an alternative recurrent approach. The \textbf{comparison table} (Table~\ref{tab:case_comparison}, excerpted from the evaluation cluster) aligns papers along key dimensions. The \textbf{guided Q\&A synthesis} extracts cross-paper patterns. The following excerpt illustrates a key trade-off in structural planning:

\begin{quote}
\small
\emph{Across the clustered papers, survey generation mainly differs in how structure is built: some methods use explicit hierarchical models such as knowledge trees or citation graphs to capture research evolution and logical flow, while others rely on iterative outline refinement for greater flexibility and efficiency. The former tends to provide stronger organization and lineage modeling at higher cost, whereas the latter is simpler and more adaptive but less explicit about evolution.}
\end{quote}

The synthesis distinguishes explicit models of research evolution from iterative outline refinement, giving the Idea Agent two concrete planning approaches to compare.

\begin{table}[H]
\centering
\small
\caption{
\textbf{Comparison table excerpt for evaluation benchmarks.}
Structured alignment along key dimensions enables systematic horizontal comparison within a research theme.
}
\label{tab:case_comparison}
\begin{tabular}{@{}p{2.0cm}p{2.5cm}p{3.2cm}p{2.5cm}p{3.2cm}@{}}
\toprule
\textbf{Benchmark} & \textbf{Evaluation Focus} & \textbf{Evaluation Methodology} & \textbf{Scope of Benchmark} & \textbf{Key Innovation} \\
\midrule
Auto\-Survey~\cite{wang2024autosurvey}& Content quality and citation quality & Multi-LLM-as-judge evaluation & Computer science (AI topics) &  multi-LLM-as-judge evaluation\\
Survey\-Gen~\cite{chen2025surveygen} & Citation quality and content consistency & Automatic metrics and human evaluation & Multi-domain (4,200 surveys) & Large-scale dataset (SurveyGen) \\
SurGE~\cite{su2025surge} & Comprehensive, citation accuracy... & Multi-dimensional evaluation with human judge & Computer science (205 surveys) & multi-dimensional evaluation \\
Agentic Auto\-Survey~\cite{Liu2025AgenticAL} & Survey quality (core, writing, depth) & 12-dimensional evaluation framework & Computer science (AI topics) & 12-dimensional evaluation framework \\
\bottomrule
\end{tabular}
\end{table}

\subsection{Code Repository Analysis}

For papers with available source code, DeepSurvey analyzes repositories to extract implementation-level details. The code analysis for the \emph{AutoSurvey} topic reveals a clear evolutionary trajectory: from monolithic single-pass systems toward modular, multi-agent, iterative frameworks. Key findings include:

\begin{itemize}[nosep]
    \item \textbf{Dominant architecture}: hierarchical pipelines combining RAG, dynamic planning, specialized agent collaboration, and iterative refinement.
    \item \textbf{Major implementation gaps}: deep understanding, evaluation alignment, long-context handling, and domain generalization remain poorly addressed at the code level.
    \item \textbf{Framework convergence}: PyTorch (v2.0.0+) with Hugging Face Transformers forms the standard stack; LangChain and LlamaIndex dominate orchestration.
\end{itemize}

These code-grounded findings complement the textual analysis by exposing implementation patterns and engineering constraints that paper prose alone cannot convey.

\subsection{Outline-driven Drafting and Citation Enforcement}

\begin{figure}[H]
  \centering
  \includegraphics[width=0.8\linewidth]{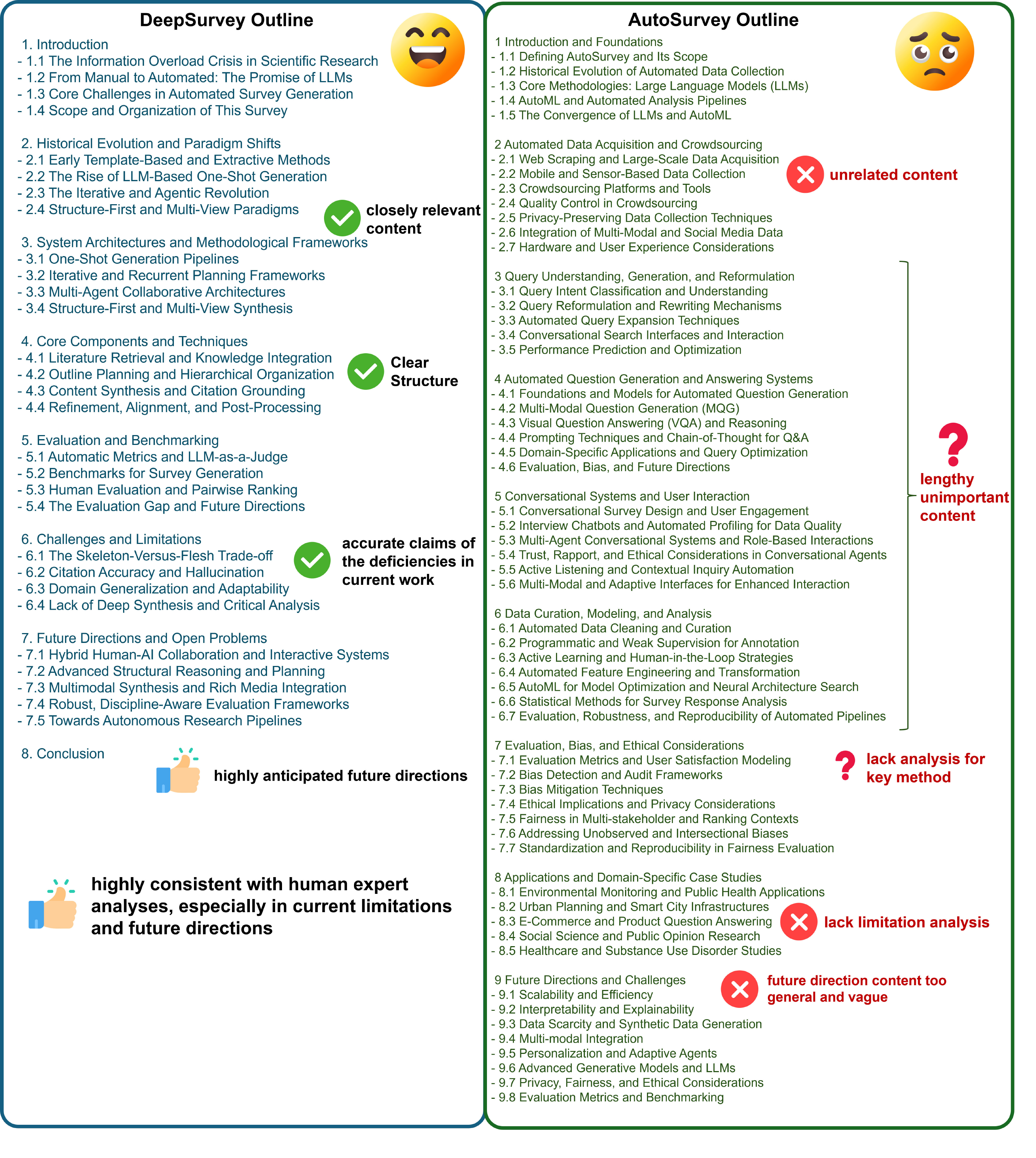}
  \caption{\textbf{Generated outlines with DeepSurvey (ours) and AutoSurvey for the topic \emph{AutoSurvey}.}
  DeepSurvey produces a focused, analytically structured outline grounded in the analysis substrate, while AutoSurvey's outline contains off-topic sections and lacks comparative synthesis.}
  \label{fig:case_study_outline}
\end{figure}

From the analysis substrate, DeepSurvey generates a hierarchical outline and assigns papers to sections. The outline for the \emph{AutoSurvey} topic comprises eight sections covering historical evolution, system architectures, core components, evaluation, challenges, and future directions. Each section receives a scoped description specifying its analytical angle and assigned paper subset.

For example, Section 2 (``Historical Evolution and Paradigm Shifts'') is assigned 18 papers covering the transition from template-based methods through LLM-based one-shot generation to agentic frameworks. The citation-anchoring process ensures that this section draws only from its assigned papers, creating a localized evidence constraint that reduces hallucination risk. After each paragraph, a citation verification mechanism checks consistency against the assigned set, triggering retry generation when errors are detected.

In this example, DeepSurvey organizes the topic around one-shot, iterative, multi-agent and structure-first approaches. The AutoSurvey outline includes sections on mobile data collection, crowdsourcing and healthcare that fall outside the intended topic. The comparison illustrates the outlines produced in this case; it does not isolate the causal effect of any one synthesis module.

\subsection{Multi-granularity Agentic Refinement}

After initial drafting, DeepSurvey applies refinement at three granularities. The centralized planning agent coordinates specialized roles (keynote reader, reviewer, reviser) through explicit action plans, with global memory preserving feedback across rounds. The refinement process validates citation-claim alignment at the paragraph level. For instance, when a draft paragraph claims that ``AutoSurvey achieves near-human performance,'' the reviewer checks whether the cited paper's reported metrics actually support this comparison. If the claim exceeds the evidence, the reviser either narrows the claim or retrieves additional supporting evidence from the keynote database. Figure~\ref{fig:case_refinement} shows the score improvement across subsections before and after refinement.

\newpage
\section{Idea Generation Case Study}
This worked example follows an additional memory-ideation trajectory initialized from a previous candidate, available component tests and a mature-idea anchor. It develops family-calibrated rank-and-pack proposals. These proposals are distinct from the slotted-retrieval system evaluated in Task 1; the search scores below are idea evaluations and do not constitute additional performance results.

\begin{table}[H]
\centering
\caption{
\textbf{Literature evidence retrieved for training-free agent memory.}
\textsc{Xcientist} generated the query \emph{Efficient Memory Retrieval and Evidence Packing} and retrieved papers that collectively characterize the design space of training-free LLM-agent memory.
}
\label{tab:case_study_rag_evidence}
\begin{tabular}{@{}p{0.25\linewidth} p{0.3\linewidth} p{0.35\linewidth}@{}}
\toprule
Evidence cluster & Representative retrieved papers & Relevance to ideation \\
\midrule
Adaptive memory at test time
& Dynamic Cheatsheet~\cite{suzgun2026dynamic}; TAME~\cite{cheng2026tame}
& Supports training-free or test-time memory updates, motivating lightweight memory evolution without model fine-tuning. \\

Hierarchical and procedural memory
& Learning Hierarchical Procedural Memory~\cite{forouzandeh2025learning}; Survey on LLM-agent memory mechanisms~\cite{zhang2025survey}
& Suggests that memory should be structured into reusable units and selected through explicit mechanisms rather than raw transcript recall. \\

Retrieval and evidence packing
& KVCOMM~\cite{ye2026kvcomm}; memory-mechanism survey~\cite{zhang2025survey}
& Highlights latency, prompt budget, compact evidence bundles, and retrieval efficiency as central constraints. \\

Trustworthiness and safety
& AgentAuditor~\cite{luo2026agentauditor}; A-MemGuard~\cite{wei2025memguard}; TrinityGuard~\cite{wang2026trinityguard}
& Motivates provenance tracking, verification, quarantine, and robustness against unreliable or poisoned memories. \\

Agent self-improvement and reflection
& Reflexion~\cite{shinn2023reflexion}; AnalogAgent~\cite{bao2026analogagent}; AutoManual~\cite{chen2024automanual}
& Provides evidence that reusable lessons and interaction traces can improve future behavior, but require explicit update and selection rules. \\
\bottomrule
\end{tabular}
\end{table}

\subsection{Literature Grounding}
Given the mature idea, \textsc{Xcientist} generates the RAG query \emph{Efficient Memory Retrieval and Evidence Packing}. This query retrieves and ranks papers relevant to training-free memory, agent memory evolution, retrieval efficiency, memory reliability, and evidence packing. The highest-ranked papers include work on adaptive memory at test time, trustworthy evolution of agent memory, hierarchical procedural memory, analogical agent improvement, KV-cache communication, and agent safety evaluation. It situates the target problem at the intersection of three concerns: how to store reusable memories without training, how to retrieve compact and useful evidence under latency constraints, and how to prevent unreliable or poisoned memories from dominating future decisions.

\subsection{Analysis of Field Structure}
On top of the retrieved papers, \textsc{Xcientist} performs an advanced analysis that separates the field into methods, consensus, open problems, and evaluation gaps. The analysis identifies several recurring method families: structured textual and procedural memory with explicit schemas, non-parametric cheatsheets or distilled rule memories, verbal reflection and lesson memory, dense-plus-lexical retrieval with compact packing, latent or KV-based working memory for efficiency, and verification or quarantine mechanisms for trustworthiness. This decomposition is useful because it does not treat ``memory'' as a single module. Instead, it separates write-time representation, read-time retrieval, evidence packing, runtime efficiency, and reliability control as distinct design axes.

The field consensus extracted by \textsc{Xcientist} further constrains the design space. Training-free external memory is preferred when online update, auditability, and deployment simplicity matter. Long-term memory should store compressed, high-signal artifacts rather than raw interaction transcripts. Read latency and prompt budget are core constraints, so retrieval should be multi-stage and evidence packing should be compact and decision-oriented. The analysis also concludes that semantic similarity alone is insufficient when wrong recalls are costly; reliability, provenance, temporal fit, and context fit should participate in retrieval decisions. These consensus statements narrow the search space away from heavier training-based or latent-memory solutions and toward explicit, inspectable retrieval rules.

\subsection{Problem Diagnosis}
The same analysis identifies the main failure modes of existing approaches. Memory retrieval remains too similarity-first and heuristic; admission, deduplication, and refinement often depend on brittle thresholds; memory growth creates duplicates, contradictions, and bloat; and evidence packing remains vulnerable to redundancy and order sensitivity. Temporal and provenance metadata are useful but noisy, and many systems lack retrieval rules that benefit from these signals without collapsing when they are absent. The analysis also incorporates prior ablation evidence: additional slotted reranking, separate handle indexing, and minority-aware provenance adjudication do not show clear marginal value in the observed setting. As a result, the system avoids proposing another auxiliary reranker or index. Instead, it redirects the idea toward a simpler and more testable retrieval principle.

\subsection{From Evidence to Root Idea}
The resulting mature idea is \emph{Family-Scoped Expected-Utility Retrieval for Atomic Note Memory}. The idea retains immutable span-grounded atomic notes as the durable memory unit, but shifts the center of the memory system from auxiliary indexes and reranking gadgets to family-scoped, reliability-aware retrieval. Each note is lightly enriched with embeddings, confidence-tagged entities and time expressions, compact QA keywords, short context descriptors, and a provenance-family identifier. Retrieval remains a bounded ANN plus lexical/facet screening stage, followed by deterministic rank-and-pack. The key change is that candidates are scored not only by query relevance, but also by temporal fit, provenance corroboration, and family reliability. Duplicate handling is simplified into family quotas and soft duplicate decay inside packing.

The resulting proposal links each retrieval change to the retrieved analysis or prior component-test record. It replaces similarity-first ranking with family-scoped expected utility while retaining atomic notes and bounded candidate screening.

\begin{table}[H]
\centering
\small
\caption{
\textbf{From advanced analysis to the root idea.}
The advanced analysis converts retrieved literature and ablation evidence into design constraints, then into the grounded mature idea.
}
\label{tab:case_study_analysis_to_idea}
\begin{tabular}{@{}p{0.24\linewidth} p{0.34\linewidth} p{0.34\linewidth}@{}}
\toprule
Analysis signal & Extracted conclusion & Design consequence \\
\midrule
Field consensus
& Training-free external memory is preferred when online update, auditability, and deployment simplicity matter.
& Avoid model fine-tuning and latent-memory modules; keep the system explicit, inspectable, and lightweight. \\

Memory representation
& High-value long-term memory should store compressed, high-signal artifacts rather than raw transcripts.
& Use immutable span-grounded atomic notes with compact metadata, instead of storing full interaction histories. \\

Retrieval limitation
& Similarity-first retrieval is brittle when wrong recalls are costly; reliability, provenance, temporal fit, and context fit should influence selection.
& Replace nearest-neighbor-style ranking with family-scoped expected utility over candidate notes. \\

Evidence packing problem
& Packed evidence can be redundant and order-sensitive, wasting prompt budget and weakening QA support.
& Use deterministic rank-and-pack with family quotas, role quotas, and soft duplicate decay. \\

Ablation evidence
& Extra slotted reranking, separate handle indexes, and minority-aware adjudication show weak or negative marginal value.
& Do not add another heavy reranker or auxiliary index; simplify the selection stack around a closed-form utility rule. \\

Evaluation gap
& End-to-end accuracy alone cannot show whether memory retrieval works.
& Add direct diagnostics: support recall, family calibration, duplicate rate, provenance coverage, temporal consistency, and order-robustness tests. \\
\bottomrule
\end{tabular}
\end{table}

\subsection{Monte-Carlo Tree Search}

After literature grounding and advanced analysis, \textsc{Xcientist} uses MCTS to refine the grounded mature idea into more concrete mechanism candidates. In this run, the root idea is \emph{Family-Scoped Expected-Utility Retrieval for Atomic Note Memory}, which scores 3.74 in simulation. The evaluator identifies several remaining defects, including \emph{validation gap}, \emph{brittle single path}, \emph{rare-regime failure}, and \emph{silent failure}. These defects then determine which expansion actions are prioritized. In particular, the search initially considers alternative-path contrast, theory transfer, and hierarchical decomposition as candidate edit operations, rather than sampling unconstrained idea variants.

\begin{figure}[H]
    \centering
    \includegraphics[width=\linewidth]{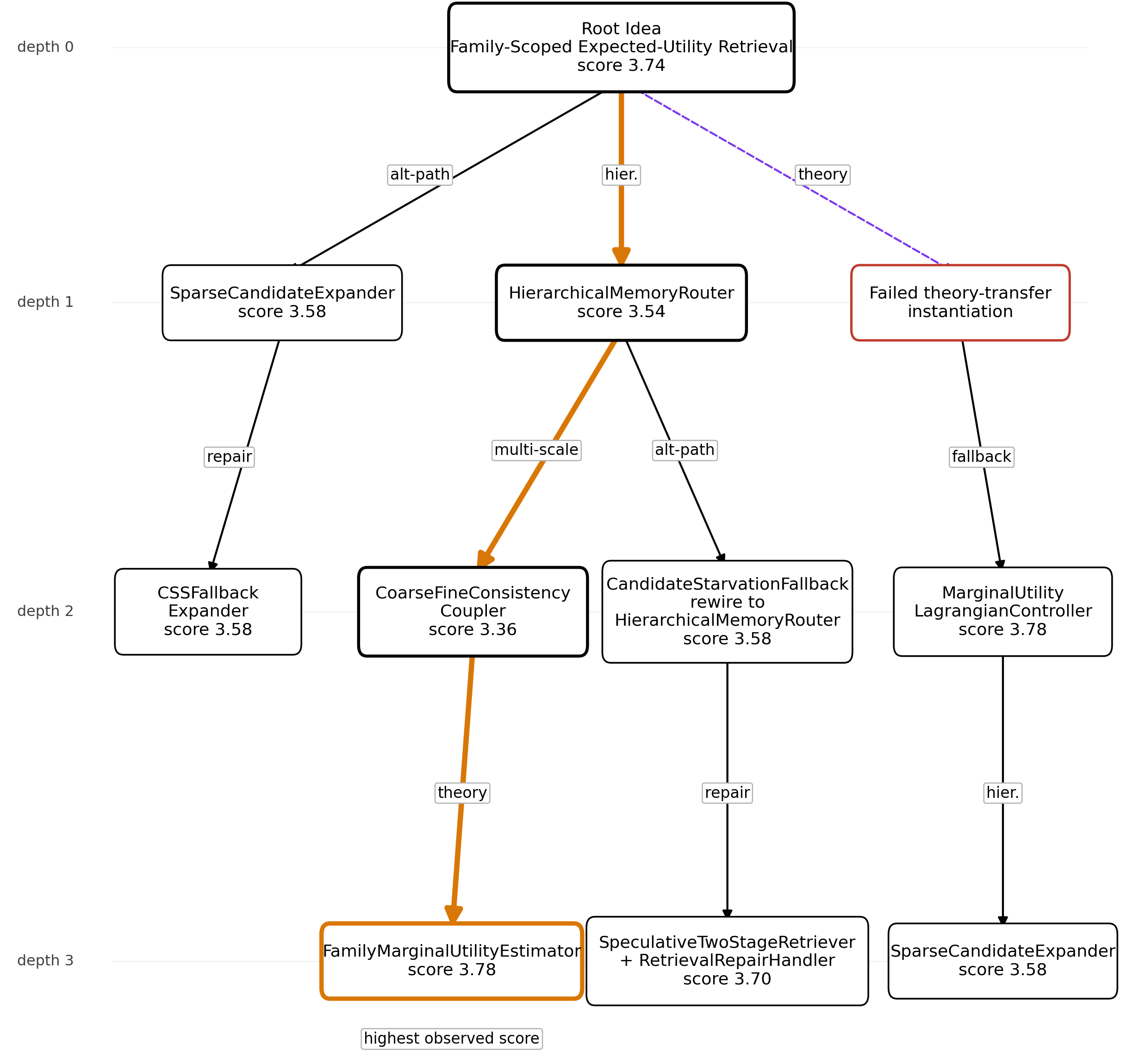}
    \caption{\textbf{Representative nodes in the compressed MCTS search tree.}
    The tree keeps only the most informative nodes from the log, showing how different expansion actions repair different defects in the root idea.}
    \label{fig:mcts_case}
\end{figure}

Figure~\ref{fig:mcts_case} visualizes a compressed MCTS search tree from this log. The figure does not include every generated node; instead, it keeps the most representative candidates needed to explain the search trajectory. The first expansion explores three qualitatively different repair directions. The alternative-path branch proposes a \emph{SparseCandidateExpander}, which directly targets candidate starvation in two-stage retrieval. The hierarchical branch proposes a \emph{HierarchicalMemoryRouter}, which separates family-level selection from note-level packing. The theory-transfer branch fails to instantiate a structured proposal at the first attempt, and is therefore retained as a failed branch in the trace. The failed branch remains in the search record.

The subsequent expansions show how the search moves from broad exploration to more targeted exploitation. The \emph{HierarchicalMemoryRouter} branch is further refined through a multi-scale coordination action, producing a \emph{CoarseFineConsistencyCoupler}. Although this child has a lower score, it exposes a concrete mechanism for checking whether coarse retrieval signals and fine-grained evidence are consistent. The highlighted path then continues to \emph{FamilyMarginalUtilityEstimator}, which obtains the highest observed score in the compressed tree. This final node sharpens the original family-scoped retrieval idea into a more explicit closed-form selection mechanism: instead of merely selecting families by a heuristic reliability score, the system estimates the marginal utility of adding evidence from a family to the final bundle. The retained path records the edits between the root and this candidate.

\subsection{Component-grounded Novelty Checking in MCTS Ideation}
Starting from the root idea \emph{Family-Scoped Expected-Utility Retrieval for Atomic Note Memory}, the system first classifies the domain as artificial intelligence and retrieves symbolic memory records for relevant component families. These records include historical ablation signals for components such as \texttt{modular\_\allowbreak atomic\_\allowbreak note\_\allowbreak enricher} and \texttt{two\_\allowbreak stage\_\allowbreak retrieval}, which provide evidence about whether a component has previously shown clear marginal utility. The root idea is then simulated and scored, with the evaluator identifying defects such as validation gap, brittle single path, rare-regime failure, and silent failure.

Before or during simulation, the evaluator retrieves component records such as \emph{AutoPrunedRetriever}, \emph{BUZZ}, \emph{MapAgent}, \emph{Reflective Memory Management} and \emph{CogMem}. It compares a candidate's proposed operation, component combination and execution path against these records.

Table~\ref{tab:evidence_graph_component_case} summarizes the evidence used for component grounding, novelty comparison and cross-domain transfer. These records identify prior mechanisms and design constraints against which the candidate can be assessed.

\begin{table}[H]
\centering
\small
\caption{
\textbf{Evidence-graph evidence used for component-grounded ideation.}
\textsc{Xcientist} retrieves component and paper nodes from the evidence graph to constrain candidate generation and novelty checking.
}
\label{tab:evidence_graph_component_case}
\begin{tabular}{@{}p{0.22\linewidth} p{0.36\linewidth} p{0.34\linewidth}@{}}
\toprule
Evidence source & Retrieved signal & Role in the MCTS search \\
\midrule
Symbolic component memory
& Historical ablation records for components such as \texttt{modular\_\allowbreak atomic\_\allowbreak note\_\allowbreak enricher}, \texttt{two\_\allowbreak stage\_\allowbreak retrieval}, and related retrieval modules.
& Prevents unconstrained feature expansion by checking whether a component has prior evidence of utility, latency cost, or inconclusive contribution. \\

Component novelty retrieval
& Top evidence nodes such as \emph{AutoPrunedRetriever}~\cite{wang2026pruning}, \emph{BUZZ}~\cite{zhao2024buzz}, \emph{FULORA}~\cite{wang2025walk}, \emph{MapAgent}~\cite{kong2025mapagent}, \emph{Reflective Memory Management}~\cite{tan2025prospect}, and \emph{CogMem}~\cite{zhang2025cogmem}.
& Provides a component-level novelty baseline, helping the evaluator decide whether a candidate is a new mechanism or a recombination of known retrieval/memory modules. \\

Theory-transfer retrieval
& Cross-domain core nodes such as \emph{Cortex}~\cite{ruan2026cortex}, \emph{SIM-RAG}~\cite{yang2025knowing}, and \emph{Dynamic Quality-Latency Aware Routing}~\cite{bao2025dynamic}.
& Supplies transferable mechanisms, including quality-latency routing, self-awareness triggers, and cache/budget trade-offs, which are instantiated as new controller components. \\

Simulation feedback
& Novelty, surprise, feasibility, impact, protocol score, detected defects, and failure modes.
& Converts retrieved evidence into a decision signal for MCTS: weak candidates are penalized, while candidates with clearer mechanisms and validation protocols become search anchors. \\
\bottomrule
\end{tabular}
\end{table}

\subsection{Idea Fusion}
After generating candidates under different idea-taste modes, the system receives five candidate ideas for the topic \emph{Training-free Memory System for LLM Agents}. These candidates share a common backbone, including \texttt{two\_\allowbreak stage\_\allowbreak retrieval}, \texttt{modular\_\allowbreak atomic\_\allowbreak note\_\allowbreak enricher}, \texttt{per\_\allowbreak slot\_\allowbreak quota\_\allowbreak and\_\allowbreak decay}, and \texttt{note\_\allowbreak attached\_\allowbreak micro\_\allowbreak handles}, but differ in their proposed core mechanisms. For example, the \emph{moonshot inventor} candidate proposes \texttt{CalibratedBayesianFamilyRankAndPack} and \texttt{ShrinkageAnalyticNullEstimator}, whereas the \emph{ambitious realist} candidate introduces a \texttt{HierarchicalMemoryRouter}. Rather than averaging these candidates or concatenating their modules, \textsc{Xcientist} treats fusion as a component-selection problem: it identifies which components carry the central novelty, which components should remain as support infrastructure, and which components should be rejected because they introduce unnecessary control complexity.

The fused draft selects the \emph{moonshot inventor} candidate as the host idea and forms the thesis \emph{Analytic Family-Calibrated Rank-and-Pack for Training-Free Atomic-Note Memory}. The selected core mechanism is \texttt{CalibratedBayesianFamilyRankAndPack}, supported by \texttt{ShrinkageAnalyticNullEstimator}. The former provides the main read-time family-aware calibration mechanism, while the latter supplies a closed-form, family-size-aware null model that makes the calibration fast, deterministic, and ablatable. At the same time, the fusion preserves mature infrastructure from other candidates: \texttt{two\_\allowbreak stage\_\allowbreak retrieval} keeps the retrieval path bounded and efficient, \texttt{modular\_\allowbreak atomic\_\allowbreak note\_\allowbreak enricher} provides the metadata needed by the calibrator, and lightweight guardrails such as quota decay and note-attached micro-handles preserve duplicate control and exact lookup without introducing a separate heavy index.

A key part of the fusion process is conflict resolution. The candidates disagree on whether family reliability should be persistent and feedback-updated or computed ephemerally within the current candidate pool. \textsc{Xcientist} resolves this by using only a small write-time prior and computing the main reliability adjustment from the bounded retrieved pool, avoiding drift-heavy mutable family state. The candidates also differ on whether to insert a router/controller layer. The fusion rejects \texttt{HierarchicalMemoryRouter}, because it adds coordination burden and weakens attribution; instead, family calibration is injected directly into the packer. Finally, the fusion avoids raw provenance-count bonuses, which can over-reward large or duplicated families, and replaces them with an analytic null model plus shrinkage. The retained design combines bounded retrieval, analytic family calibration and calibrated packing.

The initial fused draft is then passed through a local repair loop. The first repair step improves the score from 3.99 to 4.04 by narrowing the mechanism: the analytic null estimator becomes the thesis-bearing component, while deterministic packing is retained as the mature one-pass substrate with explicit degraded-mode behavior. Subsequent repair attempts explore alternative boundaries between calibration, packing, quota control, and metadata handling, but none improves over the first repaired version. The final candidate is therefore retained as \emph{Analytic Family-Calibrated Deterministic Rank-and-Pack for Training-Free Atomic-Note Memory}, with a score of 4.04. This trajectory shows that \textsc{Xcientist}'s fusion stage is auditable at the component level: each selected component has a stated role, each rejected component has an explicit reason, each conflict is resolved through a design decision, and the repair loop preserves only edits that improve the fused idea.

\newpage
\section{Experiment Validation Case Study}
This case follows the implementation and evaluation of \emph{Training-Free Slotted Evidence Retrieval for Scalable LLM Agent Memory}. The idea combines immutable notes, deterministic write-time enrichment and capacity-capped retrieval. Its four canonical components are \texttt{slotted\_\allowbreak evidence\_\allowbreak reranker}, \texttt{modular\_\allowbreak atomic\_\allowbreak note\_\allowbreak enricher}, \texttt{ultra\_\allowbreak sparse\_\allowbreak facet\_\allowbreak handle\_\allowbreak index} and \texttt{dedup\_\allowbreak minority\_\allowbreak aware\_\allowbreak provenance\_\allowbreak adjudicator}. The contract requires a separately testable implementation for each component.

\begin{table}[h]
\centering
\small
\caption{
\textbf{Contract-to-code realization in experiment validation.}
\textsc{Xcientist} converts each canonical idea component into an ablatable implementation module, enabling controlled validation rather than monolithic execution.
}
\label{tab:experiment_contract_to_code}
\begin{tabular}{@{}p{0.25\linewidth} p{0.30\linewidth} p{0.36\linewidth}@{}}
\toprule
Idea component & Implemented module & Validation role \\
\midrule
Immutable atomic-note memory
& \texttt{atomic\_\allowbreak store.py}
& Provides span-grounded \texttt{AtomicNote} and \texttt{AtomicMemoryStore} objects as the authoritative memory unit. \\

Write-time enrichment
& \texttt{note\_\allowbreak enricher.py}
& Adds cached embeddings, entity/time tags, keywords, context descriptors, and provenance labels without learned memory updates. \\

Sparse handle support
& \texttt{handle\_\allowbreak index.py}
& Supplies high-precision exact entity/time/relation handles as optional retrieval boosts rather than a separate canonical layer. \\

Slotted evidence retrieval
& \texttt{slotted\_\allowbreak reranker.py}
& Implements the main read-path mechanism: bounded ANN/lexical screening followed by capacity-capped evidence-slot assembly. \\

Deduplication guardrail
& \texttt{adjudicator.py}
& Suppresses duplicate-source flooding while preserving bounded minority contradictions. \\

System composition
& \texttt{system.py}, \texttt{run\_\allowbreak experiment.py}
& Exposes baseline, full, and component-ablation conditions through a shared runnable interface. \\
\bottomrule
\end{tabular}
\end{table}

\subsection{From Idea Contract to Runnable System}
The validation process first converts the idea contract into a self-contained implementation under \texttt{project/}. The code phase completes all nine implementation steps with validator-backed PASS verdicts. The resulting package includes \texttt{atomic\_\allowbreak store.py} for immutable span-grounded notes, \texttt{note\_\allowbreak enricher.py} for deterministic metadata extraction, \texttt{handle\_\allowbreak index.py} for exact sparse handles, \texttt{slotted\_\allowbreak reranker.py} for two-stage retrieval and evidence-slot assembly, and \texttt{adjudicator.py} for duplicate suppression and minority-conflict preservation. A composition layer, \texttt{SlottedMemorySystem}, exposes a \texttt{component\_\allowbreak config} dictionary, allowing the full system, baseline, and component ablations to be invoked through the same runtime path. This design makes implementation itself part of the audit trail: each conceptual component in the idea contract has a corresponding code module and an explicit ablation switch.

\subsection{Standard Science Validation}
After implementation, \textsc{Xcientist} runs a standard validation phase comparing a baseline condition with all components disabled against the full method with all components enabled. Both conditions use the same LoCoMo~\cite{maharana2024evaluating} subset, the same local \texttt{all-MiniLM-L6-v2} embedding model, and the same \texttt{gpt-4o-mini} answer generator. The standard science phase passes all validator checks: the overall improvement ratio exceeds the required threshold, the absolute improvement is above the minimum margin, predictions are non-empty, and the dataset identity is verified by MD5. The full method improves Overall F1 from 0.306 to 0.391, a 1.278$\times$ ratio, while reducing average token length from 2844.1 to 1017.2, a 64.23\% reduction. Improvements appear across all reported categories, including single-hop, multi-hop, temporal, open-domain, and adversarial questions.

\begin{table}[t]
\centering
\small
\caption{
\textbf{Validator-backed standard science results.}
The standard validation compares an all-disabled baseline against the full method under matched dataset, embedding model, and LLM generator conditions.
}
\label{tab:experiment_standard_science_case}
\begin{tabular}{@{}p{0.26\linewidth} p{0.18\linewidth} p{0.18\linewidth} p{0.26\linewidth}@{}}
\toprule
Metric & Baseline & Full method & Validation implication \\
\midrule
Single Hop F1
& 0.241
& 0.299
& Improves direct factual retrieval. \\

Multi Hop F1
& 0.147
& 0.188
& Higher multi-hop F1 for the complete system. \\

Temporal F1
& 0.242
& 0.310
& Higher temporal-question F1; component effects are not isolated. \\

Open Domain F1
& 0.117
& 0.223
& Higher open-domain F1 under the matched comparison. \\

Adversarial F1
& 0.617
& 0.788
& Indicates improved robustness under challenging questions. \\

Overall F1
& 0.306
& 0.391
& +0.085 absolute gain and 1.278$\times$ baseline F1. \\

Token Length
& 2844.1
& 1017.2
& 64.23\% reduction, supporting the efficiency claim. \\
\bottomrule
\end{tabular}
\end{table}

\subsection{Controlling Claim Boundaries}
The retained headline evaluation reports baseline and full-method results. Although component-test files exist, matched results were not available for every component under that evaluation. The conclusion is therefore limited to the full system's QA performance and token efficiency. Implementation switches and diagnostic artifacts do not establish each component's contribution.

\subsection{Validation Outcome}
The records connect the idea's component list to implementation modules and a matched standard comparison. They support the full-system performance and efficiency result while retaining the incomplete component tests as an explicit attribution limit.

\newpage
\section{Report Writing Case Study}
\begin{figure}[H]
\centering
\includegraphics[width=\linewidth]{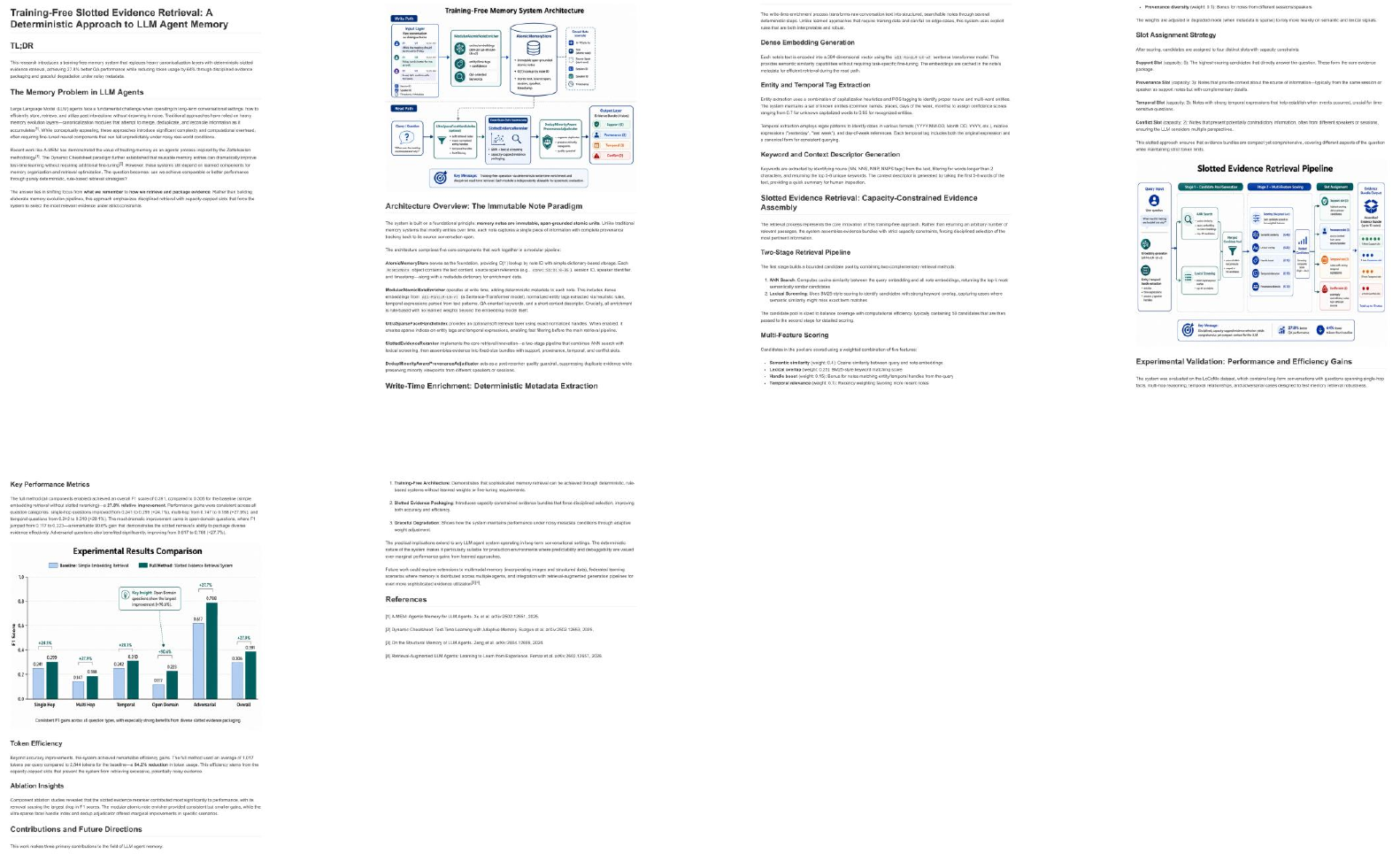}
\caption{\textbf{Example of evidence-grounded report generation.} \textsc{Xcientist} converts implementation artifacts, configuration values, experimental outputs, figure placeholders, and literature references into a structured technical report. The generated report organizes the method into problem framing, architecture description, retrieval pipeline, experimental validation, efficiency analysis, ablation insights, and future directions, while grounding narrative claims in concrete source-code components, quantitative results, and cited prior work.}
\label{fig:blog_memory}
\end{figure}

\subsection{Evidence Inputs and Report Framing}

The writing stage receives source code, results, figures and supporting papers. The generated report, \emph{Training-Free Slotted Evidence Retrieval: A Deterministic Approach to LLM Agent Memory}, describes the implemented memory system and its matched performance--cost comparison.

\subsection{Implementation-Grounded System Narrative}

The report first constructs a problem narrative. It contrasts heavy canonicalization and learned memory-evolution layers with a deterministic alternative based on immutable atomic notes and slotted retrieval. It then explains the architecture through concrete components: \texttt{Atomic\allowbreak Memory\allowbreak Store}, \texttt{Modular\allowbreak Atomic\allowbreak Note\allowbreak Enricher}, \texttt{Ultra\allowbreak Sparse\allowbreak Facet\allowbreak Handle\allowbreak Index}, \texttt{Slotted\allowbreak Evidence\allowbreak Reranker}, and \texttt{Dedup\allowbreak Minority\allowbreak Aware\allowbreak Provenance\allowbreak Adjudicator}. The named modules let readers locate the implementation of each write-path or read-path operation.

\subsection{Retrieval-Centered Technical Narrative}

The central technical narrative focuses on retrieval rather than memory storage. The report explains how notes are enriched at write time with embeddings, entity and temporal tags, QA keywords, and context descriptors, and how the read path builds a bounded candidate pool through ANN search and lexical screening. It then presents the main mechanism, \emph{slotted evidence retrieval}, in which candidates are assigned into support, provenance, temporal, and conflict slots under explicit capacity constraints. The description follows the implemented path from note representation through retrieval and evidence packing.

\begin{table}[t]
\centering
\small
\caption{
\textbf{Evidence-to-narrative transformation in report writing.}
\textsc{Xcientist} converts implementation artifacts, experimental outputs, and literature references into a structured report.
}
\label{tab:report_writing_evidence_to_narrative}
\begin{tabular}{@{}p{0.24\linewidth} p{0.34\linewidth} p{0.34\linewidth}@{}}
\toprule
Input evidence & Reported narrative element & Communication role \\
\midrule
Source-code components
& \texttt{Atomic\allowbreak Memory\allowbreak Store}, \texttt{Modular\allowbreak Atomic\allowbreak Note\allowbreak Enricher}, \texttt{Ultra\allowbreak Sparse\allowbreak Facet\allowbreak Handle\allowbreak Index}, \texttt{Slotted\allowbreak Evidence\allowbreak Reranker}, and \texttt{Dedup\allowbreak Minority\allowbreak Aware\allowbreak Provenance\allowbreak Adjudicator}.
& Grounds the architecture description in concrete implementation units rather than abstract module names. \\

Configuration values
& Candidate pool size, slot capacities, feature weights, and embedding model.
& Turns the method description into a reproducible system specification. \\

Experimental comparison
& Overall F1 improves from 0.306 to 0.391, with gains across single-hop, multi-hop, temporal, open-domain, and adversarial questions.
& Supports the central performance claim with category-level evidence. \\

Efficiency measurement
& Average token usage decreases from 2,844 to 1,017 tokens per query.
& Connects the method to the practical claim that disciplined evidence packaging improves efficiency. \\

Figure placeholders
& Architecture overview, two-stage retrieval pipeline, and experimental result chart.
& Places visual explanations at the points where they support problem framing, mechanism understanding, and empirical validation. \\

References
& A-MEM, Dynamic Cheatsheet, structural memory, and retrieval-augmented agent learning.
& Positions the report in the broader literature and supports future-work discussion. \\
\bottomrule
\end{tabular}
\end{table}

\subsection{Experimental Evidence Integration}

The report also integrates experimental evidence into the narrative. It states that the full method achieves an overall F1 score of 0.391 compared with 0.306 for the baseline, corresponding to a 27.8\% relative improvement. It further reports a 64.2\% reduction in token usage, from 2,844 tokens per query to 1,017 tokens per query. These quantitative claims are not left as unverified prose. In the quality report, \textsc{Xcientist} checks that the reported F1 values, relative improvement, token reduction, and category-level scores match the source experimental data. Thus, the report-writing stage preserves the link between metric claims and underlying artifacts.

\subsection{Post-Generation Audit}

\begin{table}[h]
\centering
\small
\caption{
\textbf{Post-generation audit of the report.}
The quality report checks whether the generated article remains faithful to the implementation, experiments, figures, and cited literature.
}
\label{tab:report_writing_audit}
\begin{tabular}{@{}p{0.24\linewidth} p{0.34\linewidth} p{0.34\linewidth}@{}}
\toprule
Audit dimension & Checked evidence & Outcome \\
\midrule
Source-code fidelity
& Verifies that all named functions and classes appear in the expected source files.
& Confirms that the report describes real implementation components. \\

Parameter alignment
& Checks candidate pool size, slot capacities, scoring weights, and embedding model against source configuration.
& Confirms that method details match the implementation. \\

Citation authenticity
& Verifies that referenced PDF files exist and that each reference is cited in the article body.
& Checks reference availability and citation usage. \\

Experimental consistency
& Compares reported F1 scores, relative improvement, token reduction, and category-level metrics against source data.
& Confirms that quantitative claims are artifact-grounded. \\

Figure placement
& Checks that architecture, retrieval-pipeline, and experimental-result figures are placed in appropriate sections.
& Ensures that visual evidence supports the report narrative. \\

Writing quality
& Evaluates content quality, engineering depth, source fidelity, research integrity, E-E-A-T signals, and AI citation readiness.
& Finds no substantive mismatch across the checked writing and source-fidelity dimensions. \\
\bottomrule
\end{tabular}
\end{table}

The post-generation audit checked source-code names, parameter values, citation files, reported metrics and figure placement. It recorded no substantive mismatch within those checks. This result concerns the generated report and its supplied artifacts; it does not replace the separate claim-level audit or resolve incomplete component comparisons.

\newpage

\section{Detailed Results and Claim Boundaries}
\label{app:detailed_results}
\begin{table}[!htbp]
\centering
\footnotesize
\setlength{\tabcolsep}{2.5pt}
\renewcommand{\arraystretch}{1.02}
\caption{\textbf{Detailed version-level results across the three case studies.}
\textbf{a}, Training-free memory-system versions evaluated on the same LoCoMo subset with \texttt{all-MiniLM-L6-v2} embeddings and \texttt{gpt-4o-mini} answer generation; higher F1 and lower token length are better.
\textbf{b}, Graph-forecasting versions evaluated on PEMS-BAY; lower MAE is better.
\textbf{c}, Generated multiscale-PINN versions and external baselines; each task entry reports mean relative L2 error followed by its reported variation, and lower error and average rank are better.}
\label{tab:ed_detailed_case_results}

\textbf{a}\quad Training-free memory system
\par\vspace{2pt}
\begin{tabular*}{\linewidth}{@{\extracolsep{\fill}}lrrrrrrr@{}}
\toprule
\textbf{Method}
& \makecell{\textbf{Single-hop}\\\textbf{F1}}
& \makecell{\textbf{Multi-hop}\\\textbf{F1}}
& \makecell{\textbf{Temporal}\\\textbf{F1}}
& \makecell{\textbf{Open-domain}\\\textbf{F1}}
& \makecell{\textbf{Adversarial}\\\textbf{F1}}
& \textbf{Overall F1}
& \textbf{Token length} \\
\midrule
A-Mem & 0.241 & 0.147 & 0.242 & 0.117 & 0.617 & 0.306 & 2844.1 \\
v1.0 & 0.175 & 0.121 & 0.158 & 0.131 & 0.724 & 0.286 & 1479.3 \\
v1.1 & 0.235 & 0.120 & 0.241 & 0.116 & 0.776 & 0.328 & 1667.8 \\
v1.2 & 0.271 & 0.169 & 0.299 & 0.138 & 0.599 & 0.335 & 807.1 \\
v2.0 & 0.138 & 0.103 & 0.305 & 0.096 & 0.821 & 0.311 & \textbf{601.6} \\
v2.1 & 0.254 & 0.184 & 0.229 & 0.153 & 0.800 & 0.358 & 813.7 \\
v2.2 & 0.249 & 0.148 & 0.378 & 0.095 & 0.814 & 0.375 & 744.3 \\
v3.0 & 0.208 & 0.168 & 0.301 & 0.150 & 0.646 & 0.313 & 2002.5 \\
v3.1 & 0.205 & 0.149 & 0.263 & 0.106 & 0.765 & 0.328 & 2083.9 \\
v3.2 & 0.285 & \textbf{0.199} & 0.286 & 0.128 & \textbf{0.832} & 0.388 & 849.9 \\
v4.0 & 0.284 & 0.176 & 0.360 & 0.184 & 0.596 & 0.346 & 1334.6 \\
v4.1 & \textbf{0.357} & 0.176 & \textbf{0.395} & 0.136 & 0.563 & 0.357 & 1436.8 \\
v4.2 & 0.299 & 0.188 & 0.310 & \textbf{0.223} & 0.788 & \textbf{0.391} & 1017.2 \\
\bottomrule
\end{tabular*}

\par\vspace{7pt}
\textbf{b}\quad Graph-structured traffic forecasting
\par\vspace{2pt}
\begin{tabular*}{\linewidth}{@{\extracolsep{\fill}}lrrrrrr@{}}
\toprule
\textbf{Method}
& \textbf{Average MAE}
& \textbf{15 min MAE}
& \textbf{30 min MAE}
& \textbf{60 min MAE} \\
\midrule
Graph WaveNet & 1.588 & 1.304 & 1.649 & 1.972 \\
v1 & 1.723 & 1.382 & 1.781 & 2.193  \\
v2 & 1.677 & 1.334 & 1.679 & 2.019  \\
v3 & 1.574 & 1.320 & 1.637 & 1.921  \\
v4 & 1.644 & 1.325 & 1.654 & 1.971  \\
v5 & \textbf{1.556} & \textbf{1.301} & \textbf{1.616} & \textbf{1.908}  \\
\bottomrule
\end{tabular*}

\par\vspace{7pt}
\textbf{c}\quad Multiscale physics-informed neural networks
\par\vspace{2pt}
\begin{tabular*}{\linewidth}{@{\extracolsep{\fill}}lcccc@{}}
\toprule
\textbf{Method}
& \makecell{\textbf{Heat-1D Multiscale}}
& \makecell{\textbf{Heat-2D Multiscale}}
& \makecell{\textbf{PINNACLE Heat}}
& \textbf{Average rank} \\
\midrule
\makecell[l]{Multiscale PINNs} & 0.142 \(\pm\) 0.036 & 0.026 \(\pm\) 0.003 & 1.215 \(\pm\) 0.306 & 4.667 \\
MMPINN & 0.091 \(\pm\) 0.010 & \textbf{0.007 \(\pm\) 0.001} & 1.678 \(\pm\) 1.784 & 4.000 \\
v1 & 0.403 \(\pm\) 0.055 & 0.303 \(\pm\) 0.089 & 0.992 \(\pm\) 0.001 & 6.333 \\
v2 & 0.284 \(\pm\) 0.028 & 0.121 \(\pm\) 0.018 & 0.993 \(\pm\) 0.006 & 5.333 \\
v3 & 0.264 \(\pm\) 0.005 & 0.427 \(\pm\) 0.037 & 1.103 \(\pm\) 0.004 & 6.667 \\
v4 & 0.080 \(\pm\) 0.001 & 0.197 \(\pm\) 0.020 & 0.826 \(\pm\) 0.002 & 3.333 \\
v5 & 0.131 \(\pm\) 0.032 & 0.107 \(\pm\) 0.012 & 0.963 \(\pm\) 0.004 & 3.333 \\
v6 & \textbf{0.067 \(\pm\) 0.012} & 0.151 \(\pm\) 0.021 & \textbf{0.431 \(\pm\) 0.014} & \textbf{2.333} \\
\bottomrule
\end{tabular*}
\end{table}

\begin{table}[!htbp]
\centering
\footnotesize
\setlength{\tabcolsep}{2.5pt}
\renewcommand{\arraystretch}{1.12}

\caption{\textbf{Evidence-supported and rejected claims across case studies.}
For each case, supporting and limiting evidence defines the scope of the final claim.}
\label{tab:ed_claim_boundaries}

\begingroup
\renewcommand{\tabularxcolumn}[1]{m{#1}}

\begin{tabularx}{\linewidth}{
    @{}
    >{\raggedright\arraybackslash}m{0.10\linewidth}
    >{\raggedright\arraybackslash}m{0.19\linewidth}
    Y
    Y
    Y
    @{}
}
\toprule
\textbf{Case}
& \textbf{Proposed idea}
& \textbf{Supporting evidence}
& \textbf{Negative or limiting evidence}
& \textbf{Final bounded claim} \\
\midrule

Memory
& Training-free atomic notes with constrained slotted evidence retrieval
& Overall F1 increased from 0.306 to 0.391, while average token length decreased from 2844.1 to 1017.2, a 64.2\% reduction.
& Matched component-disabled results were not available for every component.
& Under matched conditions, the complete system improved question-answering performance and token efficiency relative to A-Mem \cite{xu2026mem}; the available evidence supports a system-level comparison. \\

\midrule

Graph forecasting
& Diffusion-complementary residual correction through proposal-space innovation coverage
& The final average MAE was 1.556 and horizon-12 MAE was 1.908. Removing the coverage cell increased clean MAE by 0.0042 and worsened block-40\% degradation from 1.060 to 1.101.
& Removing the hard orthogonal projection changed clean MAE by only 0.0019, less than 0.12\% relative, providing little observed support for attributing a gain to the projection under this protocol.
& On PEMS-BAY, the coverage-based design improved forecasting relative to Graph WaveNet; component tests supported retaining coverage under the evaluated conditions. \\

\midrule

Multiscale PINN
& Coarse--fine fixed-complement scale separation
& Average rank improved from 6.33 to 2.33. The final version obtained the best mean relative \(L_2\) error on \texttt{heat1d\_\allowbreak multiscale} and \texttt{pinnacle\_\allowbreak heat}.
& External baselines remained better on \texttt{heat2d\_\allowbreak multiscale}: MMPINN reached 0.00661 with reported variation 0.000995, and MultiscalePINNs reached 0.0261 with reported variation 0.00331.
& The final design improved the one-dimensional multiscale and PINNACLE heat results, while external baselines remained better on the two-dimensional multiscale task. \\

\bottomrule
\end{tabularx}

\endgroup
\end{table}

\begin{table}[!htbp]
\centering
\small
\renewcommand{\arraystretch}{1.16}
\caption{\textbf{Initial process-level diagnostic observations.}
The original examples from representative AI-Scientist-v2 runs are retained independently of their application domains. They document both observed mismatches and gaps in the retained record; missing evidence alone does not establish confirmed drift under the formal definition in Section~\ref{sec:claim_drift}. These examples are not substituted for the claim-level labels used to calculate CDR.}
\label{tab:ed_diagnostic_observations}
\begin{tabularx}{\linewidth}{@{}p{0.20\linewidth}YY@{}}
\toprule
\textbf{Drift type} & \textbf{What goes wrong} & \textbf{Illustrative observation} \\
\midrule
Evidence-to-idea drift
&
The proposal evolves, but the record no longer shows which prior findings support its key design choices.
&
Many update operations were recorded, yet their links to source evidence, changes in retrieval behaviour and direct tests were not retained. \\
\midrule
Idea-to-implementation drift
&
The code implements a materially different design from the proposal, so the original idea is never directly evaluated.
&
A proposed plug-in for an existing backbone became a separate replacement model, while the controls needed to test the plug-in claim were absent. \\
\midrule
Experiment-to-claim drift
&
The experiments do not isolate the relevant components, or the final conclusion extends beyond the comparisons that were actually run.
&
The system produced quantitative results, but several component roles were not directly tested and available baseline evidence was not reflected in the final summary. \\
\bottomrule
\end{tabularx}
\end{table}

\clearpage
\section{Human Review of 20 Claim--Relation Records}
\label{sec:supp-claim-cases}
The 20 selected records below retain their original case identifiers, claim wording, evidence summaries and automated and human decisions. A supported label applies only to the stated support relation. A drifted label denotes a confirmed mismatch; an unverifiable label denotes insufficient evidence to resolve that relation.

Human review retained 14 automated labels (70.0\%). Five unverifiable records became supported after execution or result records were recovered, and one drifted record became supported after the scope of its literature attribution was clarified. Five records remained unverifiable. Because reviewers could retrieve material beyond the automated bundle, this is a descriptive comparison of selected records under different evidence access. These labels were not used to recalculate aggregate CDR.

\subsection{Evidence-to-idea}

\begin{HAclaimcase}{Case 01: Supported: A bounded design rationale}{box:supported}
\textit{Case 01. Xcientist, PINNs. Evidence-to-idea. Automated supported; human supported.}

\textbf{Claim.} Solutions of multiscale partial differential equations contain spectral structure that a standard coordinate-based neural network can struggle to represent.

\textbf{Evidence and decision.} The retained abstracts described high-frequency learning difficulties and bias affecting multiscale learning. Both assessments accepted this as motivation for a representation change. The abstracts did not establish that the proposed architecture worked; support extended only to the stated background rationale.

\end{HAclaimcase}

\begin{HAclaimcase}{Case 02: Collective literature motivation}{box:supp-case-02}
\textit{Case 02. EvoScientist; Graph-TS. Automated: drifted. Human: supported.}

\textbf{Claim.} The claim attributed motivation for adaptive spatial dependencies and explicit forecast-horizon treatment to several graph-forecasting studies.

\textbf{Evidence and decision.} The retained notes supported spatial and temporal modelling and placed missing-data work among future robustness extensions. The automated audit interpreted the final attribution as incorporating that future extension into the current design rationale. Human review instead read the sentence as a collective motivation, without assigning every feature to every cited paper or asserting a completed missing-data test. The disagreement concerned the scope of attribution, not newly observed performance.
\end{HAclaimcase}

\begin{HAclaimcase}{Case 03: Incomplete support for a joint attribution}{box:supp-case-03}
\textit{Case 03. EvoScientist; PINNs. Automated: unverifiable. Human: unverifiable.}

\textbf{Claim.} The claim cited MscaleDNN, FBPINNs and partition-of-unity methods as motivation for multiscale or local representations.

\textbf{Evidence and decision.} The record included a retrieved MscaleDNN passage, a README classification for FBPINNs, and mainly titles, citations and an internal account for partition-of-unity methods. Both assessments found the independent source material insufficient to verify the complete joint attribution. Human follow-up did not recover the missing partition-of-unity content. In the absence of a contradicting source, the result remained unverifiable rather than confirmed drift.
\end{HAclaimcase}

\begin{HAclaimcase}{Case 04: An unverified negative literature claim}{box:supp-case-04}
\textit{Case 04. AI-Scientist-v2; PINNs. Automated: unverifiable. Human: unverifiable.}

\textbf{Claim.} The claim stated that earlier local-basis and frequency-aware methods had not separately tested the sufficiency of an output basis derived from the PDE and boundary conditions.

\textbf{Evidence and decision.} Abstracts for four related works were present in the fuller archive, but did not establish their complete experimental and ablation scope. Recovering those abstracts therefore did not verify the negative assertion. Both assessments retained an unverifiable label. A source that describes a method without mentioning a test is insufficient to establish that the test was never conducted.
\end{HAclaimcase}

\begin{HAclaimcase}{Case 05: Traceable internal text without independent literature evidence}{box:supp-case-05}
\textit{Case 05. Xcientist; Memory; iter-04. Automated: unverifiable. Human: unverifiable.}

\textbf{Claim.} The claim attributed two findings to MACLA: reusable procedural memory could replace parameter updates in reusable-task settings, and structured memory retained domain and schema constraints.

\textbf{Evidence and decision.} The wording could be traced to an internal evidence-graph summary, but the record lacked an independent abstract, full text or original retrieval return. Both assessments found that the internal account could not independently verify the attributed findings. Traceability to an internal summary established the origin of the wording, not the scientific validity of the attribution.
\end{HAclaimcase}

\subsection{Idea-to-implementation}

\begin{HAclaimcase}{Case 06: A result bound to a completed implementation}{box:supp-case-06}
\textit{Case 06. EvoScientist; Graph-TS. Automated: supported. Human: supported.}

\textbf{Claim.} The claim reported that HCDSR seed 0 improved average, 30-min and 60-min MAE relative to the reference, while 15-min MAE was slightly worse.

\textbf{Evidence and decision.} The command, progress record, metrics and code snapshot in the result directory identified the same completed HCDSR run. Both assessments supported the implementation attribution and excluded smoke-test outputs. The label concerned whether the declared implementation produced the attributed result; it was not a separate judgment of every numerical comparison or broader efficacy.
\end{HAclaimcase}

\begin{HAclaimcase}{Case 07: A supported limitation on component attribution}{box:supp-case-07}
\textit{Case 07. InternAgent-1.5; PINNs. Automated: supported. Human: supported.}

\textbf{Claim.} The claim stated that the residual-gate effect could not yet be separated from random-seed effects.

\textbf{Evidence and decision.} Gated runs existed for seeds 1 and 2, whereas the ungated control was available only for seed 0. Source and run-inventory checks confirmed that matched ungated runs for seeds 1 and 2 were absent from the recorded comparison. Both assessments supported the limitation. A supported claim can therefore acknowledge an unresolved comparison rather than assert a positive component contribution.
\end{HAclaimcase}

\begin{HAclaimcase}{Case 08: An intervention that did not isolate component removal}{box:supp-case-08}
\textit{Case 08. Xcientist; PINNs; iter-02. Automated: drifted. Human: drifted.}

\textbf{Claim.} The claim described the coarse global backbone as essential across all benchmarks because disabling it caused catastrophic degradation.

\textbf{Evidence and decision.} The recorded intervention skipped coarse-model training while retaining a random or frozen coarse predictor in the final output and also altering refinement. Both assessments found that this was not a clean removal of the coarse predictor and did not isolate its contribution. The observed degradation therefore could not establish the claimed independent necessity of that component.
\end{HAclaimcase}

\begin{HAclaimcase}{Case 09: Drifted: Reversed condition identity}{box:drifted}
\textit{Case 09. AI-Scientist-v2, PINNs. Idea-to-implementation. Automated drifted; human drifted.}

\textbf{Claim.} Loss balancing helped the one-dimensional heat task (heat1d) slightly but harmed the two-dimensional (heat2d) and PINNACLE heat tasks.

\textbf{Evidence and decision.} Configuration-linked errors showed the reverse: enabling balancing slightly increased error on heat1d and reduced it on the other two tasks. Enabled/disabled relative $L_2$ errors were 0.852011/\allowbreak 0.850089 for heat1d, 1.198781/\allowbreak 1.275678 for heat2d and 1.000880/\allowbreak 1.573157 for pinnacle\_\allowbreak \allowbreak heat, with lower values better. Both assessments found that the claim assigned the disabled condition to the enabled implementation. This was a documented mismatch, not a missing-record decision.

\end{HAclaimcase}

\begin{HAclaimcase}{Case 10: Unverifiable: Source code without execution identity}{box:unverifiable}
\textit{Case 10. EvoScientist, Memory. Idea-to-implementation. Automated unverifiable; human unverifiable.}

\textbf{Claim.} The reported A-MEM results came from the implementation containing the stated implementation controls.

\textbf{Evidence and decision.} Matching source copies and their hashes did not bind each retained run to its execution-time code and dependencies. Human review therefore preserved the unverifiable label. The gap did not establish that an incorrect implementation had run.

\end{HAclaimcase}

\begin{HAclaimcase}{Case 11: A valid mathematical construction with unresolved execution identity}{box:supp-case-11}
\textit{Case 11. AI-Scientist-v2; PINNs. Automated: unverifiable. Human: unverifiable.}

\textbf{Claim.} The claim asserted exact Dirichlet boundary satisfaction because both carrier and residual terms vanished at the spatial boundary.

\textbf{Evidence and decision.} The retained sine carrier and $x(1-x)y(1-y)$ residual had that property, but earlier and later metrics pointed to different method files. Both assessments lacked adequate command or hash evidence binding the completed runs to the specified source version. The implementation-identity label therefore remained unverifiable even though the mathematical boundary property itself could be checked.
\end{HAclaimcase}

\begin{HAclaimcase}{Case 12: Unverifiable to supported: A recovered execution link}{box:recovered}
\textit{Case 12. InternAgent-1.5, PINNs. Idea-to-implementation. Automated unverifiable; human supported.}

\textbf{Claim.} The comparison at the recorded random seed (seed 0) supported an advantage for the tested heat-semigroup time factor over a cosine alternative, while holding the spatial representation fixed.

\textbf{Evidence and decision.} Human review recovered both execution records (run manifests) and checked matching code fingerprints (source hashes). This linked each completed result to its intended implementation. The seed-0 relative $L_2$ errors were approximately 0.0020101 and 0.1010106, respectively, with lower values better, using the same 128-mode spatial representation. The recovered records resolved the execution-identity gap. The decision supported this implementation comparison, not a general mechanistic claim.

\end{HAclaimcase}

\subsection{Experiment-to-claim}

\begin{HAclaimcase}{Case 13: A bounded pattern across five recorded runs}{box:supp-case-13}
\textit{Case 13. InternAgent-1.5; Graph-TS. Automated: supported. Human: supported.}

\textbf{Claim.} The claim stated that MAE increased with forecast horizon in each of five recorded runs: a reference, a named control and three proposed seeds.

\textbf{Evidence and decision.} Each 12-step MAE vector had 11 positive adjacent differences. Both assessments found that the exact metrics supported the monotonic pattern in all five runs. The conclusion was confined to these recorded vectors, rather than generalized to other runs, datasets or forecasting conditions.
\end{HAclaimcase}

\begin{HAclaimcase}{Case 14: Unequal category counts and a bounded precision caveat}{box:supp-case-14}
\textit{Case 14. AI-Scientist-v2; Memory. Automated: supported. Human: supported.}

\textbf{Claim.} The claim cautioned against assuming equally precise category means when category 3 contained 21 questions and category 4 contained 200.

\textbf{Evidence and decision.} The evaluation record covered a fixed 497-question subset and recorded no errors; both assessments verified the category counts. These counts supported the qualitative caution. They did not estimate standard errors or establish which category actually had greater sampling variance, because those conclusions would also require information about the outcome distributions and sampling process.
\end{HAclaimcase}

\begin{HAclaimcase}{Case 15: A small numerical mismatch with the performance direction intact}{box:supp-case-15}
\textit{Case 15. Xcientist; Graph-TS; iter-01. Automated: drifted. Human: drifted.}

\textbf{Claim.} The claim stated that \texttt{full\_\allowbreak pirg} outperformed \texttt{static\_\allowbreak a0} in all missing-sensor tests, with 60-min MAE reductions of 1.06--1.75\%, while the standard no-dropout test slightly favoured \texttt{static\_\allowbreak a0} at 15 min.

\textbf{Evidence and decision.} All 12 missing-sensor comparisons supported the direction, and the standard-test exception was retained. Recalculation in the source record yielded endpoints of 1.050025\% and 1.756577\%, which round to 1.05--1.76\%. Both assessments therefore identified numerical drift at the reported precision. This mismatch did not reverse the observed advantage or erase its stated exception.
\end{HAclaimcase}

\begin{HAclaimcase}{Case 16: Final error did not establish training stability}{box:supp-case-16}
\textit{Case 16. Xcientist; PINNs; iter-01. Automated: drifted. Human: drifted.}

\textbf{Claim.} The claim attributed consistent training-stability benefits to adaptive loss balancing.

\textbf{Evidence and decision.} The retained evidence contained final relative $L_2$ errors and loss histories but no corresponding definition or comparative analysis of stability; some matched results were also incomplete. Both assessments found that improved endpoint error did not establish improvement in a different property. Human review did not find a relevant stability measure or repeated-run variability analysis. The drift concerned the inference from the reported outcome to the asserted benefit.
\end{HAclaimcase}

\begin{HAclaimcase}{Case 17: Recovered metrics supported an unpaired seed comparison}{box:supp-case-17}
\textit{Case 17. EvoScientist; PINNs. Automated: unverifiable. Human: supported.}

\textbf{Claim.} The claim stated that three Fourier-residual seeds beat a No-Fourier control, with a 97.4860\% reduction based on their mean error.

\textbf{Evidence and decision.} The supplied summary listed proposed errors of 0.006899, 0.005162 and 0.005856, compared with 0.237571 for the control. It reported a reduction of 97.4860468\% calculated at original precision. The automated audit lacked independent completion metrics for two proposed seeds; human retrieval recovered them and supported the rounded percentage and all three directions. The control was run once, so this was not a three-seed paired experiment. The displayed rounded errors are not a substitute for the original-precision calculation.
\end{HAclaimcase}

\begin{HAclaimcase}{Case 18: Primary runs supported the numerical comparison}{box:supp-case-18}
\textit{Case 18. AI-Scientist-v2; PINNs. Automated: unverifiable. Human: supported.}

\textbf{Claim.} The claim reported primary-seed errors of approximately 0.00253, 0.0362 and 0.00378, all below the reference error of 0.1527.

\textbf{Evidence and decision.} The fuller record gave approximately 0.002534, 0.036155 and 0.003782, versus 0.152725, and did not substitute a retry for primary seed 0. The automated input lacked completion metrics for primary seeds 0 and 1. Human follow-up recovered the relevant metrics and confirmed the values, rounding and comparison direction. This result-level support did not resolve the source-version binding gap in Case 11.
\end{HAclaimcase}

\begin{HAclaimcase}{Case 19: A recovered comparison preserved a category-level trade-off}{box:supp-case-19}
\textit{Case 19. InternAgent-1.5; Memory. Automated: unverifiable. Human: supported.}

\textbf{Claim.} The claim stated that H3 seed 0 improved category 2, 3 and 4 F1 relative to the stated reference/control, while category 1 was lower.

\textbf{Evidence and decision.} The supplied record gave differences of approximately $-0.136896$, $+0.006982$, $+0.029585$ and $+0.012269$ on the same 497-question subset. The automated audit lacked the exact completed comparison metrics. Human review recovered reference run\_\allowbreak 0 and H3 run\_\allowbreak 3, supporting all four directions within that fixed subset. The trade-off remained part of the supported descriptive comparison.
\end{HAclaimcase}

\begin{HAclaimcase}{Case 20: Comparator identity changed the interpretation of seed effects}{box:supp-case-20}
\textit{Case 20. InternAgent-1.5; Graph-TS. Automated: unverifiable. Human: supported.}

\textbf{Claim.} The claim distinguished comparisons with a reference and a named control.

\textbf{Evidence and decision.} Against the reference, seed 1 was worse on all four MAE measures and seed 2 improved only the 15- and 30-min measures. Against the named control, seed 0 was better on all four and seeds 1 and 2 were worse on all four. The automated audit lacked the exact selected reference and full metric set. Human review identified \texttt{resumed\_\allowbreak \allowbreak experiment/\allowbreak run\_\allowbreak \allowbreak 0} as the reference and retrieved all five runs, confirming every stated direction. Support remained conditional on these explicit comparators.
\end{HAclaimcase}

\clearpage
\section{Final Designs and Idea Process Across Three Tasks}
\subsection{Task-specific designs produced by \xcientist{}}
\label{sec:supp-task-designs}

\begin{figure}[p]
    \centering
    \includegraphics[width=\linewidth,height=0.83\textheight,keepaspectratio]{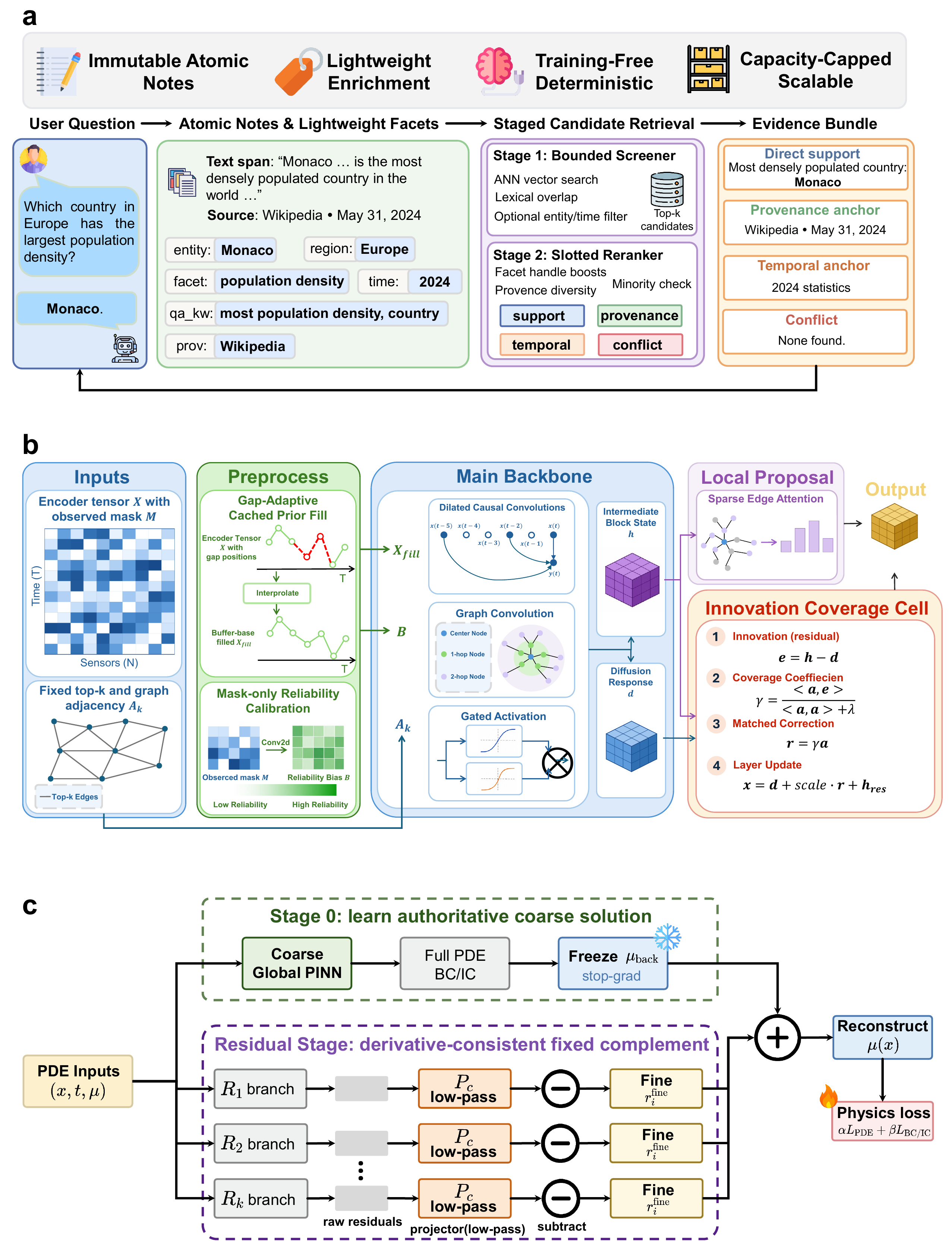}
    \caption{\textbf{Task-specific designs.} \textbf{a}, Atomic-note memory with slotted evidence retrieval. \textbf{b}, Graph WaveNet with innovation coverage. \textbf{c}, Staged coarse--fine PINNs with a fixed complement; Fourier features in the final evaluated version are not shown.}
    \label{fig:xcientist_ideas}
\end{figure}

The three research tasks led to designs with different constraints on what could change and what had to remain identifiable (Figure~\ref{fig:xcientist_ideas}). In memory, the design separated preserved source notes from the evidence selected for an answer. In graph forecasting, it retained a diffusion backbone while revising the operation that supplied local corrections. In physics-informed neural networks, it separated a learned coarse solution from later fine-scale residuals. These component boundaries connected the proposed design to its implementation and defined the comparisons needed to assess its contribution.

\paragraph{Training-free slotted evidence retrieval.}
The memory design preserved immutable atomic notes as the authoritative evidence and organized retrieval around the needs of a question (Figure~\ref{fig:xcientist_ideas}a). Lightweight enrichment attached cached embeddings and compact entity, time, keyword and provenance information without replacing the source note. A bounded first-stage screener combined approximate-nearest-neighbour retrieval with lexical overlap and optional entity or time filtering. A deterministic slotted reranker then assembled a capacity-capped bundle with roles for direct support, provenance, temporal context and conflicting evidence. Sparse facet handles supplied optional ranking boosts, while the deduplication guardrail limited repeated-source contributions and retained bounded minority contradictions. This arrangement made the selection and packing of evidence editable while keeping the underlying record available for inspection. The matched LoCoMo comparisons in the main text supported improved answer quality and reduced token use for the complete system.

\paragraph{Diffusion-referenced innovation coverage for graph forecasting.}
The final forecasting design retained Graph WaveNet as the main predictor and confined the proposed addition to a local correction path (Figure~\ref{fig:xcientist_ideas}b). Gap-adaptive cached prior filling supplied the encoder input, and the observed mask supplied a reliability bias. Sparse edge attention operated within a fixed top-$k$ neighbourhood to produce a correction proposal. The innovation coverage cell scaled that proposal according to its alignment with the state left unexplained by the diffusion response, before combining the matched correction with the backbone and residual path. The resulting design made the contribution of the coverage operation testable without replacing the forecasting backbone. It followed an earlier projection-based proposal, whose removal produced only a small change in clean-data error under the evaluated protocol. Subsequent component tests supported retaining coverage on PEMS-BAY, alongside the final system-level improvement. The projection proposal shown in Figure~\ref{fig:idea_process}b therefore belongs to an earlier design stage than the coverage-based endpoint shown here.

\paragraph{Staged coarse--fine learning with a fixed complement.}
The PINN design assigned the broad solution and its fine-scale corrections to separate stages (Figure~\ref{fig:xcientist_ideas}c). A coarse global branch was first trained using the partial differential equation and boundary and initial conditions, and was then frozen. Later additive branches produced residual functions from which a fixed projector removed the designated coarse contribution before reconstruction. The derivative-consistent formulation was intended to maintain this separation when the reconstructed solution was evaluated through the physics loss. The final experimental version also enriched the coordinate inputs with Fourier features. The retained evaluations supported lower mean error on the one-dimensional multiscale and PINNACLE heat tasks, while external baselines remained better on the two-dimensional multiscale task. The fixed-complement design thus specified how the branches should interact; its performance claim remained conditional on the benchmark results.

\clearpage

\clearpage
\subsection{Component search, candidate fusion and local repair}
\label{sec:supp-ideation-process}
\begin{figure}[!h]
    \centering
    \includegraphics[width=\linewidth,height=0.78\linewidth,keepaspectratio]{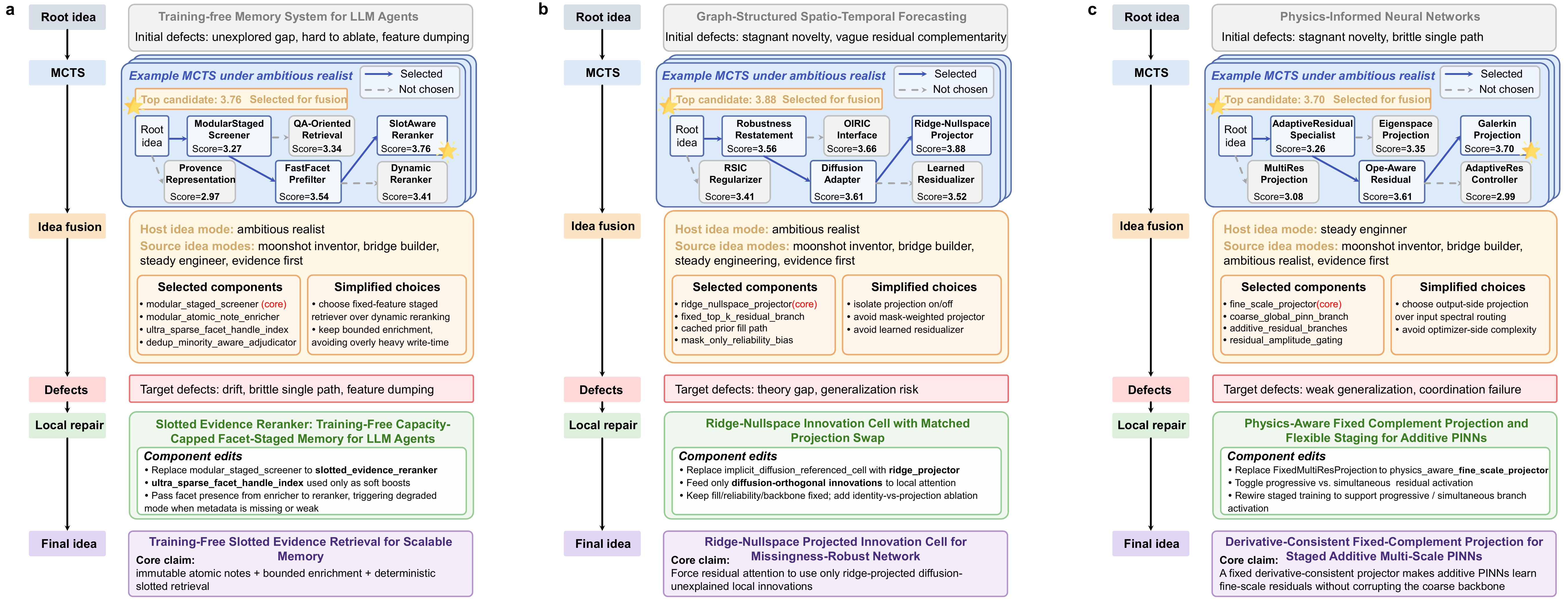}
    \caption{\textbf{Idea search, fusion and repair.} \textbf{a}, Memory. \textbf{b}, Graph forecasting. \textbf{c}, PINNs. Scores evaluate ideas, not experimental performance. The projection proposal in \textbf{b} precedes the experimentally revised coverage design in Figure~\ref{fig:xcientist_ideas}b.}
    \label{fig:idea_process}
\end{figure}

The retained ideation records show how broad research objectives were converted into component specifications and proposed tests (Figure~\ref{fig:idea_process}a--c). Monte Carlo tree search (MCTS) explored edits to structured ideas under different scientific preferences. A host candidate supplied the central question and component structure for fusion, while selected components from other candidates addressed recorded weaknesses. Local repair then revised specific components or interfaces while retaining the research objective. The panels preserve selected and unselected alternatives together with the diagnoses that motivated repair.

The displayed scores are idea-evaluation scores used within search, rather than experimental measurements or evidence that a component worked. Likewise, the ``final idea'' in each panel denotes the output of the displayed ideation and repair sequence. Implementation and experiment validation could subsequently retain, revise or reject its proposed operation. The forecasting example makes this distinction explicit: search selected a projection-based proposal, whereas later component tests motivated the coverage-based design described above.

\subsubsection{Memory: from a staged screener to an explicit evidence bundle}

The memory root left the research gap insufficiently specified and accumulated features whose contributions were difficult to separate (Figure~\ref{fig:idea_process}a). In the displayed search under the ambitious-realist preference, the selected path moved from a modular staged screener through a fast facet prefilter to a slot-aware reranker, which received the highest displayed score of 3.76. Fusion used an ambitious-realist host and drew on the other search preferences to combine screening, atomic-note enrichment, sparse facet handles and deduplication with minority-evidence handling. The retained choices favoured fixed-feature staged retrieval and bounded enrichment.

Local repair made the retrieval output more explicit. It replaced the modular screener component with a slotted evidence reranker, restricted the sparse handle index to soft boosts and passed metadata-availability information from enrichment to retrieval. This last interface allowed retrieval to enter a degraded mode when metadata were absent or weak. The resulting proposal combined immutable atomic notes, bounded enrichment and deterministic slotted retrieval. The record thus connects the final component list to concrete changes in evidence selection and fallback behaviour, rather than only to a renamed memory system.

\subsubsection{Graph forecasting: making residual complementarity testable}

The forecasting root proposed residual correction without a sufficiently explicit account of how it would complement diffusion (Figure~\ref{fig:idea_process}b). The displayed ambitious-realist search progressed through a robustness restatement and a diffusion adapter to a ridge-nullspace projector, which received the highest displayed score of 3.88. Fusion retained the projector, a residual branch with a fixed top-$k$ neighbourhood, cached prior filling and a reliability bias derived only from the observed mask. Mask-weighted projection and a learned residualizer were omitted, keeping the proposed projection operation separable from these alternatives.

Local repair replaced the implicit diffusion-referenced cell with an explicit ridge projector and proposed feeding diffusion-orthogonal innovations to local attention. It also specified an identity-versus-projection comparison with filling, reliability and the backbone held fixed. This made the intended complementarity a component-level test. The later comparison reported in the main text provided little observed support for attributing a gain to the hard projection, prompting its replacement by proposal-space innovation coverage. The ideation record and the subsequent experiment history therefore document different parts of the same revision: the former defines the proposed operation and its test, and the latter supplies the evidence for changing it.

\subsubsection{Physics-informed neural networks: specifying the coarse--fine separation}

The PINN search addressed a root design with limited differentiation from existing approaches and reliance on a single refinement path (Figure~\ref{fig:idea_process}c). The displayed ambitious-realist trajectory progressed from an adaptive residual specialist through an operator-aware residual design to Galerkin projection, with a highest displayed score of 3.70. Fusion used a steady-engineer host and retained a fine-scale projector, a coarse global PINN branch, additive residual branches and residual-amplitude gating. It favoured output-side projection over input spectral routing and avoided additional optimizer-side complexity.

Local repair addressed the recorded generalization and coordination risks by replacing the fixed multiresolution projection with a physics-aware fine-scale projector. It also made progressive and simultaneous residual activation explicit alternatives in the staged training interface. The resulting derivative-consistent fixed-complement proposal specified how residual branches should refine the solution while preserving the designated role of the coarse backbone. Later evaluations tested the design across the three heat-equation settings and retained the two-dimensional limitation described in the main text. The preserved search record explains the intended separation and the comparisons it motivated; the completed experiments determine the scope of the supported conclusion.

\clearpage

\section{Key Instructions and Prompt Contracts}
\label{app:prompt_contracts}
\begin{instruction}{Schema-Bound Evidence-Graph Evidence Extraction}
You are a Senior Research Analyst extracting structured metadata from academic papers.

~\\
CRITICAL GROUNDING RULES:\\
1. Insight must provide independent analysis beyond summary. Describe prerequisites, trade-offs, design choices, or potential defects.\\
2. Quote must be verbatim text from the original paper. Do not paraphrase or fabricate.\\
3. Summary must be a concise factual description.\\
4. Keywords must capture representative concepts for retrieval and filtering.\\

~\\
PRIORITY: RECALL. Extract all relevant cores, components, problems, innovations, limitations, and future work.

~\\
Core versus component: \\
- A core contribution is the largest top-level contribution that is not contained by another entity and can stand alone.\\
- A component is a module, technique, or building block that is part of a core contribution.\\
- If unsure, ask whether the entity can exist independently or is always part of something larger.\\

Pattern-derived entity markers are suggestions only. Read the full paper and extract relevant entities regardless of whether they are marked. 

~\\
Input: \\
- Paper title: \{title\}\\
- Full paper text: \{marked\_text\}\\

~\\
Return JSON containing: \\
- metadata: title, domain, paper\_type, structured\_summary, code\_url\\
- problems\\
- core\_contributions\\
- core\_relations\\
- components\\
- innovations\\
- limitations\\
- future\_work\\

~\\
For every evidential item, preserve:\\
- keywords\\
- factual summary\\
- independent insight\\
- verbatim quote\\
- relation to the relevant core contribution\\

~\\
Components must include design choices and potential defects.
Return JSON only.
\end{instruction}

\begin{instruction}{Baseline and Dataset Relation Extraction Instruction}
You are a Research Graph Builder focusing on relationship extraction.

~\\
For baselines and datasets, summary, insight, and quote must describe the relationship or comparison, not merely the entity itself.

~\\
PRIORITY: RECALL > PRECISION.\\
Extract baselines and datasets from experimental results, related work, appendices, and theoretical comparisons.

~\\
A baseline may be:\\
- a directly compared competitor;\\
- a predecessor the work builds upon or repairs;\\
- a related alternative or inspiration;\\
- a method in a theoretical comparison;\\
- an appendix comparison;\\
- an evaluated method for a dataset or analysis paper;\\
- an ablation variant.\\

~\\
Constraints:\\
- Never include the paper's own full core contribution as a baseline.\\
- Verify that every extracted entity appears in the paper.\\
- Preserve the related core, retrieval keywords, metrics, relationship summary, independent insight, and verbatim supporting quote.\\

~\\
Input:\\
- Proposed core methods: \{core\_names\}\\
- Full paper text: \{marked\_text\}\\

~\\
Return JSON with `baselines` and `datasets` only.
\end{instruction}

\begin{instruction}{Cross-Paper Analytical Synthesis and Citation-Constrained Drafting}
You are an expert research analyst. Given clusters of paper keynotes:\\
1. Propose a small set of high-value questions involving relationships among multiple papers, such as comparisons, shared assumptions, conflicting claims, technical lineage, unresolved gaps, or future directions.\\
2. Answer each question by synthesizing evidence across the related papers.\\
3. Produce cross-cluster analysis that identifies patterns, differences, connections, unresolved issues, and research gaps.

~\\
Grounding constraints:\\
- Cite only papers supplied in the input.\\
- Use exact paper-title citations.\\
- Do not repeat the input verbatim.\\
- Prefer analytical synthesis over paper-by-paper listing.\\
- Keep each answer concise while preserving the evidence chain.\\

~\\
Inputs:\\
- Cluster content: \{cluster\_content\}\\
- Related papers: \{related\_papers\_content\}\\
- Cross-cluster analyses: \{cluster\_analysis\_content\}\\
\end{instruction}

\begin{instruction}{Survey-Led Advanced Analysis and Root-Idea Calibration}
You are the lead author preparing a top-tier paper on topic \{topic\}.

~\\
Inputs:\\
- Mature idea and its source: \{mature\_idea\}, \{mature\_idea\_source\}\\
- Refinement scope and its source: \{refinement\_scope\},
  \{refinement\_scope\_source\}\\
- Survey contents: \{survey\_contents\}\\
- Curated cited-paper capsules: \{papers\}\\
- Optional experiment findings: \{experiment\_findings\}\\

~\\
Core principles:\\
- The survey drives the agenda. Method clusters, unresolved mechanism bottlenecks, and evaluation blind spots must originate from the survey.\\
- Paper capsules may substantiate survey-derived bottlenecks, baselines, feasibility constraints, and concrete instantiations, but must not redefine the main problem axis.\\
- This stage is a 1.0-to-1.1 calibrator, not a 2.0 invention stage.\\
- Keep an explicit mature idea on the same topic, hypothesis, and primary mechanism axis. Apply localized corrections only.\\
- Treat an explicit refinement scope as a hard edit boundary.\\
- Experiment findings may invalidate a weak component or motivate a local replacement, but must not trigger an unsupported paradigm shift.\\
- Prioritize mechanism bottlenecks over evaluation tooling.\\
- Do not introduce a new primary mechanism family before identifying the specific weak, brittle, underspecified, or unsupported part of the current idea.\\
- Prefer repairing an existing rule, objective, representation, write policy, consolidation rule, or training contract over adding a gate, router, controller, monitor, or threshold wrapper.\\
- If no valuable mechanism-level local patch exists, preserve the mature idea.\\
- Preserve training-free or inference-time character unless new training is indispensable and explicitly justified.\\

~\\
Required analysis:\\
1. Map survey-led method clusters and summarize their assumptions, training signals, and operating constraints.\\
2. Extract unresolved mechanism bottlenecks from the survey, then identify evaluation blind spots that prevent clean measurement.\\
3. Produce one search-ready root idea that applies the smallest meaningful repair while preserving the main method axis.\\
4. Specify only the validation tools needed to falsify the proposed repair.\\

~\\
Output strict JSON containing:\\
- key\_methods, field\_consensus, existing\_problems, evaluation\_gaps;\\
- preserve\_current\_idea decision;\\
- grounded mature idea and refinement scope;\\
- exactly one root\_idea with contribution, method, risks, target defects, rationale, and evidence anchors;\\
- at most one local divergent seed;\\
- bounded cross-domain inspiration;\\
- concise synthesis.
\end{instruction}

\begin{instruction}{Memory-Guided MCTS Idea Expansion}
You control the expansion step of a memory-guided MCTS that iteratively rewrites research ideas.

~\\
Mission:\\
- Surface strong mechanism-level research concepts rather than incremental safeguards or evaluation-only proposals.\\
- Prefer bold but coherent mechanism commitments.\\
- Import cross-domain mechanisms only when they repair a diagnosed weakness.\\
- Never present a benchmark, protocol, or dataset as the primary contribution.\\

~\\
Inputs:\\
- Topic: \{topic\}\\
- Current idea state: \{current\_summary\}\\
- Literature context: \{paper\_context\}\\
- Retrieved natural-language memory: \{memory\_bundle\}\\
- Allowed edit operators: \{edit\_operators\}\\
- Global constraints: \{constraints\}\\

~\\
Apply exactly one provided edit operator per child. Never invent operators.

~\\
Each child must:\\
1. Target at least one explicit defect.\\
2. Explain why the selected operator repairs the defect without feature dumping, unfair comparisons, hidden failure modes, or resource violations.\\
3. Provide a structured idea payload containing title, abstract, core contribution, method, experiments, risks, and tags.\\
4. Reference the memory snippets actually used.\\
5. Introduce a concrete algorithmic intervention. Instrumentation-only ideas are invalid unless they support a substantive mechanism change.\\
6. Sharpen a scientific thesis, repair a weak assumption, propose a stronger principle, or reframe the parent idea on the same method axis.\\
7. State whether the idea clears an expert-review bar and why.\\

~\\
Return up to \{max\_children\} mutually distinct children as strict JSON.
\end{instruction}

\begin{instruction}{Evidence-Aware MCTS Referee Evaluation}
You score research ideas encountered during a memory-guided MCTS search.

~\\
Inputs:\\
- topic and fixed root domains;\\
- mature-idea anchor and refinement boundary;\\
- component-level edit plan;\\
- candidate idea;\\
- canonical defect registry;\\
- symbolic memory derived from component ablations.\\

~\\
Scoring policy:\\
- Prefer concrete mechanism-level edits over vague incremental changes.\\
- Treat evaluator, contract, audit, gate, router, controller, and threshold additions as scaffolding unless they enable a distinct task-solving mechanism.\\
- Reward validation only when it is the lightest protocol needed to falsify the core mechanism.\\
- Penalize feature dumping, unsupported complexity jumps, scope drift, and unjustified changes from training-free to trained systems.\\
- Interpret symbolic ablation memory as follows:
  * removing a component helped: the component family is risky or redundant;
  * removing a component hurt: the component family is beneficial;
  * inconclusive evidence must not dominate scoring.\\

Independently score novelty, surprise, feasibility, clarity, impact, risk, conciseness, alignment, complexity penalty, and protocol quality.\\
Return failure modes, fairness protocol, actionable feedback, defect-fix summary, and all remaining canonical defects.
Return strict JSON only.\\
\end{instruction}

\begin{instruction}{Evidence-Graph Novelty Instruction}
Evaluate mechanism novelty using:\\
- candidate idea state: \{idea\_state\};\\
- component explanations: \{components\_with\_explanations\};\\
- nearest evidence-graph evidence nodes: \{retrieved\_nodes\}.\\

~\\
Rules:\\
- Score mechanism novelty only; ignore impact, feasibility, and writing.\\
- High textual similarity does not force low novelty if the mechanism shift is concrete and substantive.\\
- Low similarity does not imply novelty when the idea is vague or generic.\\
- If one structurally sound component fundamentally changes capability, do not simply average it away with standard components.\\
- Distinguish cosmetic restatement, modest variation, meaningful recombination, clear mechanism departure, and strongly differentiated mechanism.\\

~\\
Return strict JSON with retrieval similarity, perceived novelty, rubric score, and one grounded rationale.
\end{instruction}

\begin{instruction}{Cross-Mode Idea Fusion and Bounded Local Repair}
You are a component-level fusion agent.

~\\
Inputs:\\
- topic, mature idea, refinement scope, root domains, and analysis summary;\\
- candidate ideas generated from different idea-taste modes.

~\\
Fusion rules:\\
- Do not union all ideas.\\
- Select exactly one dominant core mechanism.\\
- Retained components from other candidates must act only as support modules, protocols, or guardrails.\\
- Copy reused component names exactly.\\
- Protocol, evaluator, audit, gate, router, controller, or threshold  components cannot become the main novelty merely because they are easy to combine.\\
- Reject components that are redundant, conflicting, outside the refinement scope, or only repair complexity created by another weak component.\\
- The final idea must express one method with one clear causal chain.

~\\
Return:\\
- the fused idea;\\
- host mode;\\
- selected and rejected components with reasons;\\
- conflicts and resolutions;\\
- one fused core thesis;\\
- why the result is stronger;\\
- a minimal validation plan.
\end{instruction}

\begin{instruction}{Contract-Governed Code Enablement}
Experiment workspace rules:\\
- Treat `idea.json` as the only canonical structured experiment input.\\
- Restrict reads and writes to declared workspace-local paths.\\
- Keep runtime code self-contained under `project/`; reference repositories are reference-only.\\
- Return structured failure evidence when work cannot complete.\\
- Only the dedicated integrator may write the final ablation artifact.

~\\
Planner mandate:\\
- Translate the full idea into an executable code DAG supporting baseline, full-method, and per-component ablation conditions.\\
- Ground every step in real prepared targets.\\
- Map every canonical component to implementation nodes and an ablation mechanism.\\
- Declare input paths, allowed write roots, required outputs, verification commands, repair limits, and done conditions.\\
- End with `final\_integration\_smoke` using real prepared data and model/API bindings.\\
- Require provenance mapping for any code copied from reference repositories.

~\\
Worker mandate:\\
- Implement exactly one assigned step contract inside its declared write scope.\\
- Prioritize validator feedback on repair attempts.\\
- Make the real integration changes needed for the experiment path.\\
- Never use path injection, editable installs, or external local-path imports to hide dependencies.\\
- Use real data and real prepared models or APIs; do not use synthetic data, random embeddings, or mocks.

~\\
Validator mandate:\\
- Judge from the step contract, worker report, changed files, and real execution evidence, not summaries.\\
- Import success alone is insufficient.\\
- Fail changes outside the project scope, hidden external dependencies, missing provenance manifests, or fake smoke tests.\\
- Pass the final integration smoke only after a bounded real end-to-end run.
\end{instruction}

\begin{instruction}{Component-Complete Science Validation and Convergence}
Science-planning mandate:\\
- Build standard baseline-versus-full-method comparisons using the same real prepared dataset and target bindings.\\
- Build exactly one ablation step for every canonical component.\\
- Derive components only from `idea.json.components`, preserving exact names, explanations, and order.\\
- Distinguish smoke/debug tasks from final full-dataset experiments.\\
- Do not invent datasets, models, API bindings, or substitute targets.

~\\
Worker mandate:\\
- Execute exactly one science contract through the real command chain.\\
- Preserve raw evidence before summarizing.\\
- For ablation, test exactly one canonical component and preserve the exact method-context change.

~\\
Validator mandate:\\
- Judge from raw evidence and artifacts, not self-reported summaries.\\
- Fail runs that do not use declared prepared targets and real data.\\
- For ablation, require exact component identity and an explicit description of the tested method context.\\
- Return PASS, PARTIAL, or FAIL with structured evidence.

~\\
Component-coverage contract:\\
- Every component is a required validation target.\\
- Do not rename, merge, split, omit, or reorder components.
- Fixed components are not exempt; define a degraded fallback or report a blocker.

~\\
Convergence contract:\\
- Prefer machine-readable status and validator reports over narrative summaries.\\
- Code must pass before standard science; standard science must pass before ablation science.\\
- Smoke, debug, and subset runs cannot satisfy final evidence requirements.\\
- Do not declare completion if implementation flaws invalidate scientific conclusions.\\
- Final ablation evidence must cover exactly all canonical components in the same order, with no extras or omissions.
\end{instruction}

\begin{instruction}{Source-Fidelity and Claim-Boundary Audit}
You are a technical report auditor.

~\\
Extract and verify:\\
- function and class names;\\
- parameters and default values;\\
- file paths;\\
- architecture and component relationships;\\
- quantitative statistics and experimental claims;\\
- citations and references;\\
- figures and pseudocode.

~\\
Source-fidelity checks:\\
1. Verify that every named function and class exists in the source.\\
2. Verify that every parameter value and type matches the implementation.\\
3. Verify that every referenced file path exists.\\
4. Verify that architecture descriptions match actual component relationships.\\
5. Verify experimental values against raw result artifacts.

~\\
Research-integrity checks:\\
1. Verify that every cited paper exists in the evidence workspace.\\
2. Read the cited source and confirm that the attributed claim is supported.\\
3. Verify that every reference is cited in the report body.\\
4. Flag fabricated statistics, unsupported conclusions, and missing evidence.

~\\
Refinement rules:\\
- Verify every reported issue before changing the report.
- Always fix confirmed critical source-fidelity, code-accuracy, or fabricated claim issues.\\
- Make minimal and precise changes.\\
- Preserve correct content, authorial intent, and code-block integrity.\\
- Apply consistency fixes globally when a function name, parameter, or path is corrected.\\
- Skip unverified, ambiguous, or purely stylistic changes.

~\\
Return a structured audit containing scores, checked evidence, critical issues, and prioritized repairs.
\end{instruction}

\clearpage

\end{document}